%% file: example_paper.tex
\documentclass{article}

\usepackage{microtype}
\usepackage{graphicx}
\usepackage{subfigure}
\usepackage{booktabs} 

\usepackage{hyperref}

\usepackage[accepted]{icml2024}

\usepackage{amsmath}
\usepackage{amssymb}
\usepackage{mathtools}
\usepackage{amsthm}
\usepackage{graphicx}
\usepackage{booktabs}
\usepackage{algorithm}
\usepackage{algorithmic}
\usepackage[switch]{lineno}
\usepackage{multirow}
\usepackage{subfigure}
\input{math_defi.tex}

\usepackage[capitalize,noabbrev]{cleveref}

\theoremstyle{plain}
\newtheorem{theorem}{Theorem}[section]

\theoremstyle{definition}

\newtheorem{assumption}[theorem]{Assumption}
\theoremstyle{remark}

\usepackage[textsize=tiny]{todonotes}

\icmltitlerunning{An Empirical Examination of Balancing Strategy for Counterfactual Estimation on Time Series}

\begin{document}

\twocolumn[
\icmltitle{An Empirical Examination of Balancing Strategy \\ for Counterfactual Estimation on Time Series}




\begin{icmlauthorlist}
\icmlauthor{Qiang Huang}{aff1,aff2}
\icmlauthor{Chuizheng Meng
}{aff4}
\icmlauthor{Defu Cao}{aff4}
\icmlauthor{Biwei Huang}{aff5}
\icmlauthor{Yi Chang}{aff1,aff2,aff3}
\icmlauthor{Yan Liu}{aff4}
\end{icmlauthorlist}

\icmlaffiliation{aff1}{School of Artificial Intelligence, Jilin University, Changchun, Jilin, China}
\icmlaffiliation{aff2}{International Center of Future Science, Jilin University, Changchun, Jilin, China}

\icmlaffiliation{aff3}{Engineering Research Center of Knowledge-Driven Human-Machine Intelligence, MOE, Changchun, Jilin, China}

\icmlaffiliation{aff4}{Department of Computer Science, University of Southern California, California, Los Angeles, United States}
\icmlaffiliation{aff5}{Halicioğlu Data Science Institute, University of California San Diego, San Diego, California, United States}

\icmlcorrespondingauthor{Yi Chang}{yichang@jlu.edu.cn}
\icmlcorrespondingauthor{Yan Liu}{yanliu.cs@usc.edu}

\icmlkeywords{Treatment Effect, Time Series, Balancing Strategy}

\vskip 0.3in
]



\printAffiliationsAndNotice{} 

\begin{abstract}
Counterfactual estimation from observations represents a critical endeavor in numerous application fields, such as healthcare and finance, with the primary challenge being the mitigation of treatment bias. The balancing strategy aimed at reducing covariate disparities between different treatment groups serves as a universal solution. However, when it comes to the time series data, the effectiveness of balancing strategies remains an open question, with a thorough analysis of the robustness and applicability of balancing strategies still lacking. This paper revisits counterfactual estimation in the temporal setting and provides a brief overview of recent advancements in balancing strategies. More importantly, we conduct a critical empirical examination for the effectiveness of the balancing strategies within the realm of temporal counterfactual estimation in various settings on multiple datasets. Our findings could be of significant interest to researchers and practitioners and call for a reexamination of the balancing strategy in time series settings.
\end{abstract}

\section{Introduction}
Temporal counterfactual outcome estimation \cite{cao2023estimating, morgan2015counterfactuals,hernan2010causal,pearl2009causal,brodersen2015inferring,liu2022practical, meng2023costar} aims to predict what the outcome would have been under different treatment scenarios. It is a crucial task in various real-world applications, such as health care \cite{prosperi2020causal,richens2019counterfactual, cao2023estimatingws}, finance \cite{lundberg1992counterfactuals,dhar1998data,castro2019does}, social media~\cite{zhang2022counterfactual}, and e-commerce \cite{hua2021markdowns,goswami2022theory}. For example, in personalized medicine \cite{jain2002personalized,pazzagli2018methods,porcher2019identifying}, counterfactual outcome provides a more comprehensive understanding of how the patient may respond to each treatment over time, allowing for a data-driven decision regarding the most suitable treatment strategy; In e-commerce, counterfactual inference can give companies guidance on when and to whom to issue coupons to different groups of users for increasing sales.

Estimating counterfactual outcomes is inherently challenging, principally for two reasons. Firstly, the intrinsic nature of observed data precludes the direct observation of counterfactuals, rendering their estimation a complex endeavor as acknowledged in the literature \cite{mandel1996counterfactual,pearl2009causal,boninger1994counterfactual,hassanpour2019learning}. These outcomes, delineating the hypothetical scenario under alternative treatments, are intrinsically unobservable. Furthermore, the confounding variables further convolute this task \cite{mcnamee2003confounding,vanderweele2013definition,jager2008confounding}, since they affect both the treatment and the outcome, leading to treatment bias and obscuring the true causal effect. This results in a disparity between the distributions of observed (factual) and unobserved (counterfactual) outcome and makes   accurate estimation of counterfactuals extremely difficult.

To address above challenges, the research community harnessed balancing technologies and developed a series of work to address the confounding issue. Strategies like Inverse Probability of Treatment Weighting (IPTW) \cite{chesnaye2022introduction,allan2020propensity}, stratification \cite{imbens2015causal,miratrix2013adjusting} and matching \cite{abadie2004implementing,abadie2006large}, create pseudo-populations to mitigate distributional discrepancies between treatment groups. Alternative approaches, such as G-computation \cite{robins1986new,robins1987graphical}, iteratively simulate potential outcomes based on Monte Carlo. With the advance of deep learning, a new wave of methodologies has emerged to uncover complex, nonlinear relationships in data. These methods, as explored in studies by \cite{shalit2017estimating,hartford2017deep,atan2018deep,yao2018representation,hassanpour2019learning}, are geared towards encoding covariates into a latent space. The goal is to derive representations that are devoid of treatment-related information and minimize the correlations between the representations and the treatment.

\begin{figure}[t]
\centering
\subfigure[Slight treatment bias.]{
\begin{minipage}{0.5\linewidth}
\centering
\includegraphics[width=1.6in]{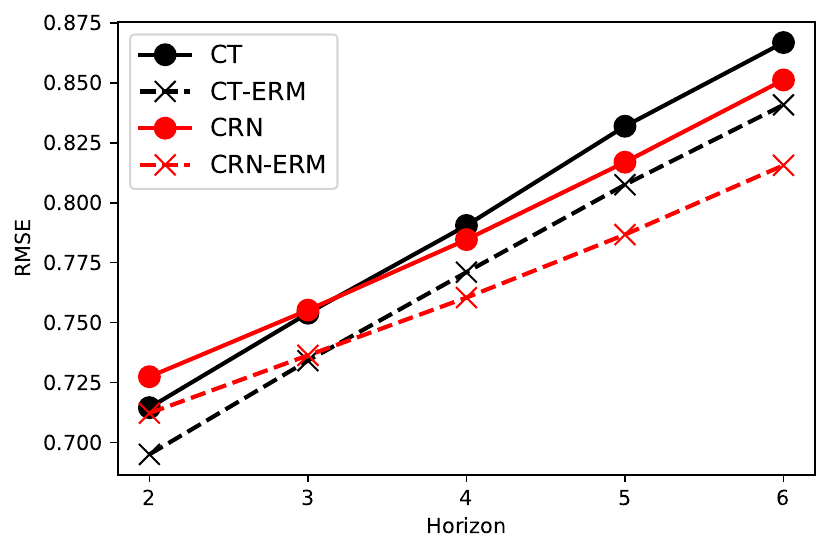}
\end{minipage}%
}%
\centering
\subfigure[Significant treatment bias.]{
\begin{minipage}{0.5\linewidth}
\centering
\includegraphics[width=1.6in]{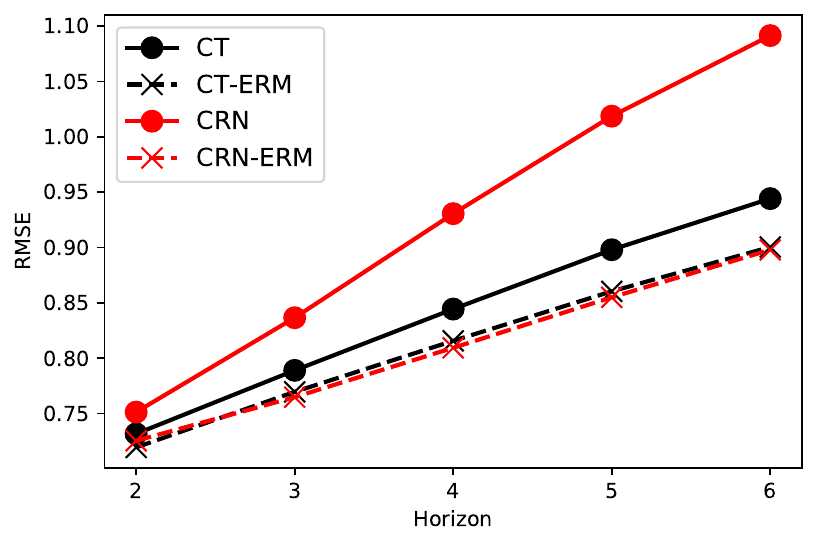}
\end{minipage}%
}%
\centering
\caption{Performance comparison between balanced models (Causal Transformer and CRN) and their ERM variants in multi-step counterfactual estimation on Tumor Growth dataset, the X-axis represents Horizon, Y-axis represents RMSE, lower is better.}
\label{example_comparison}
\vspace{-3mm}
\end{figure}

Despite the outlined effectiveness of balancing strategies in estimating counterfactual outcomes, our initial empirical investigation into the time series scenarios unveiled a counterintuitive phenomenon: in a temporal setting, models that forego balancing strategies—specifically, empirical risk minimization (ERM)—demonstrate superior performance in counterfactual estimation tasks compared to their counterparts that impose balancing strategies, even in the presence of treatment bias.   Figure \ref{example_comparison} illustrates two examples, i.e., Causal Transformer \cite{melnychuk2022causal} and CRN  \cite{bica2020time}. We observed that their ERM variants consistently outperforms Causal Transformer and CRN in multi-step prediction tasks (in either slight and significant treatment bias conditions).  This surprising observation deviates from the prevailing perception on the effectiveness of balancing strategies and calls for an in-depth examination.

In this study, we undertake a comprehensive examination of contemporary methodologies devised to mitigate treatment bias via balancing techniques in time series analysis. We not only demonstrate how these balancing strategies perform but also discuss the underlying reasons through more in-depth empirical analyses. This paper elucidates the development in balancing methods for temporal counterfactual estimation, sheds light into when and how balance can be achieved in time series scenario, and discusses when it may be an elusive goal through empirical studies. We hope that the analysis results provide better understanding on balancing strategy in temporal counterfactual estimation and useful insights to researchers and practitioners.

The organization of this paper is as follows: Section 2 provides a review of the problem setting and basic assumptions of temporal counterfactual outcomes. In Section 3, we briefly revisit the advancements in balancing strategies from the perspective of balancing types and discusses popular methods in practical applications. Section 4 conducts empirical studies of these temporal causal models on benchmark datasets and discusses the effectiveness of balancing strategies in a temporal setting. Finally, Section 5 summarizes our empirical analysis and sheds light into promising directions to fundamentally address the problem of counterfactual estimation for time series.

\begin{table*}[t]
\centering
\caption{Summary of partial literature for counterfactual estimation.}
\vskip 0.10in
\label{tab:my-table}
\scalebox{0.9}{
\begin{tabular}{ccccc}
\hline
\multicolumn{1}{c|}{Taxonomy}                      & \multicolumn{1}{c|}{Methods}                                                                                             & \multicolumn{1}{c|}{Blancing Type}        & \multicolumn{1}{c|}{Balancing Tech}        & Backbone Type        \\ \hline
\multicolumn{1}{c|}{\multirow{6}{*}{Non-temporal}} & \multicolumn{1}{c|}{\begin{tabular}[c]{@{}c@{}}Statistical Method\\ (Re-weighting,Stratification,Matching)\end{tabular}} & \multicolumn{1}{c|}{Pseudo-Population}    & \multicolumn{1}{c|}{Re-sampling}           & N/A                  \\
\multicolumn{1}{c|}{}                              & \multicolumn{1}{c|}{CB-IV}                                                                                               & \multicolumn{1}{c|}{IV-Regression}        & \multicolumn{1}{c|}{IPM Balancing}         & DNN                  \\
\multicolumn{1}{c|}{}                              & \multicolumn{1}{c|}{Deep-Treat}                                                                                          & \multicolumn{1}{c|}{Representation-Based} & \multicolumn{1}{c|}{Propensity Inverse}    & DNN                  \\
\multicolumn{1}{c|}{}                              & \multicolumn{1}{c|}{CITE}                                                                                                & \multicolumn{1}{c|}{Representation-Based} & \multicolumn{1}{l|}{Contrastive Balancing} & DNN                  \\
\multicolumn{1}{c|}{}                              & \multicolumn{1}{c|}{CFR}                                                                                                 & \multicolumn{1}{c|}{Representation-Based} & \multicolumn{1}{c|}{IPM Balancing}         & DNN                  \\
\multicolumn{1}{c|}{}                              & \multicolumn{1}{c|}{MIM-DRCFR}                                                                                           & \multicolumn{1}{c|}{Representation-Based} & \multicolumn{1}{c|}{MI Minimization}       & DNN                  \\ \hline
\multicolumn{1}{c|}{\multirow{5}{*}{Temporal}}     & \multicolumn{1}{c|}{MSM}                                                                                                 & \multicolumn{1}{c|}{Reweighting}          & \multicolumn{1}{c|}{Propensity Inverse}    & \multicolumn{1}{l}{Linear Regression} \\
\multicolumn{1}{c|}{}                              & \multicolumn{1}{c|}{GNET}                                                                                                & \multicolumn{1}{c|}{G-compupation}        & \multicolumn{1}{c|}{Monte-Carlo}           & LSTM                 \\
\multicolumn{1}{c|}{}                              & \multicolumn{1}{c|}{RMSN}                                                                                                & \multicolumn{1}{c|}{Reweighting}          & \multicolumn{1}{c|}{Propensity Inverse}    & LSTM                 \\
\multicolumn{1}{c|}{}                              & \multicolumn{1}{c|}{CRN}                                                                                                 & \multicolumn{1}{c|}{Representation-Based} & \multicolumn{1}{c|}{Gradient Reversal}     & LSTM                 \\
\multicolumn{1}{c|}{}                              & \multicolumn{1}{c|}{Causal Transformer (CT)}                                                                                  & \multicolumn{1}{c|}{Representation-Based} & \multicolumn{1}{c|}{Domain Confusion}      & Transformer          \\ \hline
\multicolumn{5}{c}{\begin{tabular}[c]{@{}c@{}}Statistical methods \cite{rosenbaum1983central,miratrix2013adjusting,abadie2004implementing}, CB-IV \cite{wu2022instrumental},\\ Deep-Treat \cite{atan2018deep}, CITE \cite{li2022contrastive}, CFR \cite{shalit2017estimating}, MIM-DRCFR \cite{cheng2022learning},\\ MSM\cite{robins2000marginal},GNET\cite{li2021g},RMSN \cite{lim2018forecasting},CRN \cite{Bica2020Estimating},CT\cite{melnychuk2022causal}, \\ IV denotes the Instrumental Variable, IPM represents the Integral Probability Metric and MI denotes the Mutual Information. \end{tabular}}                                                                                                                                                                                                                        \\ \hline
\end{tabular}
}
\end{table*}

\section{Foundations}
\begin{figure}[t]
    \centering
\includegraphics[width=3.2in]{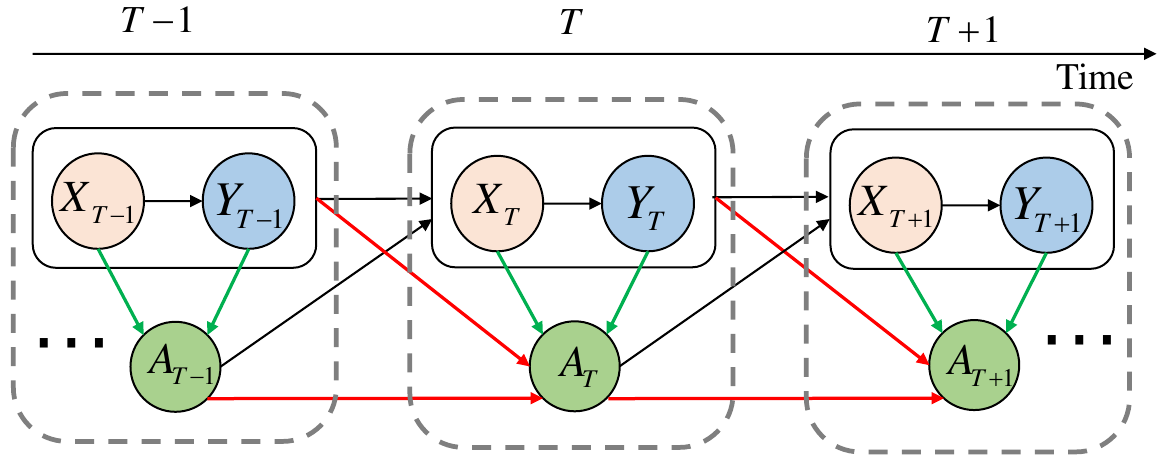}
    \caption{Causal structure in temporal setting, take time step $T-1$ to $T+1$ as an example, where green and red arrows denote the current and time-varying treatment bias, respectively.}
    \label{causal_structure}
\vspace{-5mm}
\end{figure}

\noindent \textbf{Notations}. Let $i$ denote the $i$-th individual (a single unit of analysis within a study, e.g., a patient) with historical trajectories over multiple time steps from $t=1,...T^{(i)}$. For each time step $t$, each individual $i$ has the following observations: the time-varying covariates $\boldsymbol{X}_t^{(i)} \in \mathcal{R}^{d_x}$, where $d_x$ is the dimension of the time-varying covariates; the static covariates $\boldsymbol{V}^{(i)}$ that do not change over time; the treatment assignments $\boldsymbol{A}_t^{(i)} \in \{a_1,...,a_{d_a}\}$ where $d_a$ is the number of treatment variables; the outcome $\boldsymbol{Y}_t^{(i) \in \mathcal{R}^{d_Y}}$. Then we let $\mathcal{D} = \{\{\boldsymbol{x}_t^{(i)}, \boldsymbol{a}_t^{(i)}, \boldsymbol{y}_t^{(i)}\}_{t=1}^{T^{(i)}}\}_{i=1}^N$ denote the observation dataset. For simplicity, we omit the superscript $(i)$ for each individual unless needed and use $\boldsymbol{X}_t$ to represent both static and time-varying covariates at time step $t$.

\textbf{Causal Structure}. Figure \ref{causal_structure} illustrates the causal structure and relationships between key variables in a temporal setting. $X_t$ represents the time-varying covariates at time step $t$, $Y_t$ denotes the outcome at time $t$, and $A_t$ is the treatment assigned at time step $t$. The diagram shows that at each time step, the treatment assignment $A_t$ is influenced by the time-varying covariates $X_t$ as well as the history, indicating the presence of time-varying confounders. The outcome $Y_t$ is affected by the historical treatment $A_{t-1}$, previous outcome $Y_{t-1}$, and current covariates $X_t$.

\textbf{Target}: Estimating $\E(\vy_{t+\tau}[\bar{\va}_{t:t+\tau-1}] \vert \bar{\mH}_t)$, where $\bar{\mH}_t = (\bar{\mX}_t, \bar{\mA}_{t-1}, \bar{\mY}_t, \mV)$ is the observed history. $\bar{\mX}_t = (\vx_1, \vx_2, \dots, \vx_t)$, $\bar{\mA}_{t-1} = (\va_1, \va_2, \dots, \va_{t-1})$, $\bar{\mY}_t = (\vy_1, \vy_2, \dots, \vy_t)$, and $\mV = \vv$. $\tau \geq 1$ denotes the projection horizon for a $\tau$-step-ahead prediction. $\bar{\va}_{t:t+\tau-1} = (\va_t, \va_{t+1}, \dots, \va_{t+\tau - 1})$ is the sequence of the applied treatments in the future $\tau$ discrete time steps.

\section{Balancing Techniques}
Table \ref{tab:my-table} provides a concise overview of counterfactual estimation methods from the perspective of balancing strategy. These methods vary in their balancing techniques, ranging from re-sampling and propensity inverse to more complex strategies like gradient reversal and domain confusion. It can be seen that the development of the solutions follow the paths towards leveraging advanced machine learning for causal inference. The table only lists a selection of representative methods. Many other models based on the balancing strategies not mentioned in the table, such as \cite{hartford2017deep,yao2018representation,hassanpour2019learning,gilbertson2016controlling,wodtke2020regression,liu2020estimating}, because the balancing strategies adopted by these unlisted methods are already included in Table \ref{tab:my-table}.

Representation-based balancing strategies have received significant attention in recent years, especially those based on deep learning, given  its high performance in estimation accuracy. Therefore, in this work, we focus our discussion on  representation-based balancing strategies for counterfactual estimation. Specifically, we choose the following three popular balancing methods for our empirical study:

\textbf{Adversarial Gradient Reversal (AGR)} \cite{Bica2020Estimating}. It aims to build a treatment classifier $G_A$ taking the representation $\boldsymbol{h}_t$ as input, and maximize the following classification loss to obtain the representation $\boldsymbol{h}_t$ that is invariant to the treatment assignment:
\begin{equation}
    \mathcal{L}_{G_A}(\theta_A,\theta_R) = -\sum_{j=1}^{d_a}\mathbb{I}_{(\boldsymbol{A}_t=a_j)}logG_A(\boldsymbol{h}_t;\theta_A),
\end{equation}
where $\theta_R$ denotes the parameters for generating the representation $\boldsymbol{h}_t$,  $\theta_A$ denotes the parameter of the treatment classifier $G_A$, $\mathbb{I}$ 
is the indicator function. The adversarial loss aims to make the inferred representation $\boldsymbol{h}_t$ is not predictive to the treatment assignment.

\textbf{Counterfactual Domain Confusion (CDC)} \cite{melnychuk2022causal}. The CDC balancing method is designed to ensure these representations are non-predictive of the current treatment assignment. First, the CDC method is also developed to fit the treatment classifier network $G_A$ using the representation $\boldsymbol{h}_t$ by minimizing classification loss:
\begin{equation}
    \mathcal{L}_{G_A}(\theta_A,\theta_R) = -\sum_{j=1}^{d_a}\mathbb{I}_{(\boldsymbol{A}_t=a_j)}logG_A(\boldsymbol{h}_t;\theta_A),
\end{equation}
then CDC method proposes to minimize the cross-entropy between a uniform distribution over treatment categorical space and predictions of $G_A$ via the following objective:
\begin{equation}
    \mathcal{L}_{conf}(\theta_A,\theta_R) = -\sum_{j=1}^{d_a}\frac{1}{d_a}logG_A(\boldsymbol{h}_t;\theta_A),
\end{equation}
\textbf{PS-based Contrastive Balancing (PCB)}  \cite{li2022contrastive}. The balancing method utilizes a contrastive learning paradigm to infer the balanced representation based on the propensity score, which is a two-stage procedure. First, we pre-train a propensity score estimator $e(\cdot)$ supervised by the treatment and get the propensity score for every instance. 

Then we can build the positive and negative sample sets $\boldsymbol{X}^+$, $\boldsymbol{X}^-$ based on the above calculated propensity scores, where $\boldsymbol{X}^+$ are identified by propensity scores near 
0.5, suggesting a higher likelihood of random treatment assignment, and $\boldsymbol{X}^-$ are characterized by propensity scores closer to 0 or 1, indicating a more pronounced inclination towards a specific treatment. Then, for every anchor sample $i$, the contrastive loss for balancing the representation is as follows:
 \begin{equation}
     \mathcal{L}_c = \sum_{i=1}^N -log\frac{exp(\boldsymbol{h}_i \cdot \boldsymbol{h}_k^+/\tau)}{\sum_{j=1}^Kexp(\boldsymbol{h}_i \cdot \boldsymbol{h}_j^-/\tau)},
 \end{equation}
where $\boldsymbol{h}_k^+$ is the representation of the randomly selected one positive sample from $\boldsymbol{X}^+$, $\boldsymbol{h}_j^-$ is the representation for every negative sample from $\boldsymbol{X}^-$, $\tau$ determines how much the contrastive loss inclines to the hard negative samples. 
Minimizing the loss can force the representation of units similar to that of positive samples while different from that of negative ones.

\section{Empirical Study}
\subsection{Experiment Settings}

\subsubsection{Baselines and Metric}
Although there are many proposed sequential models for temporal counterfactual estimation, we focus on several state-of-the-art models for examination, including (1) Causal Transformer (CT) \cite{melnychuk2022causal}: A transformer-based counterfactual outcome prediction model with domain confusion module (CDC) for balanced representations; (2) Counterfactual Recurrent Network(CRN) \cite{Bica2020Estimating}: a sequence-to-sequence model with adversarial gradient reversal (AGR) for balancing; (3) Recurrent Marginal Structural Networks(RMSN) \cite{lim2018forecasting}: an LSTM-based model with propensity reweighting to adjust for time-dependent confounders; (4) G-Net \cite{li2021g}: a sequential deep learning model based on G-computation; and (5) Marginal Structural Model (MSM) \cite{robins2000marginal}: a linear marginal structural model based on the inverse-probability-of-treatment weighted estimator. 

To ensure a thorough evaluation, we also explore the impact of the PCB balancing strategy, which is not inherently temporal, on both CT and CRN models. We refer to the balanced representation module as BRM and designate models without BRM as ERM (Empirical Risk Minimization) for clarity in our discussion. We evaluate these models by the Root Mean Square Error (RMSE) (i.e., lower is better).

\subsubsection{Datasets}

\textbf{Pure synthetic dataset}. Synthetic datasets are generated through autoregressive iterations based on a predefined causal structure between variables, such as the data generation process in \cite{bica2020time}. Such data closely aligns with the causal assumptions formulated by practitioners, making it highly suitable for use in evaluating temporal counterfactual estimation.

\textbf{Tumor Growth Simulator} \cite{lim2018forecasting,Bica2020Estimating,melnychuk2022causal}: The simulator employs a currently popular and widely accepted biomedical model to simulate the temporal evolution of tumor volume, adhering to recognized practices in the field of biomedical research. The model consists of two types of binary treatment: (1) Radiotherapy when assigned to a patient has an immediate effect on the outcome of the next step; (2) Chemotherapy affects several future outcomes with certain exponentially decaying effects, and the tumor volume as outcome. 

\textbf{Semi-synthetic MIMIC III} \cite{johnson2016mimic}: This dataset underwent standardized preprocessing procedures specifically designed for MIMIC-III data. It comprises ICU data aggregated at hourly intervals. The covariates within this dataset encompass 25 vital signs (dynamic features over time) and 3 static attributes, namely gender, ethnicity, and age. For additional information regarding to the data generation process, please refer to the Appendix of \cite{melnychuk2022causal}.

\textbf{M5 dataset} \cite{makridakis2022m5}. This dataset, provided by Walmart, captures the unit sales data of a diverse range of products sold across the United States, structured as grouped time series. It encompasses sales figures for 3,049 distinct products, which are segmented into three main categories: Hobbies, Foods, and Household. The original dataset does not have any label information on counterfactual outcomes. Therefore, we resort to  factual evaluation and discuss the results in the Appendix due to space limit.

\subsubsection{Evaluation}
In temporal counterfactual estimation, we evaluate the models in a variety of scenarios to verify the effectiveness of different balancing strategies, including:

\textbf{Standard supervised learning}. The scenario involves using the full historical data, including previous covariates and treatment sequences, to predict future counterfactual outcomes over several time steps by using the standard train-test training procedure.

\begin{figure}[t]
    \centering
\includegraphics[width=3.2in]{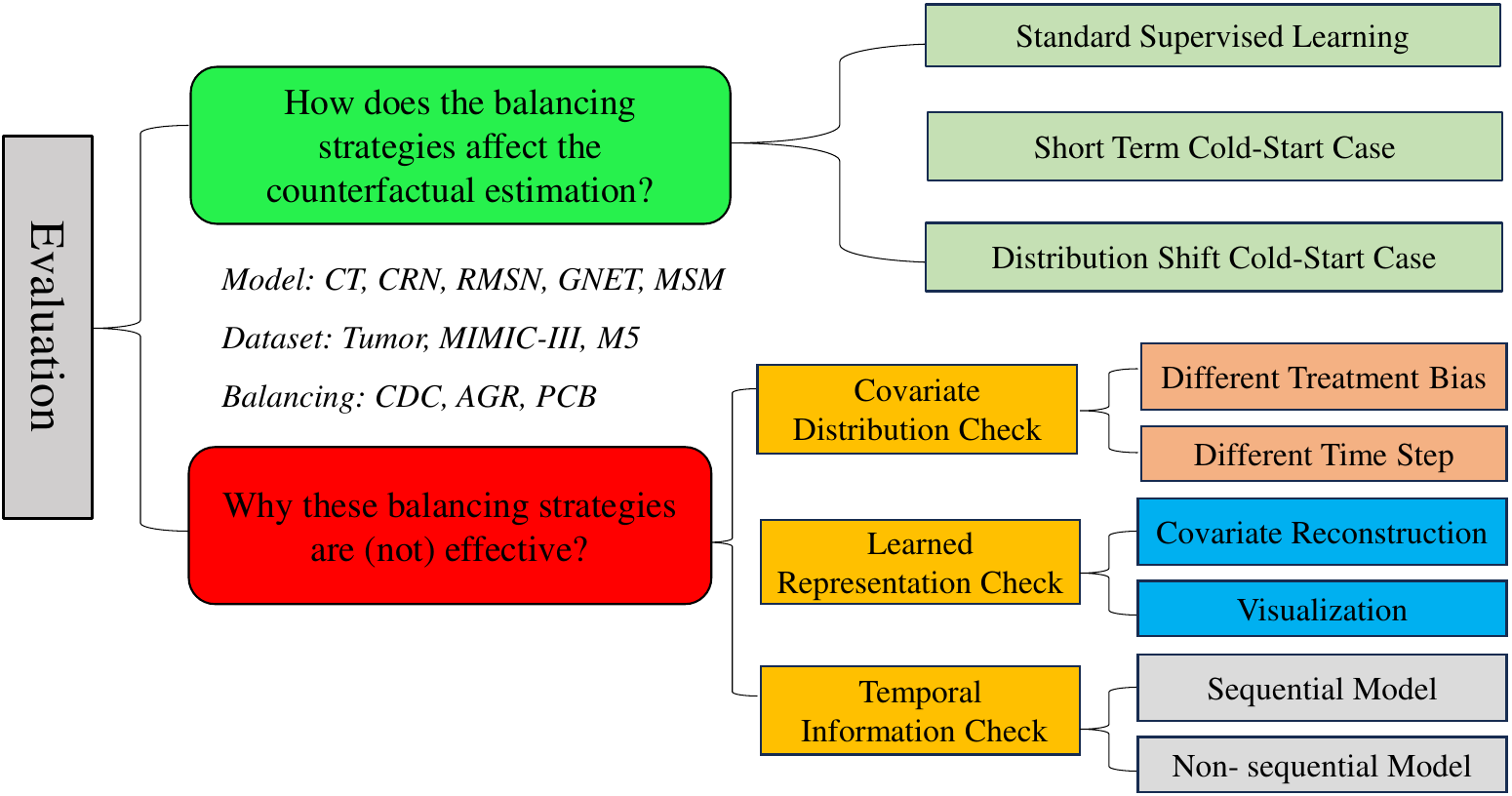}
\vspace{-3mm}
    \caption{Road map of the experimental evaluation in this study.}
    \label{road_map}
\vspace{-5mm}
\end{figure}

\textbf{Short-term history cold-start}. The scenario involves developing a short-term history test set for assessing the causal models. In this context, the model has access only to the covariates and treatment sequence information from $K$ steps before the predicted time step. This limitation in the extent of historical information available to the model is referred to as the truncation size.

\textbf{Distribution shift cold-start}. The scenario involves the case where there is distributional difference between the training data and the test data, particularly relevant in counterfactual estimation. It specifically pertains to the discrepancy in treatment bias between the data used during the training phase and that used in the testing phase.

\begin{table*}[t]
\caption{RMSE (mean $\pm$ std) with $\gamma=1,3,10$ on Tumor dataset.}
\label{standard_rmse_tumor}
\centering
\vskip 0.10in
\resizebox{0.9\textwidth}{!}{
\begin{tabular}{|c|c|l|l|l|l|l|}
\hline
\multirow{10}{*}{$\gamma=1$} & Methods     & \multicolumn{1}{c|}{$\tau$=1}        & \multicolumn{1}{c|}{2}               & \multicolumn{1}{c|}{3}               & \multicolumn{1}{c|}{4}               & \multicolumn{1}{c|}{5}               \\ \cline{2-7} 
                             & CT          & 0.808$\pm$0.064                      & 0.715$\pm$0.071                      & 0.754$\pm$0.067                      & 0.791$\pm$0.059                      & 0.832$\pm$0.059                      \\ \cline{2-7} 
                             & CT-PCB      & 0.901$\pm$0.056                      & 0.829$\pm$0.060                      & 0.898$\pm$0.078                      & 0.965$\pm$0.097                      & 1.033$\pm$0.117                      \\ \cline{2-7} 
                             & CT w/o BRM  & 0.794$\pm$0.068                      & 0.695$\pm$0.057                      & 0.734$\pm$0.056                      & 0.771$\pm$0.054                      & 0.808$\pm$0.051                      \\ \cline{2-7} 
                             & CRN         & 0.818$\pm$0.044                      & 0.727$\pm$0.046                      & 0.755$\pm$0.044                      & 0.785$\pm$0.041                      & 0.817$\pm$0.041                      \\ \cline{2-7} 
                             & CRN-PCB     & 1.071$\pm$0.103                      & 0.962$\pm$0.154                      & 0.961$\pm$0.144                      & 0.974$\pm$0.133                      & 0.992$\pm$0.120                      \\ \cline{2-7} 
                             & CRN w/o BRM & 0.845$\pm$0.062                      & 0.712$\pm$0.035                      & 0.736$\pm$0.031                      & 0.760$\pm$0.028                      & 0.787$\pm$0.030                      \\ \cline{2-7} 
                             & RMSN        & 1.116$\pm$0.116                      & 1.047$\pm$0.154                      & 1.072$\pm$0.096                      & 1.090$\pm$0.065                      & 1.113$\pm$0.050                      \\ \cline{2-7} 
                             & GNET        & 0.867$\pm$0.070                      & 0.982$\pm$0.079                      & 1.150$\pm$0.119                      & 1.219$\pm$0.146                      & 1.255$\pm$0.176                      \\ \cline{2-7} 
                             & MSM         & \multicolumn{1}{c|}{1.204$\pm$0.042} & \multicolumn{1}{c|}{1.693$\pm$0.123} & \multicolumn{1}{c|}{2.028$\pm$0.151} & \multicolumn{1}{c|}{2.227$\pm$0.164} & \multicolumn{1}{c|}{2.314$\pm$0.167} \\ \hline
\multirow{9}{*}{$\gamma=3$}  & CT          & 1.033$\pm$0.122                      & 0.827$\pm$0.097                      & 0.913$\pm$0.127                      & 0.984$\pm$0.135                      & 1.043$\pm$0.145                      \\ \cline{2-7} 
                             & CT-PCB      & 1.615$\pm$0.292                      & 1.750$\pm$0.553                      & 2.003$\pm$0.638                      & 2.178$\pm$0.686                      & 2.290$\pm$0.701                      \\ \cline{2-7} 
                             & CT w/o BRM  & 1.003$\pm$0.108                      & 0.785$\pm$0.084                      & 0.855$\pm$0.109                      & 0.916$\pm$0.128                      & 0.965$\pm$0.142                      \\ \cline{2-7} 
                             & CRN         & 1.070$\pm$0.108                      & 1.139$\pm$0.358                      & 1.380$\pm$0.518                      & 1.560$\pm$0.604                      & 1.690$\pm$0.641                      \\ \cline{2-7} 
                             & CRN-PCB     & 1.582$\pm$0.120                      & 1.426$\pm$0.259                      & 1.472$\pm$0.286                      & 1.549$\pm$0.322                      & 1.599$\pm$0.350                      \\ \cline{2-7} 
                             & CRN w/o BRM & 1.046$\pm$0.111                      & 0.795$\pm$0.056                      & 0.853$\pm$0.079                      & 0.907$\pm$0.105                      & 0.957$\pm$0.129                      \\ \cline{2-7} 
                             & RMSN        & 1.266$\pm$0.078                      & 1.154$\pm$0.132                      & 1.240$\pm$0.143                      & 1.290$\pm$0.166                      & 1.327$\pm$0.189                      \\ \cline{2-7} 
                             & GNET        & 1.334$\pm$0.299                      & 1.114$\pm$0.092                      & 1.243$\pm$0.095                      & 1.286$\pm$0.106                      & 1.311$\pm$0.116                      \\ \cline{2-7} 
                             & MSM         & \multicolumn{1}{c|}{1.749$\pm$0.139} & \multicolumn{1}{c|}{2.404$\pm$0.445} & \multicolumn{1}{c|}{2.742$\pm$0.516} & \multicolumn{1}{c|}{2.896$\pm$0.545} & \multicolumn{1}{c|}{2.924$\pm$0.547} \\ \hline
\multirow{9}{*}{$\gamma=10$} & CT          & \multicolumn{1}{c|}{5.746$\pm$1.709} & \multicolumn{1}{c|}{6.722$\pm$2.875} & \multicolumn{1}{c|}{7.326$\pm$2.989} & \multicolumn{1}{c|}{7.497$\pm$2.941} & \multicolumn{1}{c|}{7.493$\pm$2.829} \\ \cline{2-7} 
                             & CT-PCB      & 5.742$\pm$1.603                      & 6.387$\pm$2.988                      & 6.896$\pm$3.292                      & 7.015$\pm$3.433                      & 7.132$\pm$3.479                      \\ \cline{2-7} 
                             & CT w/o BRM  & \multicolumn{1}{c|}{4.395$\pm$0.958} & \multicolumn{1}{c|}{4.338$\pm$1.939} & \multicolumn{1}{c|}{4.840$\pm$1.997} & \multicolumn{1}{c|}{4.991$\pm$1.818} & \multicolumn{1}{c|}{5.080$\pm$1.736} \\ \cline{2-7} 
                             & CRN         & \multicolumn{1}{c|}{4.963$\pm$0.345} & \multicolumn{1}{c|}{6.526$\pm$2.435} & \multicolumn{1}{c|}{7.085$\pm$3.204} & \multicolumn{1}{c|}{7.289$\pm$3.384} & \multicolumn{1}{c|}{7.326$\pm$3.210} \\ \cline{2-7} 
                             & CRN-PCB     & 5.280$\pm$0.619                      & 8.982$\pm$2.904                      & 10.403$\pm$3.541                     & 10.617$\pm$3.603                     & 10.304$\pm$3.394                     \\ \cline{2-7} 
                             & CRN w/o BRM & \multicolumn{1}{c|}{4.707$\pm$0.394} & \multicolumn{1}{c|}{6.502$\pm$2.026} & \multicolumn{1}{c|}{7.342$\pm$2.174} & \multicolumn{1}{c|}{7.690$\pm$2.174} & \multicolumn{1}{c|}{7.727$\pm$2.093} \\ \cline{2-7} 
                             & RMSN        & \multicolumn{1}{c|}{5.109$\pm$0.400} & \multicolumn{1}{c|}{5.339$\pm$1.815} & \multicolumn{1}{c|}{5.479$\pm$1.895} & \multicolumn{1}{c|}{5.331$\pm$1.801} & \multicolumn{1}{c|}{5.125$\pm$1.621} \\ \cline{2-7} 
                             & GNET        & \multicolumn{1}{c|}{3.893$\pm$0.367} & \multicolumn{1}{c|}{4.010$\pm$1.307} & \multicolumn{1}{c|}{4.958$\pm$1.595} & \multicolumn{1}{c|}{5.375$\pm$1.725} & \multicolumn{1}{c|}{5.465$\pm$1.730} \\ \cline{2-7} 
                             & MSM         & \multicolumn{1}{c|}{5.837$\pm$0.616} & \multicolumn{1}{c|}{2.040$\pm$0.672} & \multicolumn{1}{c|}{3.039$\pm$0.999} & \multicolumn{1}{c|}{3.870$\pm$1.274} & \multicolumn{1}{c|}{4.617$\pm$1.525} \\ \hline
\end{tabular}
}
\end{table*}

\begin{table*}[th]
\centering
\caption{RMSE (mean $\pm$ std) on MIMIC-III dataset.}
\vskip 0.10in
\resizebox{0.75\textwidth}{!}{
\label{mimic_3_rmse}
\begin{tabular}{|c|l|l|l|l|l|l|}
\hline
Methods     & \multicolumn{1}{c|}{$\tau$=1} & \multicolumn{1}{c|}{2} & \multicolumn{1}{c|}{3} & \multicolumn{1}{c|}{4} & \multicolumn{1}{c|}{5} & \multicolumn{1}{c|}{6} \\ \hline
CT          & 0.25$\pm$0.06                 & 0.42$\pm$0.09          & 0.55$\pm$0.15          & 0.65$\pm$0.21          & 0.73$\pm$0.25          & 0.79$\pm$0.29          \\ \hline
CT-PCB      & 0.70$\pm$0.24                 & 1.15$\pm$0.67          & 1.30$\pm$0.70          & 1.41$\pm$0.74          & 1.50$\pm$0.78          & 1.56$\pm$0.81          \\ \hline
CT w/o BRM  & 0.23$\pm$0.06                 & 0.41$\pm$0.08          & 0.53$\pm$0.13          & 0.63$\pm$0.17          & 0.70$\pm$0.21          & 0.75$\pm$0.23          \\ \hline
CRN         & 0.24$\pm$0.03                 & 0.50$\pm$0.08          & 0.68$\pm$0.15          & 0.82$\pm$0.24          & 0.97$\pm$0.38          & 1.13$\pm$0.54          \\ \hline
CRN-PCB     & 0.30$\pm$0.06                 & 0.97$\pm$0.74          & 1.13$\pm$0.81          & 1.30$\pm$0.86          & 1.45$\pm$0.92          & 1.59$\pm$0.99          \\ \hline
CRN w/o BRM & 0.22$\pm$0.02                 & 0.47$\pm$0.13          & 0.63$\pm$0.20          & 0.77$\pm$0.29          & 0.89$\pm$0.39          & 1.01$\pm$0.49          \\ \hline
RMSN        & 0.29$\pm$0.08                 & 0.59$\pm$0.19          & 0.78$\pm$0.27          & 0.91$\pm$0.30          & 0.99$\pm$0.29          & 1.06$\pm$0.28          \\ \hline
GNET        & 0.40$\pm$0.12                 & 0.72$\pm$0.14          & 0.98$\pm$0.23          & 1.17$\pm$0.30          & 1.34$\pm$0.35          & 1.47$\pm$0.40          \\ \hline
\end{tabular}
}
\end{table*}

This empirical study is structured around three critical questions: (1) \emph{How do these temporal models perform for the counterfactual estimation?}
(2) \emph{Does the balanced representation module contribute to the temporal counterfactual estimation?}
(3) \emph{Why is (not) the balanced representation module effective for counterfactual estimation?} To enhance readability, the road map of the experimental evaluation for this examination study has been shown in Figure \ref{road_map}. 

\subsection{Counterfactual Estimation in Standard Supervised Learning}
\textbf{Results on synthetic Tumor dataset}.
Here, we report the performance of the counterfactual prediction on the Tumor Growth dataset. Specifically, let $\gamma$ control the magnitude of the time-dependent confounding bias, i.e., the extent of correlation between covariates and treatment. Let $\tau$ denote the projection horizon for multi-step ahead prediction. In this evaluation setting, we set the $\gamma= {1,3,10}$ to test the model performance in the varying magnitude of confounding bias, projection horizon to $\tau=5$, and we report the average RMSE and standard deviation (STD) for 5 runs, the values of these parameters in this setting (also the subsequent setting) are chosen following previous work \cite{melnychuk2022causal}. The results are shown in Table \ref{standard_rmse_tumor}. From this table, we can have the following observations:

    (1) When $\gamma$ is relatively small (e.g., 1 and 3), the performance of balanced (e.g., CT and CT-PCB) and ERM models is similar. However, when $\gamma$ becomes larger (e.g., 10), the performance of the non-balanced models is better than the balanced ones in most cases, and the PCB balancing module performs the worst.

    (2) With the increase of $\gamma$, the variance of the performance for the models with BRM increases significantly. That means, the existence of BRM will introduce the high variance, making the models unstable in the case where the confounding bias is large. 

    (3) The performance of GNET, RMSN and MSM is undesirable when the $\gamma$ is small. However, they perform better and are more stable methods when $\gamma$ gets large (e.g., 10).

Overall, we did not observe a positive contribution from the balanced representation module to counterfactual estimation, instead tending to make the model's predictions unstable.

\textbf{Experiments on semi-synthetic MIMIC III dataset}. We adopt similar settings as in Tumor dataset to evaluate the performance of these sequential models on the semi-synthetic MIMIC III dataset, as shown in Table \ref{mimic_3_rmse}. We omit the experiment results of the MSM model because the model cannot converge on this dataset. Similar to the Tumor dataset, the balanced representation module fails to improve the performance on the MIMIC III dataset.

\subsection{Counterfactual Estimation in Cold-start Case}
Experimental results in standard supervised learning contradict previous claims about the effectiveness and robustness of balanced representation. Cold-start cases pose significant challenges to counterfactual estimation, while the balanced representation module could help alleviate the issue. Thus we further examine the following evaluation scenarios:

\textbf{Short term history cold-start}. We created a short-term history test set to evaluate causal models' performance in cold start situations. We used a history of truncation size, assessing models' ability to predict counterfactual outcomes with limited historical data. We report the results for $\gamma=1,3,10$ in Table \ref{short_history_rmse_tumor}. It can seen that even under the cold start case, the models with balancing modules fail to help with counterfactual estimation.

\begin{table*}[t]
\vspace{-2mm}
\centering \caption{ \centering RMSE (mean $\pm$ std) with $\gamma=1,3,10$ for the cold start case of short-term history on Tumor dataset, lower is better.}
\vskip 0.10in
\label{short_history_rmse_tumor}
\centering
\resizebox{0.85\textwidth}{!}{
\begin{tabular}{|c|c|l|l|l|l|l|}
\hline
\multirow{8}{*}{$\gamma=1$}  & Methods     & \multicolumn{1}{c|}{$\tau$=1}        & \multicolumn{1}{c|}{2}               & \multicolumn{1}{c|}{3}               & \multicolumn{1}{c|}{4}                & \multicolumn{1}{c|}{5}                \\ \cline{2-7} 
                             & CT          & 0.827$\pm$0.052                      & 0.715$\pm$0.074                      & 0.770$\pm$0.077                      & 0.819$\pm$0.075                       & 0.869$\pm$0.075                       \\ \cline{2-7} 
                             & CT w/o BRM  & 0.822$\pm$0.055                      & 0.716$\pm$0.064                      & 0.770$\pm$0.062                      & 0.820$\pm$0.062                       & 0.873$\pm$0.066                       \\ \cline{2-7} 
                             & CRN         & 0.842$\pm$0.037                      & 0.752$\pm$0.049                      & 0.794$\pm$0.048                      & 0.836$\pm$0.048                       & 0.879$\pm$0.049                       \\ \cline{2-7} 
                             & CRN w/o BRM & 0.868$\pm$0.077                      & 0.733$\pm$0.055                      & 0.769$\pm$0.055                      & 0.805$\pm$0.055                       & 0.841$\pm$0.057                       \\ \cline{2-7} 
                             & RMSN        & 1.134$\pm$0.117                      & 1.031$\pm$0.143                      & 1.079$\pm$0.099                      & 1.121$\pm$0.084                       & 1.165$\pm$0.083                       \\ \cline{2-7} 
                             & GNET        & 0.899$\pm$0.053                      & 6.062$\pm$1.338                      & 7.162$\pm$1.425                      & 7.385$\pm$1.442                       & 7.180$\pm$1.423                       \\ \cline{2-7} 
                             & MSM         & \multicolumn{1}{c|}{1.242$\pm$0.050} & \multicolumn{1}{c|}{1.906$\pm$0.102} & \multicolumn{1}{c|}{2.286$\pm$0.122} & \multicolumn{1}{c|}{2.508$\pm$0.129}  & \multicolumn{1}{c|}{2.600$\pm$0.125}  \\ \hline
\multirow{7}{*}{$\gamma=3$}  & CT          & 1.080$\pm$0.103                      & 0.902$\pm$0.068                      & 1.045$\pm$0.093                      & 1.157$\pm$0.117                       & 1.257$\pm$0.137                       \\ \cline{2-7} 
                             & CT w/o BRM  & 1.082$\pm$0.105                      & 0.888$\pm$0.065                      & 1.034$\pm$0.069                      & 1.146$\pm$0.071                       & 1.241$\pm$0.071                       \\ \cline{2-7} 
                             & CRN         & 1.147$\pm$0.103                      & 1.330$\pm$0.274                      & 1.660$\pm$0.418                      & 1.900$\pm$0.477                       & 2.073$\pm$0.493                       \\ \cline{2-7} 
                             & CRN w/o BRM & 1.127$\pm$0.103                      & 0.923$\pm$0.116                      & 1.036$\pm$0.137                      & 1.138$\pm$0.156                       & 1.226$\pm$0.171                       \\ \cline{2-7} 
                             & RMSN        & 1.336$\pm$0.067                      & 1.230$\pm$0.177                      & 1.343$\pm$0.178                      & 1.414$\pm$0.190                       & 1.476$\pm$0.208                       \\ \cline{2-7} 
                             & GNET        & 1.394$\pm$0.287                      & 4.869$\pm$0.286                      & 6.032$\pm$0.519                      & 6.546$\pm$0.749                       & 6.746$\pm$0.947                       \\ \cline{2-7} 
                             & MSM         & \multicolumn{1}{c|}{1.866$\pm$0.101} & \multicolumn{1}{c|}{3.158$\pm$0.233} & \multicolumn{1}{c|}{3.613$\pm$0.263} & \multicolumn{1}{c|}{3.821$\pm$0.273}  & \multicolumn{1}{c|}{3.859$\pm$0.270}  \\ \hline
\multirow{7}{*}{$\gamma=10$} & CT          & \multicolumn{1}{c|}{5.385$\pm$1.876} & \multicolumn{1}{c|}{5.732$\pm$2.804} & \multicolumn{1}{c|}{7.312$\pm$3.635} & \multicolumn{1}{c|}{8.222$\pm$3.640}  & \multicolumn{1}{c|}{8.573$\pm$3.542}  \\ \cline{2-7} 
                             & CT w/o BRM  & \multicolumn{1}{c|}{4.234$\pm$0.828} & \multicolumn{1}{c|}{2.706$\pm$0.483} & \multicolumn{1}{c|}{3.959$\pm$0.510} & \multicolumn{1}{c|}{4.112$\pm$0.456}  & \multicolumn{1}{c|}{4.873$\pm$0.772}  \\ \cline{2-7} 
                             & CRN         & \multicolumn{1}{c|}{5.057$\pm$0.474} & \multicolumn{1}{c|}{8.377$\pm$1.832} & \multicolumn{1}{c|}{9.189$\pm$3.031} & \multicolumn{1}{c|}{9.527$\pm$3.497}  & \multicolumn{1}{c|}{9.619$\pm$3.488}  \\ \cline{2-7} 
                             & CRN w/o BRM & \multicolumn{1}{c|}{4.798$\pm$0.508} & \multicolumn{1}{c|}{8.620$\pm$1.676} & \multicolumn{1}{c|}{9.737$\pm$2.008} & \multicolumn{1}{c|}{10.177$\pm$2.083} & \multicolumn{1}{c|}{10.199$\pm$1.986} \\ \cline{2-7} 
                             & RMSN        & \multicolumn{1}{c|}{5.313$\pm$0.562} & \multicolumn{1}{c|}{7.433$\pm$1.207} & \multicolumn{1}{c|}{7.429$\pm$1.171} & \multicolumn{1}{c|}{7.116$\pm$1.060}  & \multicolumn{1}{c|}{6.737$\pm$0.913}  \\ \cline{2-7} 
                             & GNET        & \multicolumn{1}{c|}{3.854$\pm$0.370} & \multicolumn{1}{c|}{6.614$\pm$0.892} & \multicolumn{1}{c|}{7.774$\pm$1.029} & \multicolumn{1}{c|}{8.222$\pm$1.069}  & \multicolumn{1}{c|}{8.264$\pm$1.054}  \\ \cline{2-7} 
                             & MSM         & \multicolumn{1}{c|}{5.980$\pm$0.742} & \multicolumn{1}{c|}{2.663$\pm$0.465} & \multicolumn{1}{c|}{3.962$\pm$0.684} & \multicolumn{1}{c|}{5.042$\pm$0.864}  & \multicolumn{1}{c|}{6.016$\pm$1.029}  \\ \hline
\end{tabular}
}
\end{table*}

\textbf{Distribution shift cold-start}. In this case,  we trained models using data with significant confounding bias ($\gamma=10$) and tested their performance on data with varying bias levels ($\gamma={1,3,8}$). This approach aims to determine if balancing modules enhance counterfactual prediction when there's a distribution shift between the source and target domains, as those balancing modules are designed to balance representations across different groups. Results in Table \ref{distribution_shift_rmse_tumor} show that similar to the short-term history and standard supervised learning cases, there is little difference between balanced models and non-balanced models with low $\gamma$ (e.g., 1). Meanwhile, the balancing module introduces high variance in cases of significant treatment bias (e.g., $\gamma=8$).

\begin{table*}[t]
\caption{RMSE (mean $\pm$ std) with $\gamma=1,3,8$ for the cold start case of distribution shift on Tumor dataset, lower is better.}
\vskip 0.10in
\centering
\label{distribution_shift_rmse_tumor}
\resizebox{0.8\textwidth}{!}{
\begin{tabular}{|c|c|l|l|l|l|l|}
\hline
\multirow{8}{*}{$\gamma=1$} & Methods     & \multicolumn{1}{c|}{$\tau$=1}        & \multicolumn{1}{c|}{2}               & \multicolumn{1}{c|}{3}               & \multicolumn{1}{c|}{4}               & \multicolumn{1}{c|}{5}               \\ \cline{2-7} 
                            & CT          & 1.654$\pm$1.190                      & 2.491$\pm$2.630                      & 3.030$\pm$3.469                      & 3.452$\pm$4.001                      & 3.595$\pm$4.049                      \\ \cline{2-7} 
                            & CT w/o BRM  & 0.941$\pm$0.077                      & 0.991$\pm$0.084                      & 1.076$\pm$0.105                      & 1.177$\pm$0.129                      & 1.263$\pm$0.138                      \\ \cline{2-7} 
                            & CRN         & 1.566$\pm$0.069                      & 2.065$\pm$0.146                      & 2.067$\pm$0.322                      & 2.095$\pm$0.376                      & 2.127$\pm$0.390                      \\ \cline{2-7} 
                            & CRN w/o BRM & 1.339$\pm$0.079                      & 1.308$\pm$0.068                      & 1.466$\pm$0.098                      & 1.575$\pm$0.103                      & 1.631$\pm$0.094                      \\ \cline{2-7} 
                            & RMSN        & 1.374$\pm$0.097                      & 1.188$\pm$0.072                      & 1.220$\pm$0.061                      & 1.273$\pm$0.074                      & 1.338$\pm$0.115                      \\ \cline{2-7} 
                            & GNET        & 1.074$\pm$0.050                      & 4.238$\pm$0.043                      & 4.884$\pm$0.117                      & 5.023$\pm$0.233                      & 5.050$\pm$0.369                      \\ \cline{2-7} 
                            & MSM         & \multicolumn{1}{c|}{1.160$\pm$0.033} & \multicolumn{1}{c|}{0.507$\pm$0.043} & \multicolumn{1}{c|}{0.763$\pm$0.063} & \multicolumn{1}{c|}{0.976$\pm$0.081} & \multicolumn{1}{c|}{1.167$\pm$0.096} \\ \hline
\multirow{7}{*}{$\gamma=3$} & CT          & 1.889$\pm$0.987                      & 2.600$\pm$2.463                      & 3.148$\pm$3.207                      & 3.554$\pm$3.676                      & 3.718$\pm$3.693                      \\ \cline{2-7} 
                            & CT w/o BRM  & 1.211$\pm$0.141                      & 1.138$\pm$0.148                      & 1.285$\pm$0.170                      & 1.405$\pm$0.200                      & 1.521$\pm$0.216                      \\ \cline{2-7} 
                            & CRN         & 1.810$\pm$0.098                      & 2.231$\pm$0.396                      & 2.375$\pm$0.668                      & 2.456$\pm$0.752                      & 2.506$\pm$0.740                      \\ \cline{2-7} 
                            & CRN w/o BRM & 1.619$\pm$0.103                      & 1.734$\pm$0.289                      & 1.971$\pm$0.309                      & 2.106$\pm$0.300                      & 2.162$\pm$0.287                      \\ \cline{2-7} 
                            & RMSN        & 1.704$\pm$0.117                      & 1.504$\pm$0.216                      & 1.548$\pm$0.224                      & 1.570$\pm$0.199                      & 1.595$\pm$0.169                      \\ \cline{2-7} 
                            & GNET        & 1.297$\pm$0.096                      & 3.909$\pm$0.075                      & 4.313$\pm$0.202                      & 4.372$\pm$0.335                      & 4.387$\pm$0.458                      \\ \cline{2-7} 
                            & MSM         & \multicolumn{1}{c|}{1.605$\pm$0.185} & \multicolumn{1}{c|}{0.617$\pm$0.104} & \multicolumn{1}{c|}{0.930$\pm$0.161} & \multicolumn{1}{c|}{1.193$\pm$0.207} & \multicolumn{1}{c|}{1.430$\pm$0.250} \\ \hline
\multirow{7}{*}{$\gamma=8$} & CT          & \multicolumn{1}{c|}{4.942$\pm$1.938} & \multicolumn{1}{c|}{5.723$\pm$3.801} & \multicolumn{1}{c|}{6.532$\pm$4.389} & \multicolumn{1}{c|}{6.937$\pm$4.525} & \multicolumn{1}{c|}{7.091$\pm$4.416} \\ \cline{2-7} 
                            & CT w/o BRM  & \multicolumn{1}{c|}{3.495$\pm$0.622} & \multicolumn{1}{c|}{3.287$\pm$1.403} & \multicolumn{1}{c|}{3.776$\pm$1.577} & \multicolumn{1}{c|}{3.925$\pm$1.445} & \multicolumn{1}{c|}{4.080$\pm$1.442} \\ \cline{2-7} 
                            & CRN         & \multicolumn{1}{c|}{4.124$\pm$0.235} & \multicolumn{1}{c|}{5.346$\pm$1.846} & \multicolumn{1}{c|}{5.782$\pm$2.433} & \multicolumn{1}{c|}{5.957$\pm$2.570} & \multicolumn{1}{c|}{6.019$\pm$2.427} \\ \cline{2-7} 
                            & CRN w/o BRM & \multicolumn{1}{c|}{3.871$\pm$0.281} & \multicolumn{1}{c|}{5.218$\pm$1.717} & \multicolumn{1}{c|}{5.932$\pm$1.849} & \multicolumn{1}{c|}{6.249$\pm$1.847} & \multicolumn{1}{c|}{6.311$\pm$1.775} \\ \cline{2-7} 
                            & RMSN        & \multicolumn{1}{c|}{4.163$\pm$0.255} & \multicolumn{1}{c|}{4.201$\pm$1.535} & \multicolumn{1}{c|}{4.326$\pm$1.581} & \multicolumn{1}{c|}{4.240$\pm$1.486} & \multicolumn{1}{c|}{4.123$\pm$1.324} \\ \cline{2-7} 
                            & GNET        & \multicolumn{1}{c|}{3.167$\pm$0.260} & \multicolumn{1}{c|}{4.643$\pm$0.749} & \multicolumn{1}{c|}{5.269$\pm$1.086} & \multicolumn{1}{c|}{5.539$\pm$1.289} & \multicolumn{1}{c|}{5.625$\pm$1.401} \\ \cline{2-7} 
                            & MSM         & \multicolumn{1}{c|}{4.788$\pm$0.438} & \multicolumn{1}{c|}{1.608$\pm$0.544} & \multicolumn{1}{c|}{2.403$\pm$0.813} & \multicolumn{1}{c|}{3.065$\pm$1.040} & \multicolumn{1}{c|}{3.659$\pm$1.249} \\ \hline
\end{tabular}
}
\end{table*}

\subsection{Why Balancing Module Does Not Work for the Counterfactual Estimation?}

The above study shows that using a balanced representation module in models does not help (and often worsens) in various scenarios. In this section, we analyze the covariate distribution among treatment groups and the models' learned representations, in order to offer insights into why the balancing module is ineffective.

\begin{figure}[t]
\vspace{-3mm}
\centering
\subfigure[$\gamma=0$]{
\begin{minipage}{0.25\linewidth}
\centering
\includegraphics[width=0.7in]{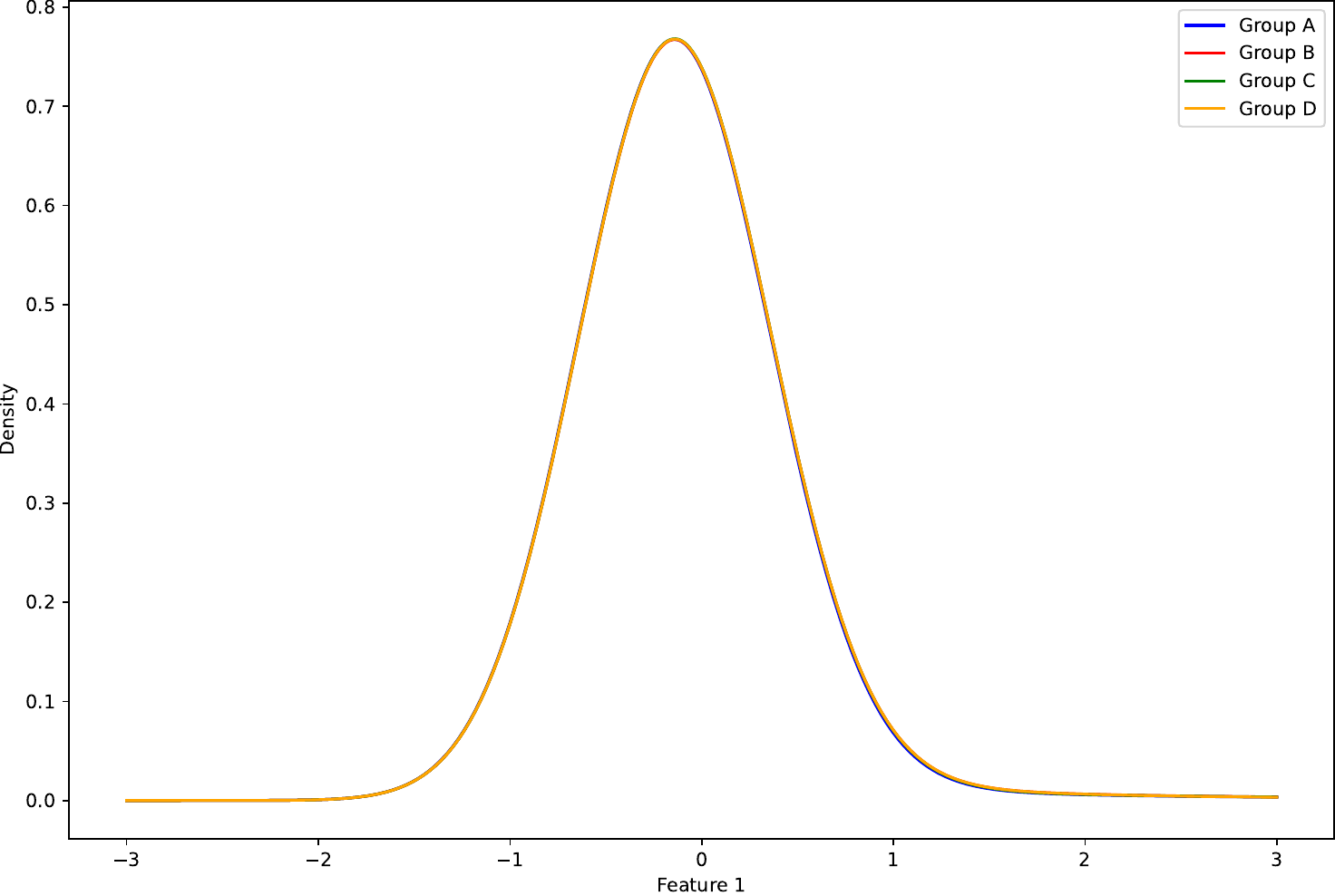}
\end{minipage}%
}%
\centering
\subfigure[$\gamma=1$]{
\begin{minipage}{0.25\linewidth}
\centering
\includegraphics[width=0.7in]{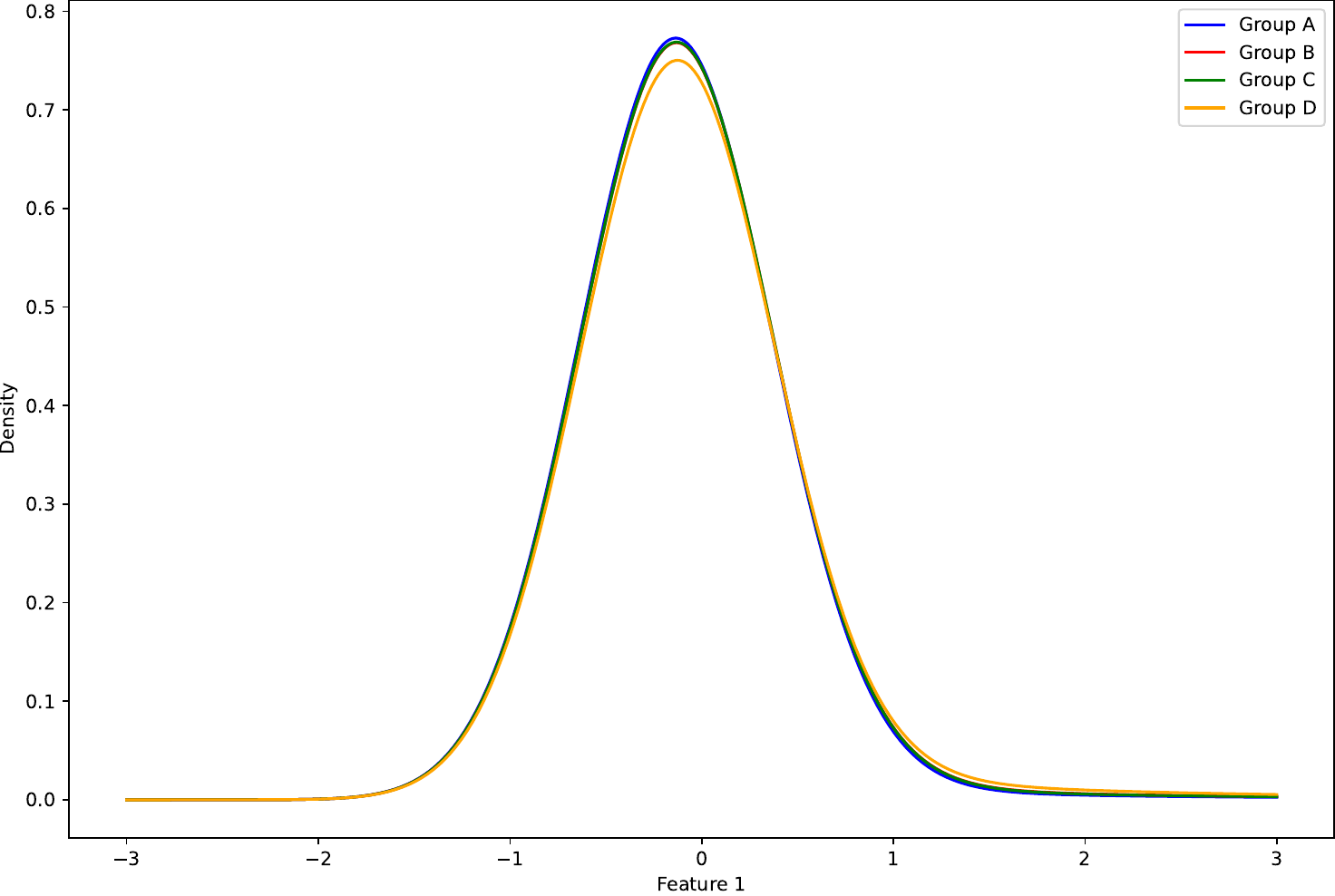}
\end{minipage}%
}%
\centering
\subfigure[$\gamma=3$]{
\begin{minipage}{0.25\linewidth}
\centering
\includegraphics[width=0.7in]{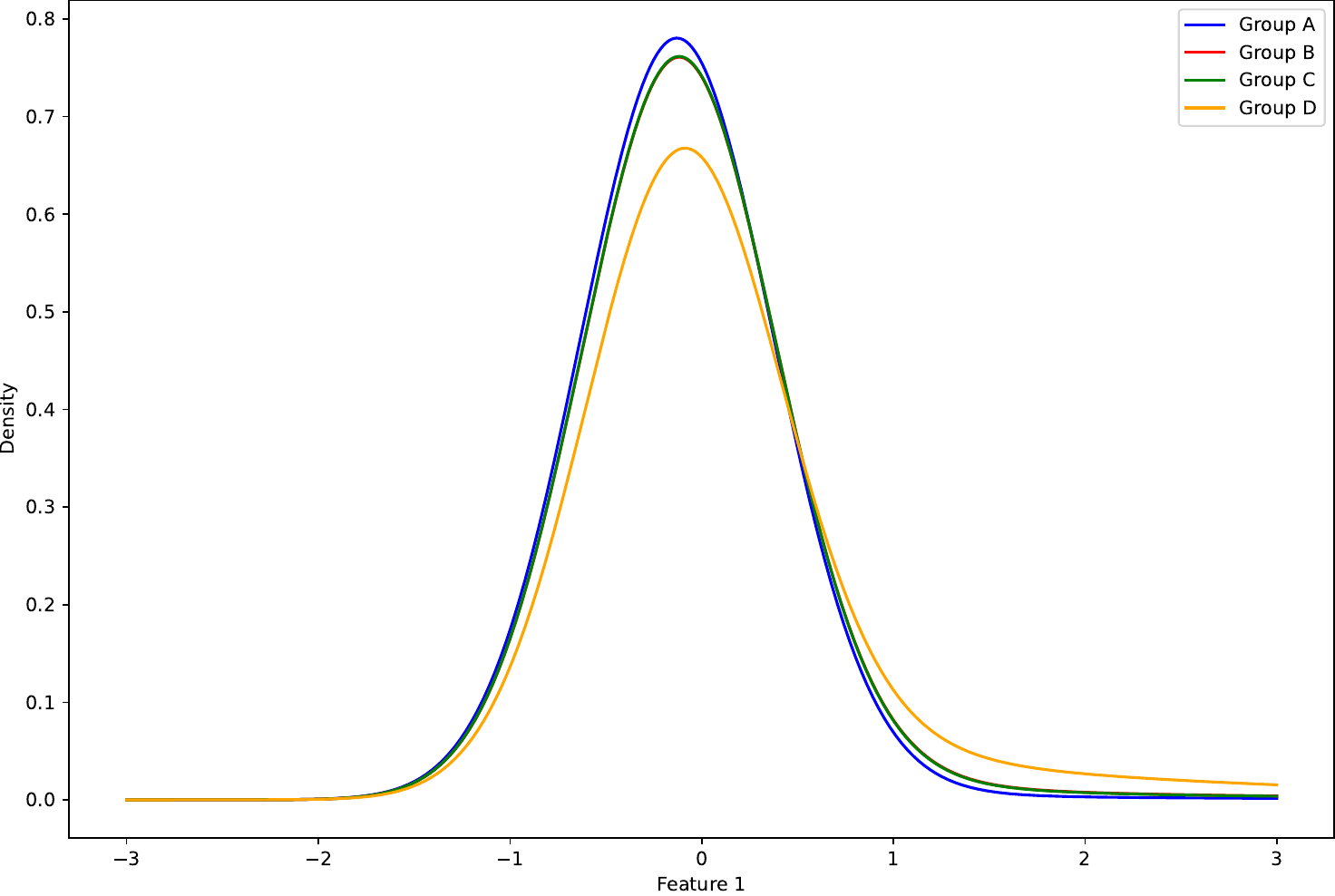}
\end{minipage}%
}%
\centering
\subfigure[$\gamma=10$]{
\begin{minipage}{0.25\linewidth}
\centering
\includegraphics[width=0.7in]{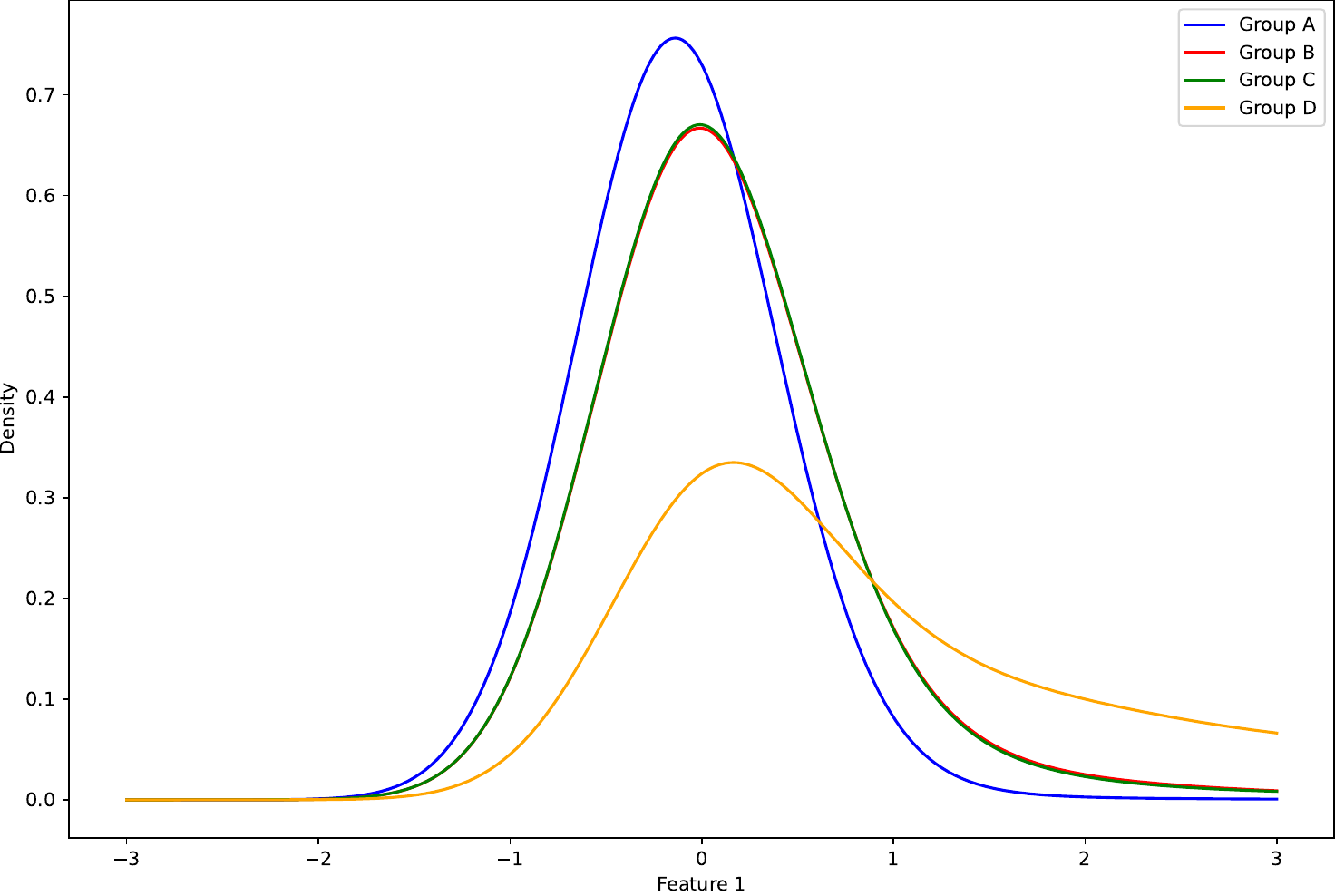}
\end{minipage}%
}%
\centering
\\
\subfigure[$\gamma=0$]{
\begin{minipage}{0.25\linewidth}
\centering
\includegraphics[width=0.7in]{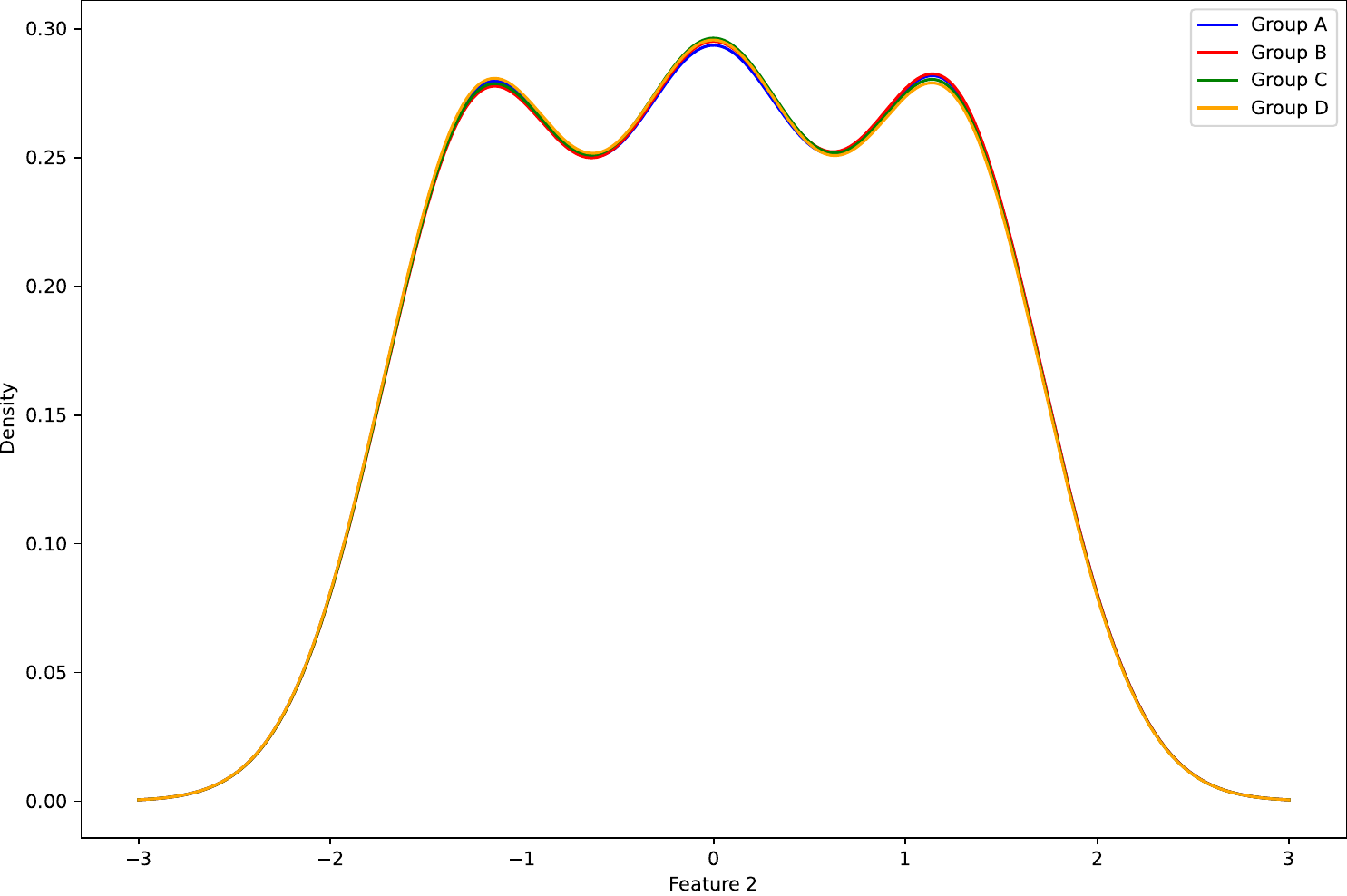}
\end{minipage}%
}%
\centering
\subfigure[$\gamma=1$]{
\begin{minipage}{0.25\linewidth}
\centering
\includegraphics[width=0.7in]{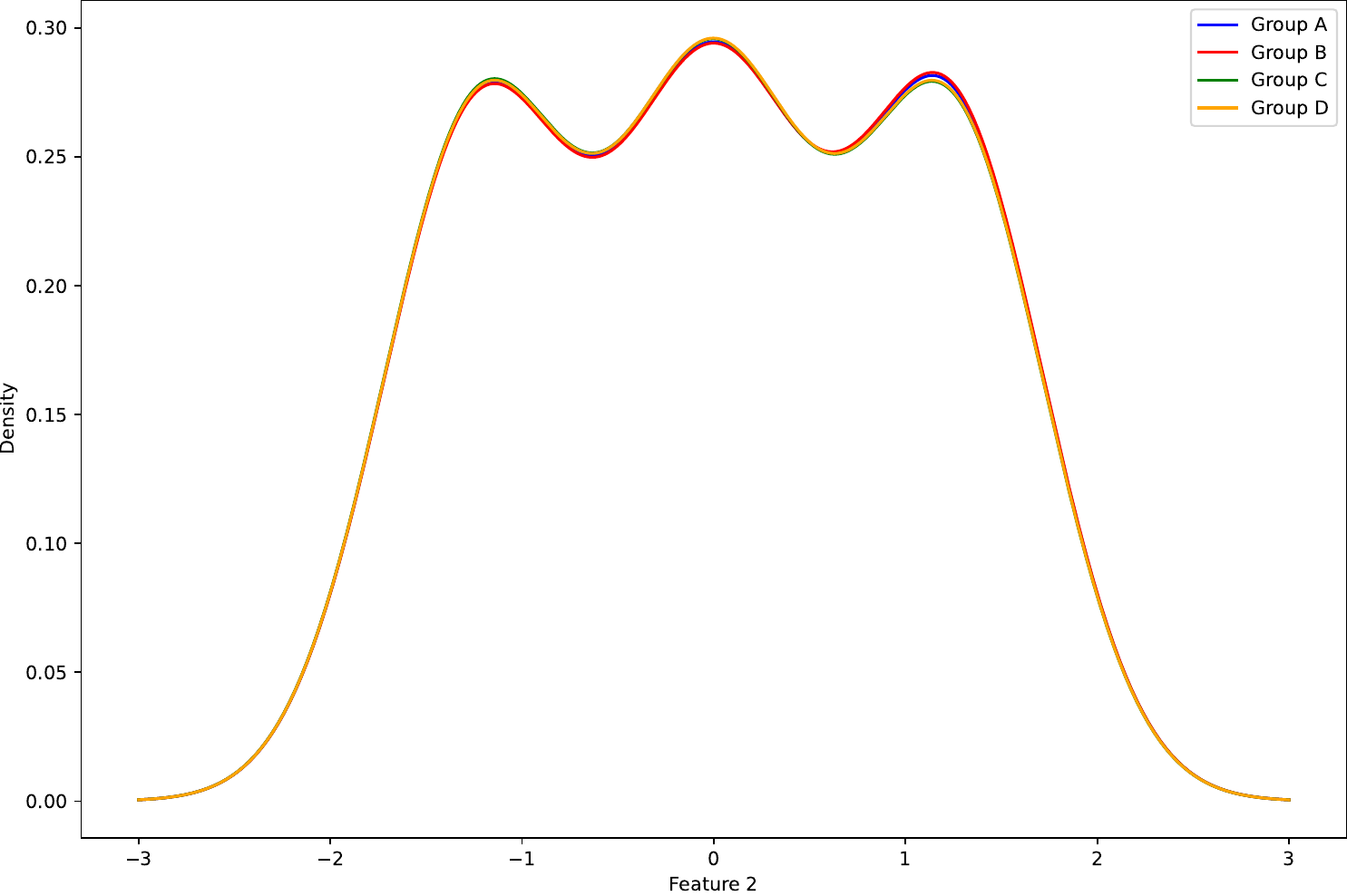}
\end{minipage}%
}%
\centering
\subfigure[$\gamma=3$]{
\begin{minipage}{0.25\linewidth}
\centering
\includegraphics[width=0.7in]{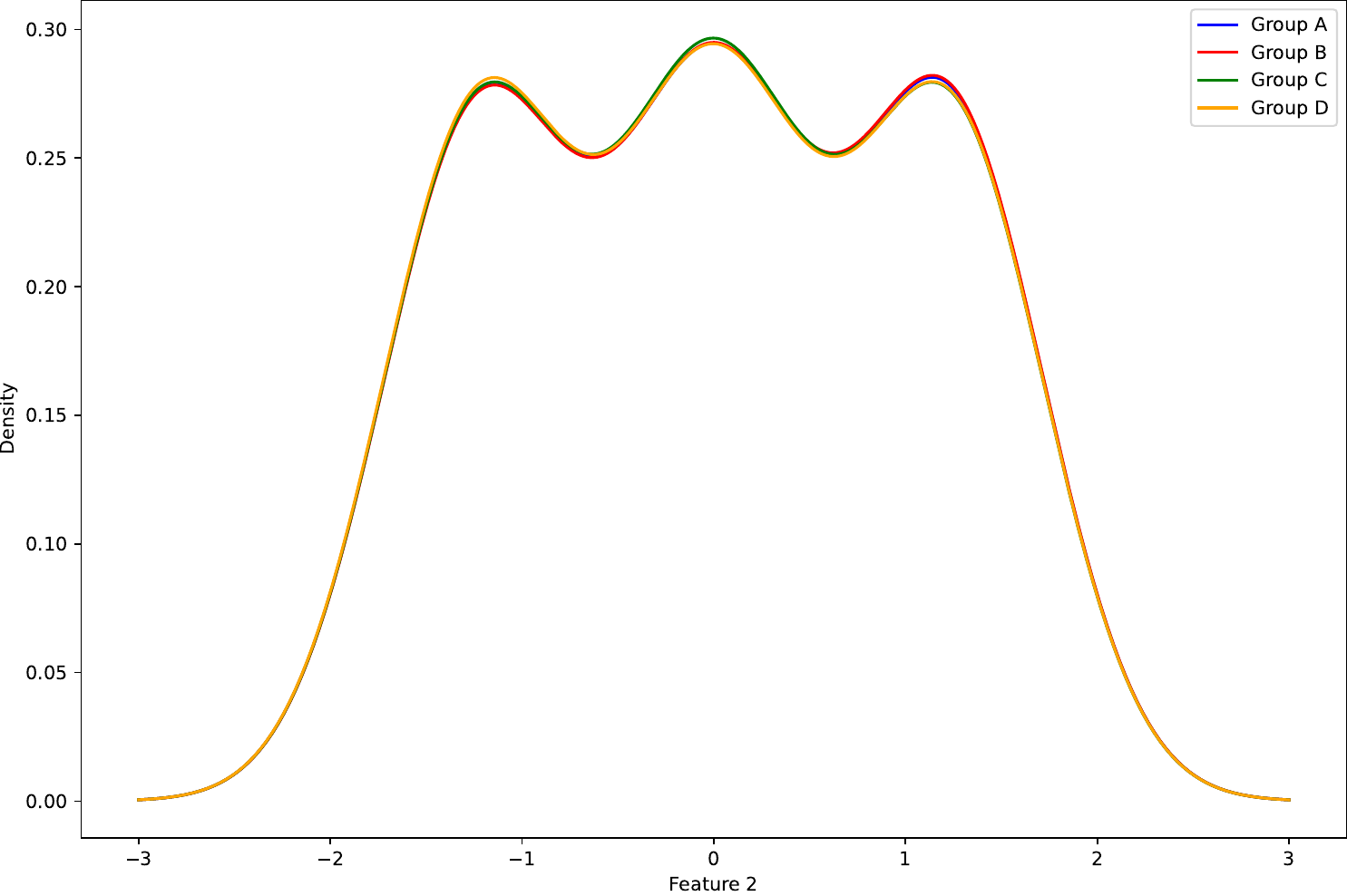}
\end{minipage}%
}%
\centering
\subfigure[$\gamma=10$]{
\begin{minipage}{0.25\linewidth}
\centering
\includegraphics[width=0.7in]{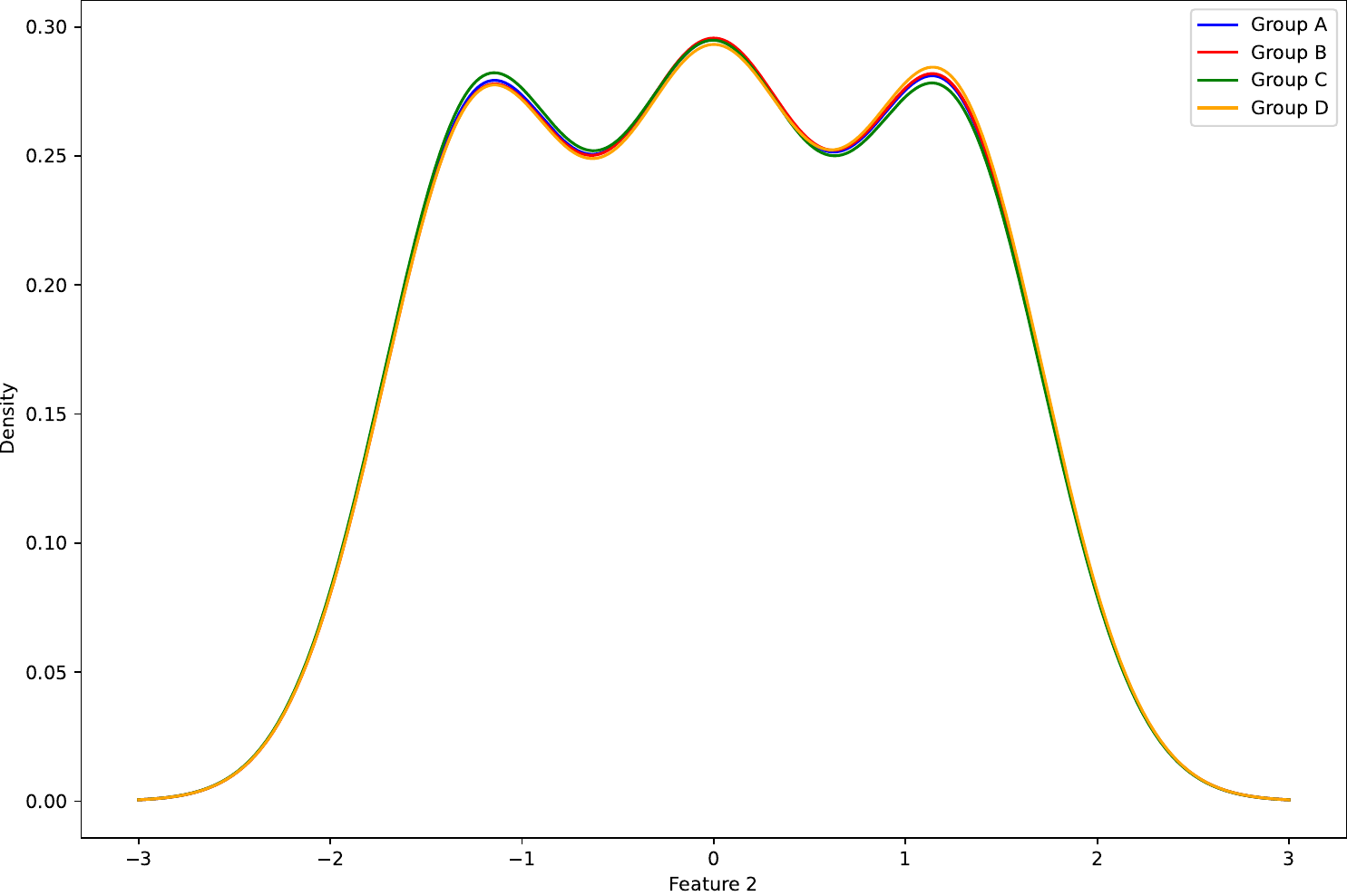}
\end{minipage}%
}%
\vspace{-.2cm}
\caption{Distribution shape for the two covariates under different strengths of treatment bias on Tumor. (a)-(f) for the first covariate, (g)-(l) for the second covariate.}
\label{fig_dis_cov}
\vspace{-5mm}
\end{figure}

\textbf{Checking on covariate distribution}. We investigate how covariate distributions vary among groups under different treatment bias levels. The balancing module aims to equalize these distributions for counterfactual estimation, but altering distributions might lower the model's accuracy for individual samples. If group differences in covariate distributions are not significant, the benefits of balancing may not outweigh its drawbacks. We examine the two covariates on this dataset averaging on history, using kernel density estimation (KDE) and a Gaussian kernel function to plot their distribution shapes for $\gamma={0,1,3,10}$. Figure  \ref{fig_dis_cov} shows the Gaussian fit distributions for two covariates. For the first covariate, significant distribution differences among groups only appear at higher treatment bias levels ($\gamma$ equals 0, 1, 3 show minimal differences). In contrast, the distribution of the second covariate among different groups is almost identical, regardless of treatment bias intensity.

To check if our observation is general, we generate time-series data using the universal method from \cite{bica2020time} and analyze the covariate distribution across treatment groups. This involves generating initial covariates and treatments using Gaussian and Bernoulli distributions, and then creating time-series under the proposed causal structure. The generated dataset features 10-dimensional covariates, a maximum sequence length of 30, and two-dimensional treatment variables, following a $p$-order autoregressive process, the detailed data generation can be found in Appendix.

\begin{figure}[t]
\vspace{-3mm}
\centering
\subfigure[$t=1$]{
\begin{minipage}{0.25\linewidth}
\centering
\includegraphics[width=0.7in]{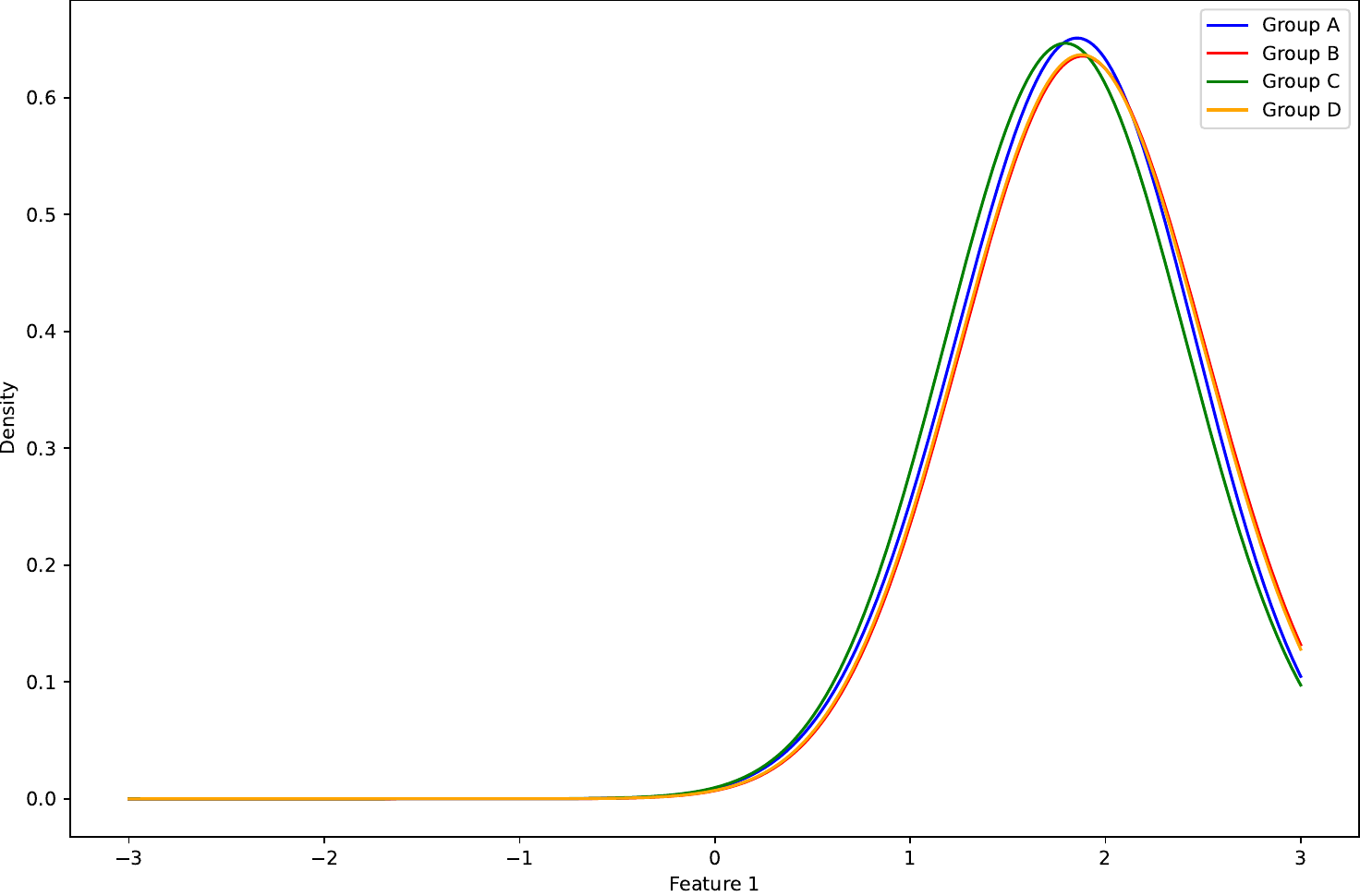}
\end{minipage}%
}%
\centering
\subfigure[$t=4$]{
\begin{minipage}{0.25\linewidth}
\centering
\includegraphics[width=0.7in]{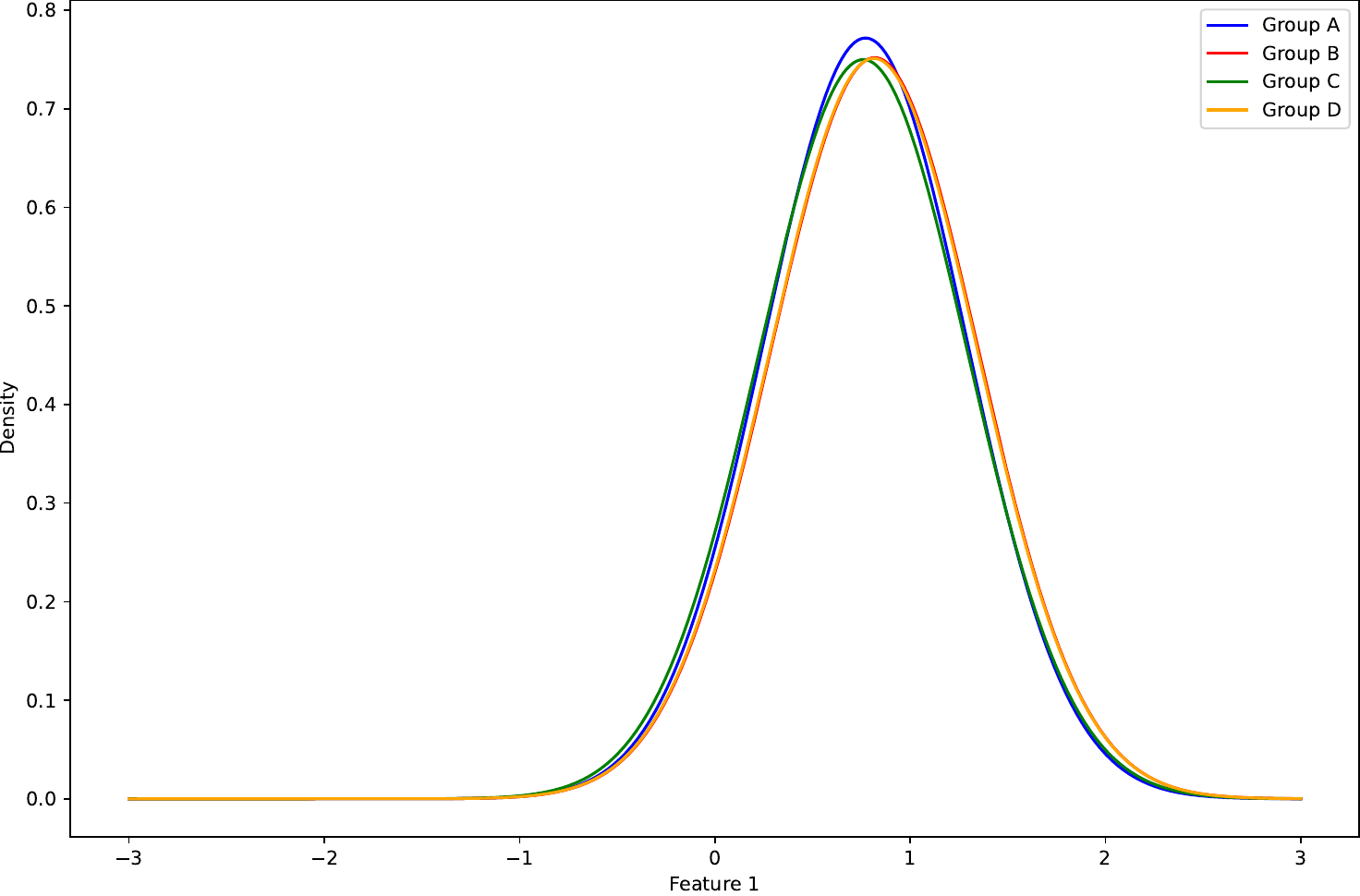}
\end{minipage}%
}%
\centering
\subfigure[$t=8$]{
\begin{minipage}{0.25\linewidth}
\centering
\includegraphics[width=0.7in]{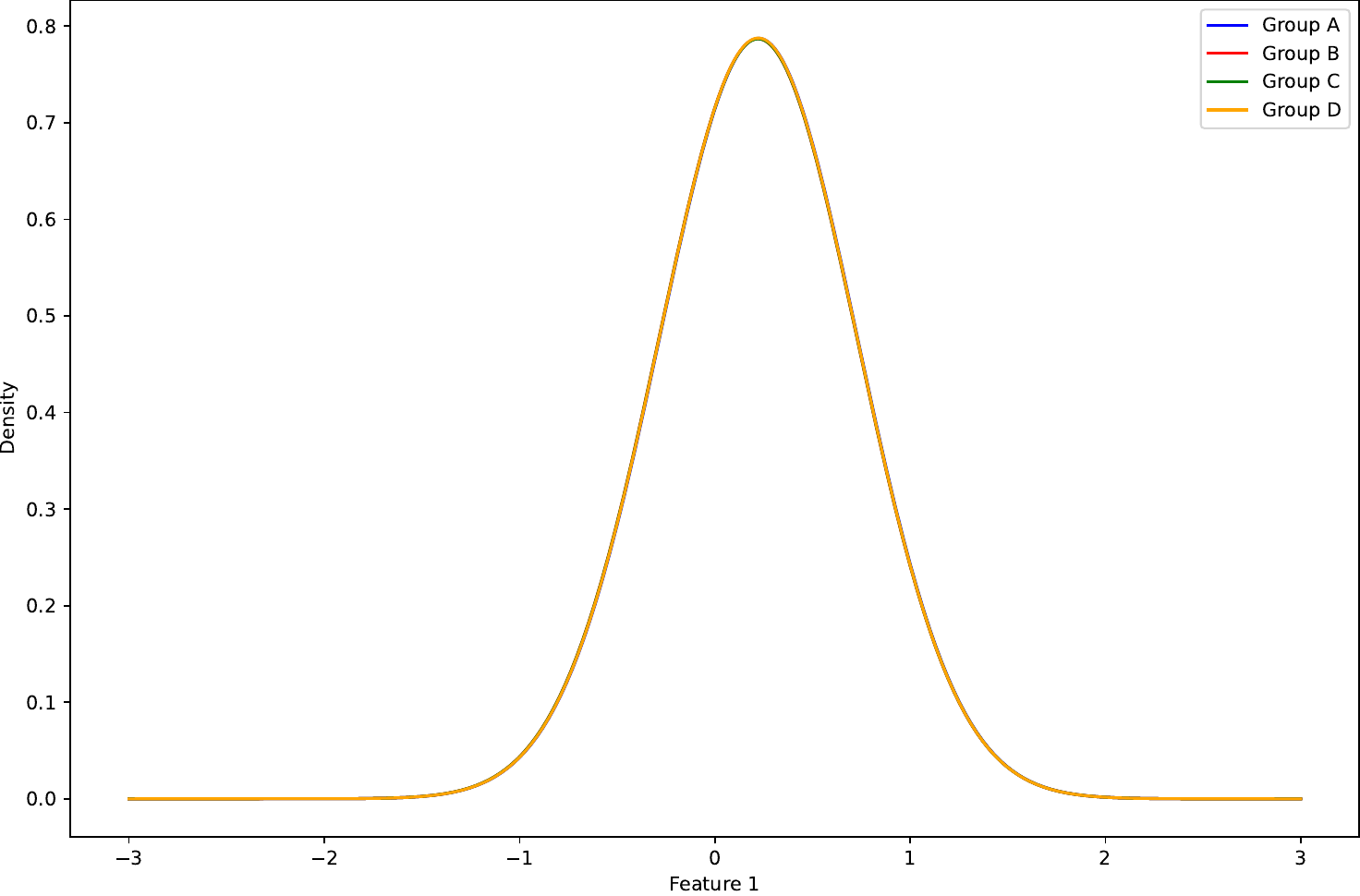}
\end{minipage}%
}%
\centering
\subfigure[$t=16$]{
\begin{minipage}{0.25\linewidth}
\centering
\includegraphics[width=0.7in]{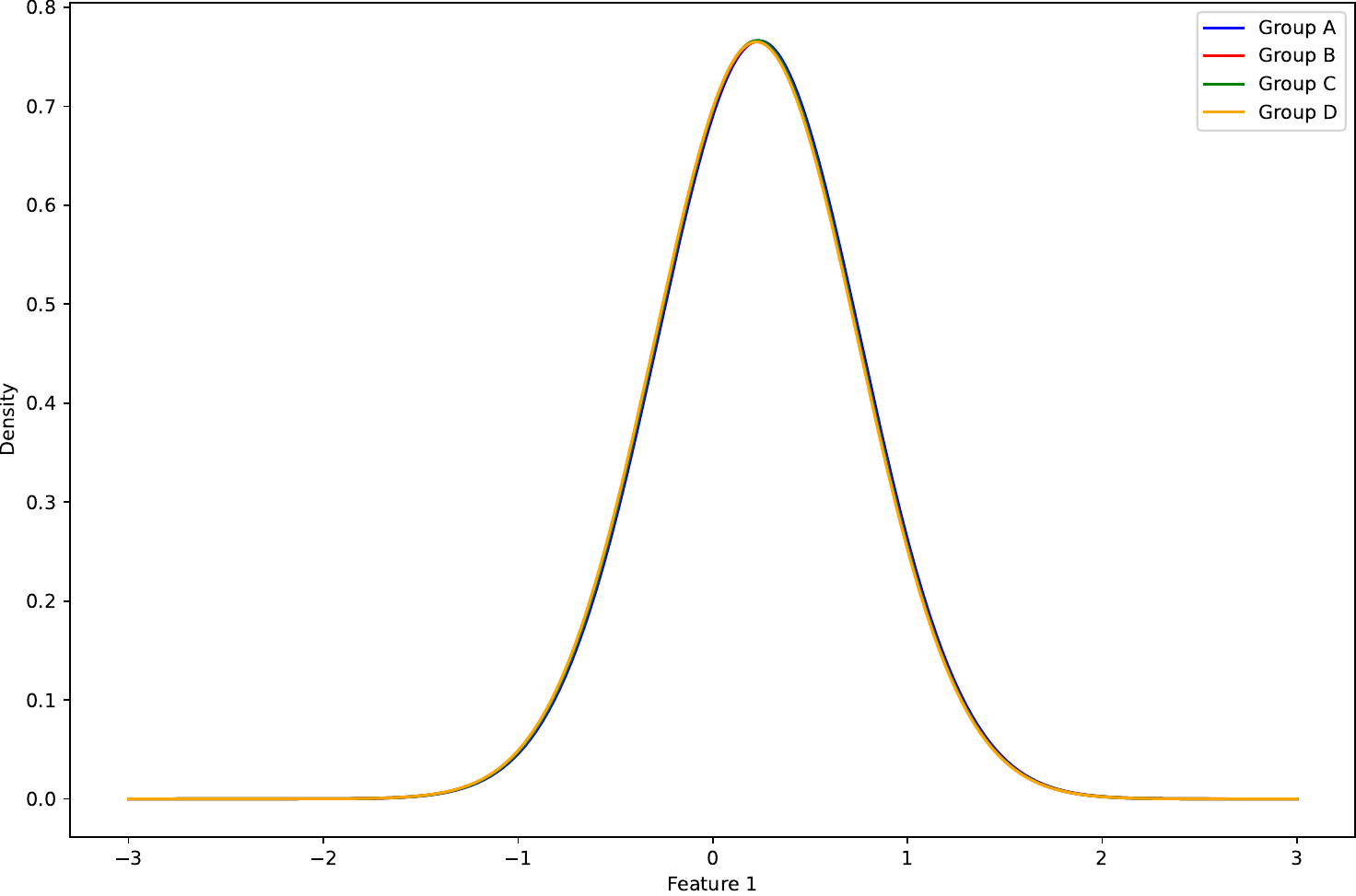}
\end{minipage}%
}%
\\
\subfigure[$t=1$]{
\begin{minipage}{0.25\linewidth}
\centering
\includegraphics[width=0.7in]{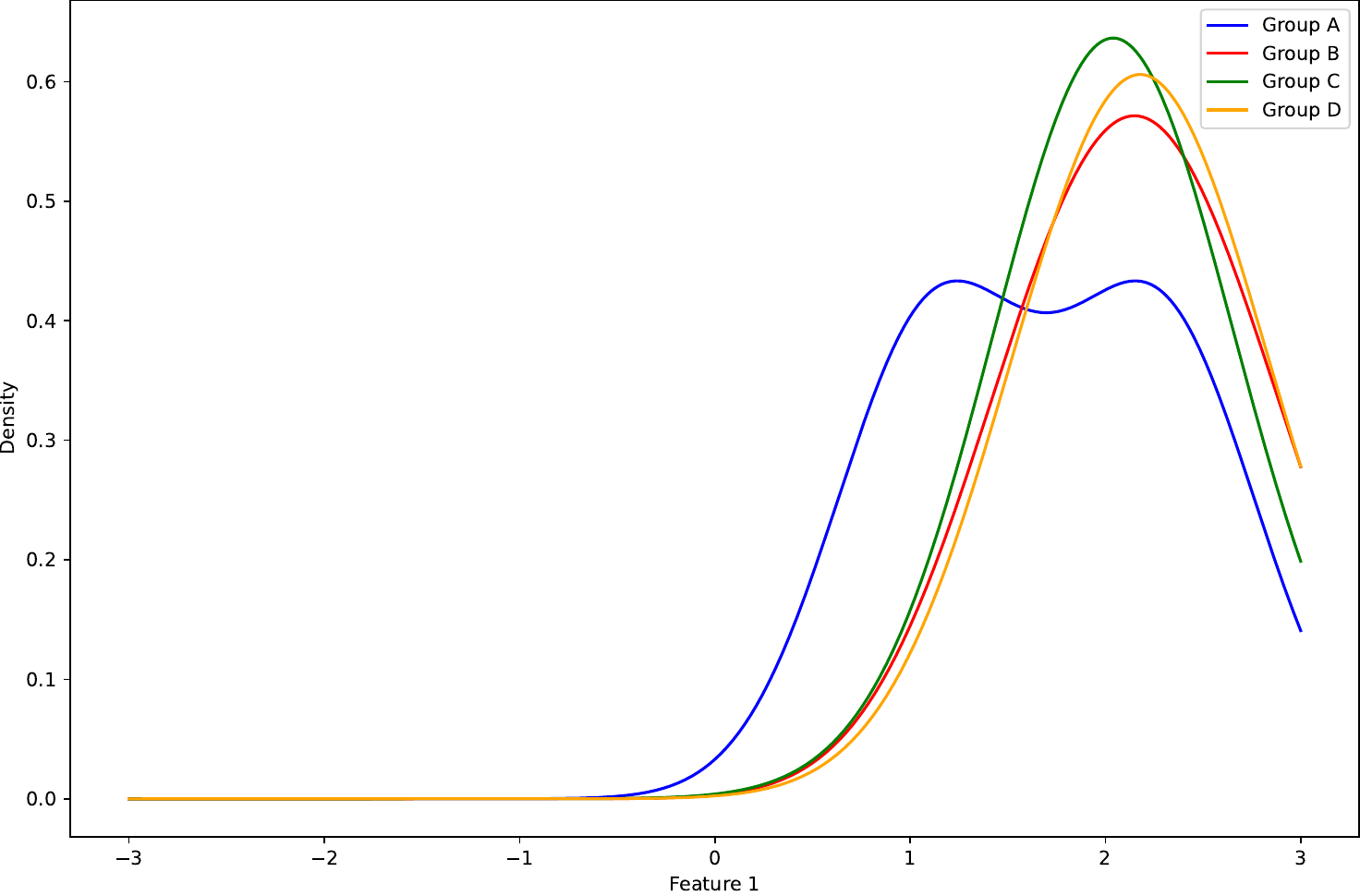}
\end{minipage}%
}%
\centering
\subfigure[$t=4$]{
\begin{minipage}{0.25\linewidth}
\centering
\includegraphics[width=0.7in]{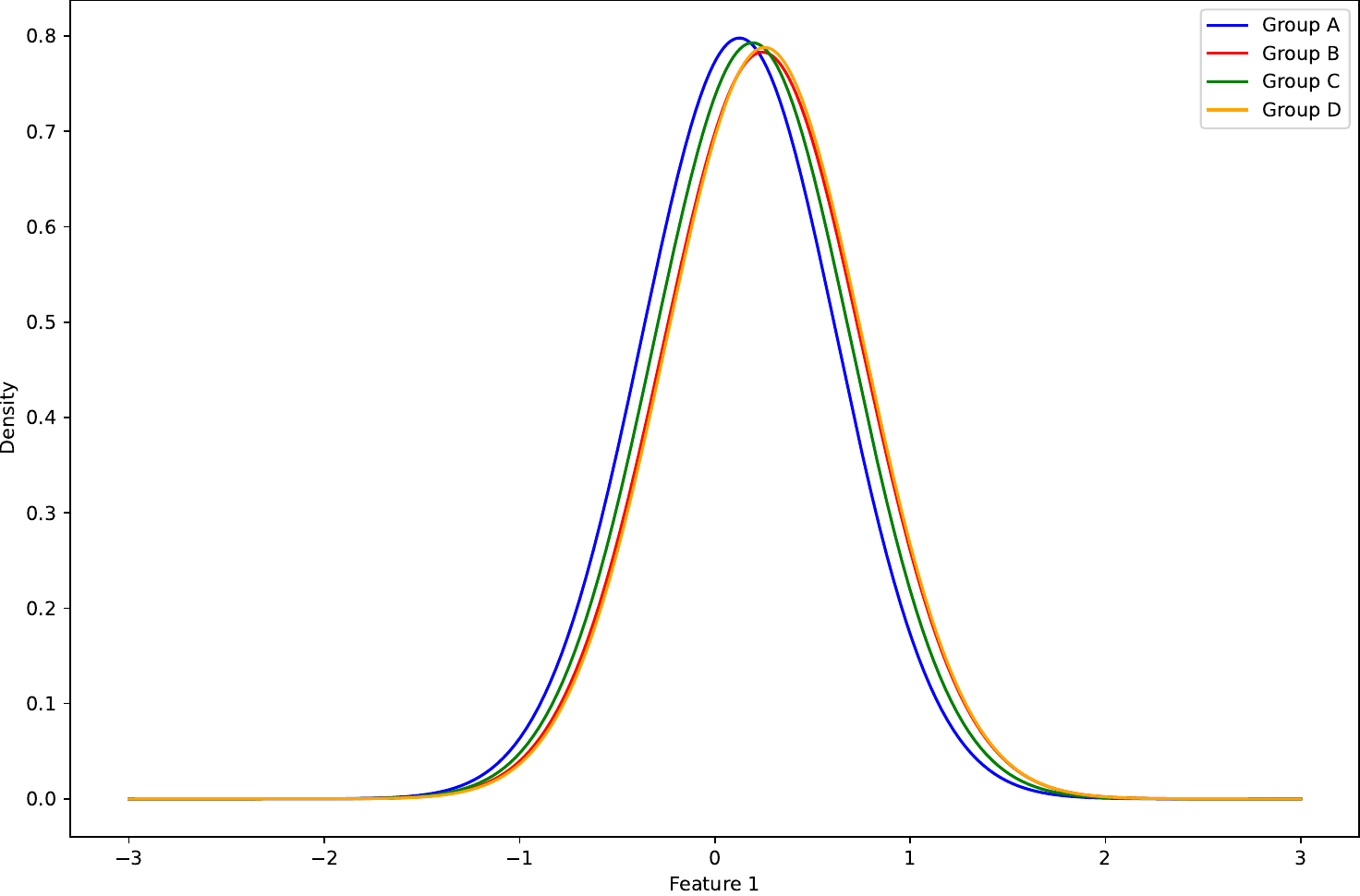}
\end{minipage}%
}%
\centering
\subfigure[$t=8$]{
\begin{minipage}{0.25\linewidth}
\centering
\includegraphics[width=0.7in]{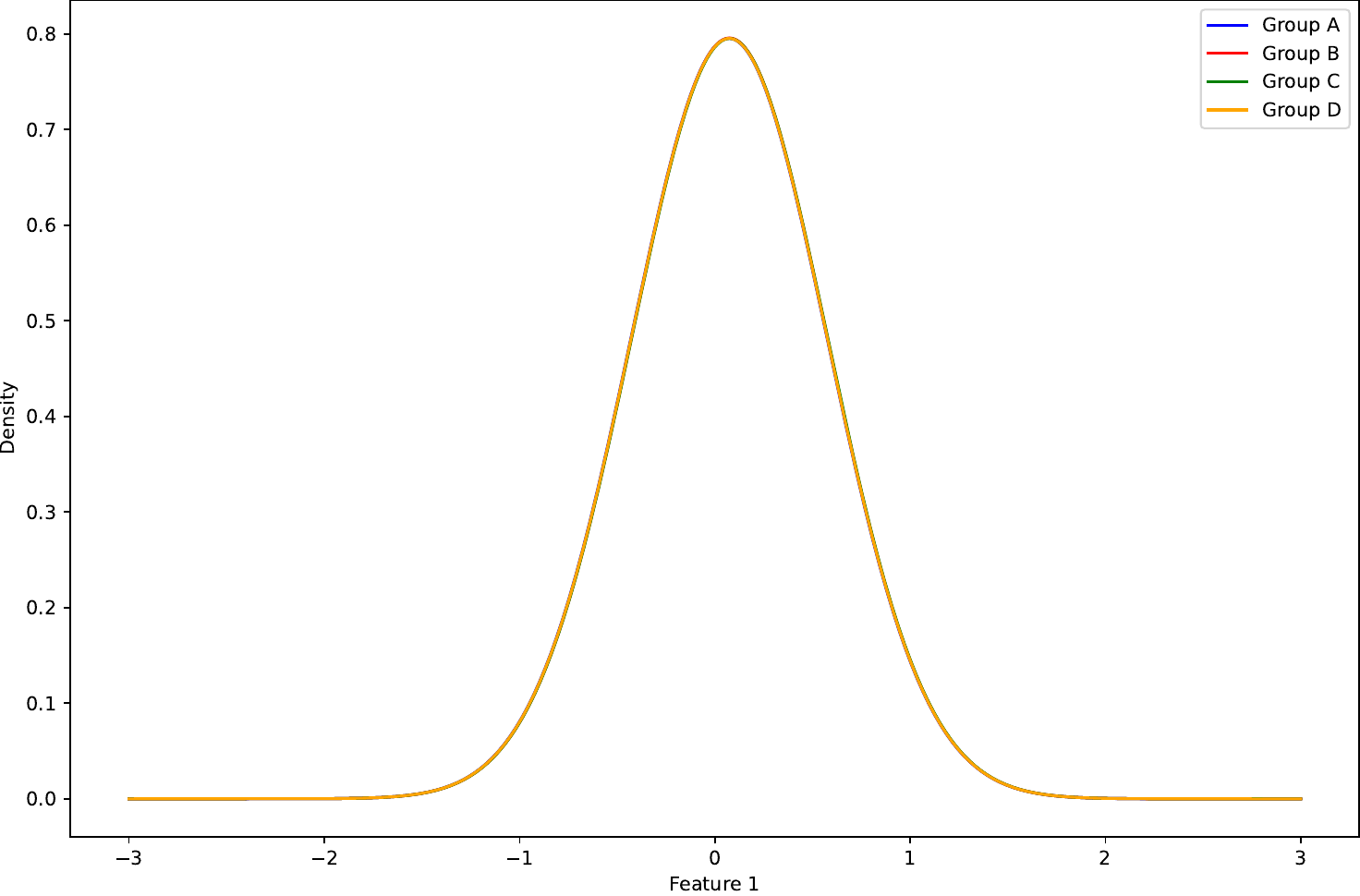}
\end{minipage}%
}%
\centering
\subfigure[$t=16$]{
\begin{minipage}{0.25\linewidth}
\centering
\includegraphics[width=0.7in]{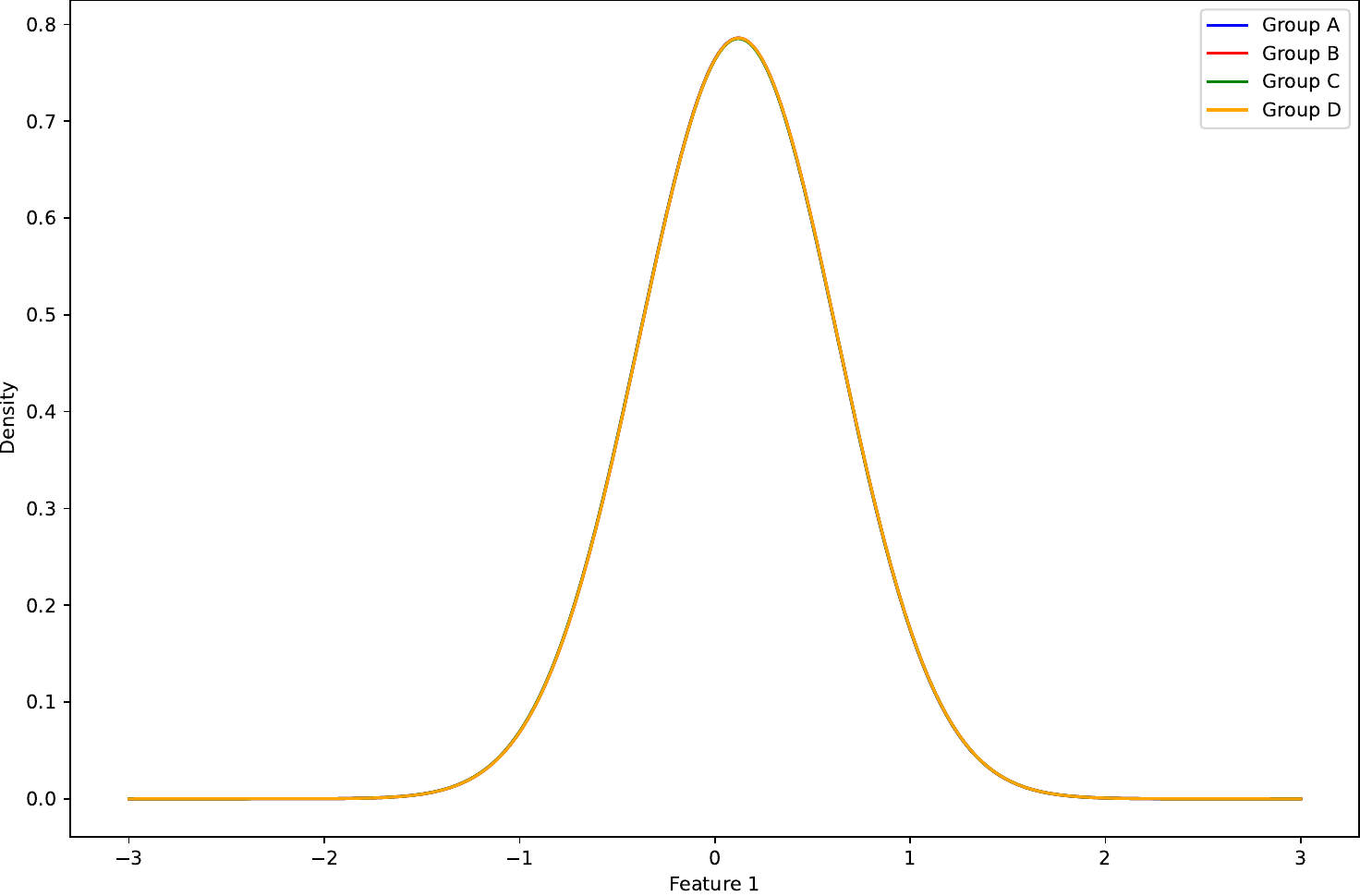}
\end{minipage}%
}%
\vspace{-.2cm}
\caption{Distribution shape for the covariate under different strengths of treatment bias and time-steps on pure synthetic dataset, (a)-(d) for $\gamma=0.2$, (e)-(h) for $\gamma=0.4$.}
\label{fig_dis_cov_pure}
\vspace{-6mm}
\end{figure}

We analyzed covariate distribution differences across treatment groups under various gamma values (indicating treatment bias strength) in our generated time-series data. Due to space constraints, we only show the distribution for one covariate, but similar patterns exist in others. Using KDE, we plotted covariate distributions for time steps 1, 4, 8 and 16 under $\gamma=0.2$ and $0.4$. Figure \ref{fig_dis_cov_pure} reveals that early in the time series (e.g., time-step=1,4), covariate distribution differences across groups grow with increasing gamma value. However, at later time steps (e.g., time-step=8, 16), these differences diminish, suggesting treatment bias lessens over time and making balance-focused methods less effective.

\begin{figure}[h]
\vspace{-4mm}
\centering
\subfigure[$\gamma=1$, Balanced CT]{
\begin{minipage}{0.5\linewidth}
\centering
\includegraphics[width=1.6in]{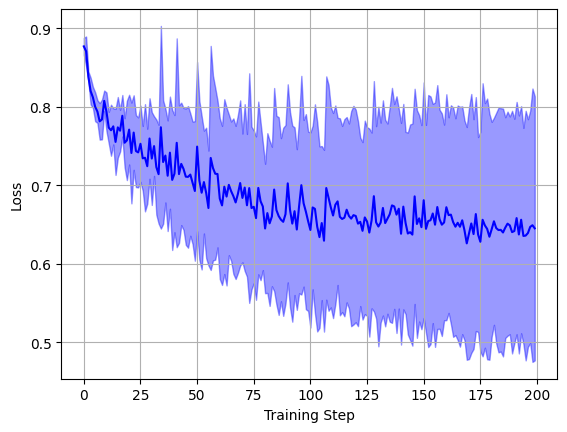}
\end{minipage}%
}%
\subfigure[$\gamma=1$, Non-balanced CT]{
\begin{minipage}{0.5\linewidth}
\centering
\includegraphics[width=1.6in]{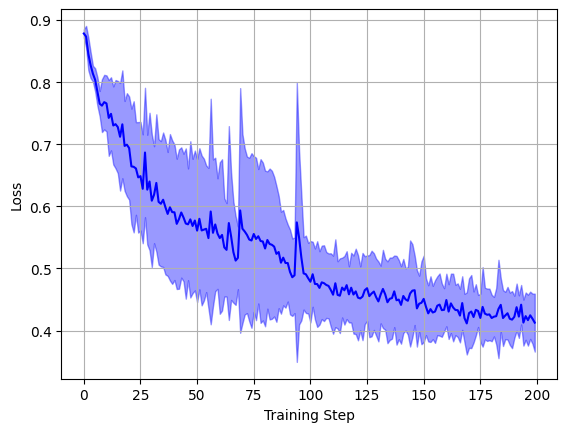}
\end{minipage}%
}
\\
\subfigure[$\gamma=10$, Balanced CT]{
\begin{minipage}{0.5\linewidth}
\centering
\includegraphics[width=1.6in]{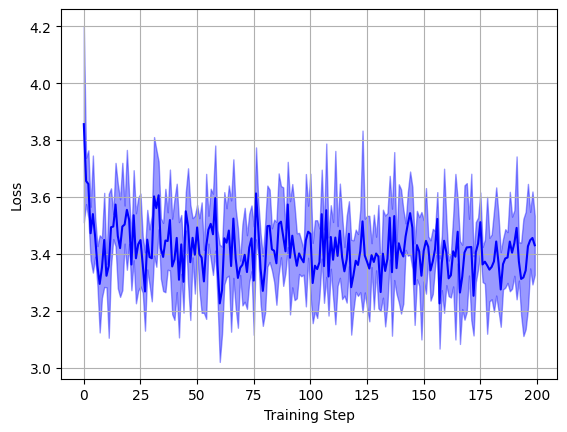}
\end{minipage}%
}%
\centering
\subfigure[$\gamma=10$, Non-balanced CT]{
\begin{minipage}{0.5\linewidth}
\centering
\includegraphics[width=1.6in]{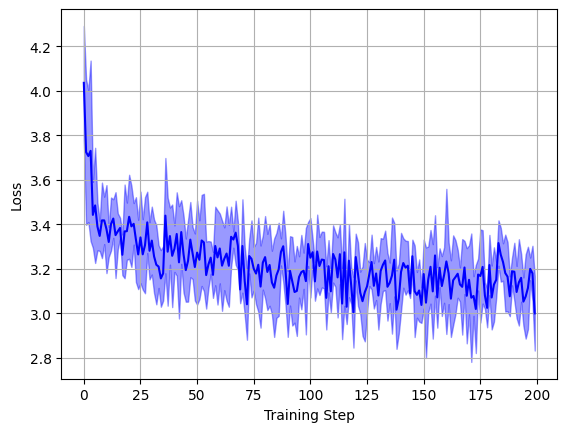}
\end{minipage}%
}%

\centering
\vspace{-3mm}
\caption{Reconstruction loss curves for Causal Transformer.}
\centering
\label{fig_reconsreuction_loss}
\vspace{-6mm}
\end{figure}

\begin{figure}[h]
\vspace{-4mm}
\centering
\subfigure[$t=1$, CT, Balanced]{
\begin{minipage}{0.5\linewidth}
\centering
\includegraphics[width=1.5in]{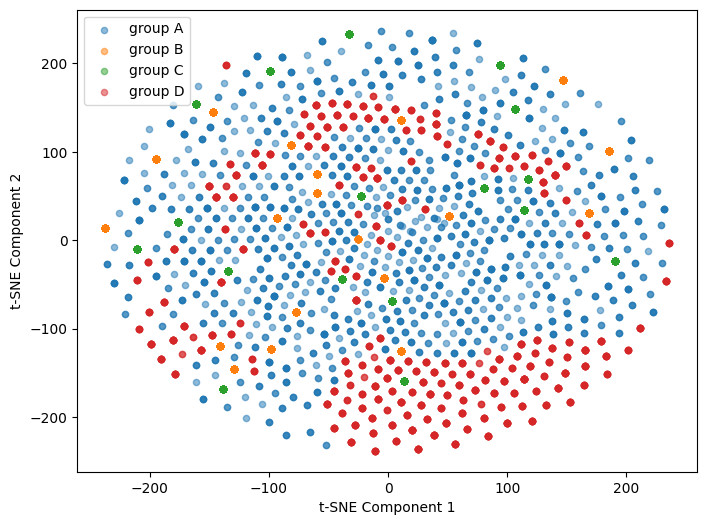}
\end{minipage}%
}%
\centering
\subfigure[$t=10$, CT, Balanced]{
\begin{minipage}{0.5\linewidth}
\centering
\includegraphics[width=1.5in]{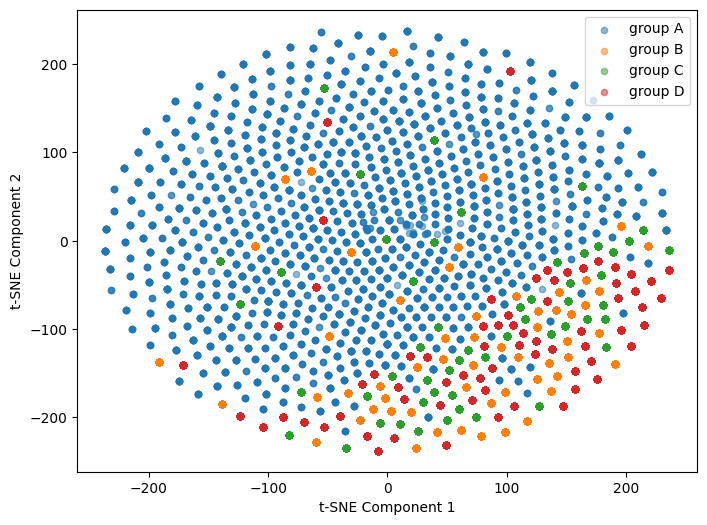}
\end{minipage}%
}%
\\
\centering
\subfigure[$t=1$, CT, Non-balanced]{
\begin{minipage}{0.5\linewidth}
\centering
\includegraphics[width=1.5in]{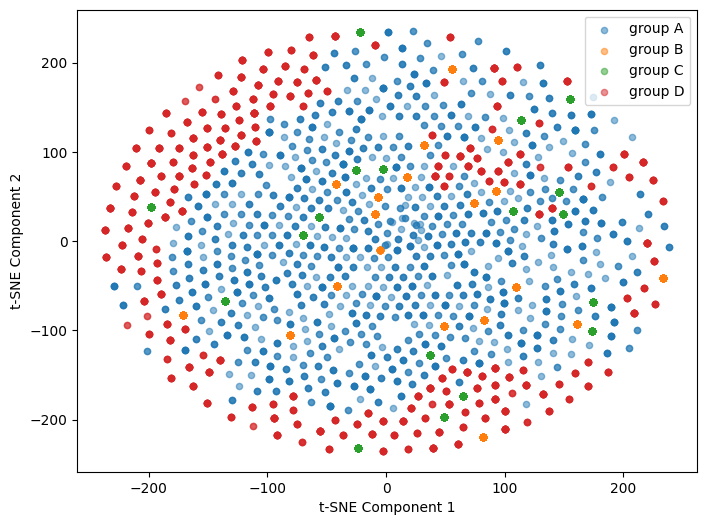}
\end{minipage}%
}%
\centering
\subfigure[$t=10$, CT, Non-balanced]{
\begin{minipage}{0.5\linewidth}
\centering
\includegraphics[width=1.5in]{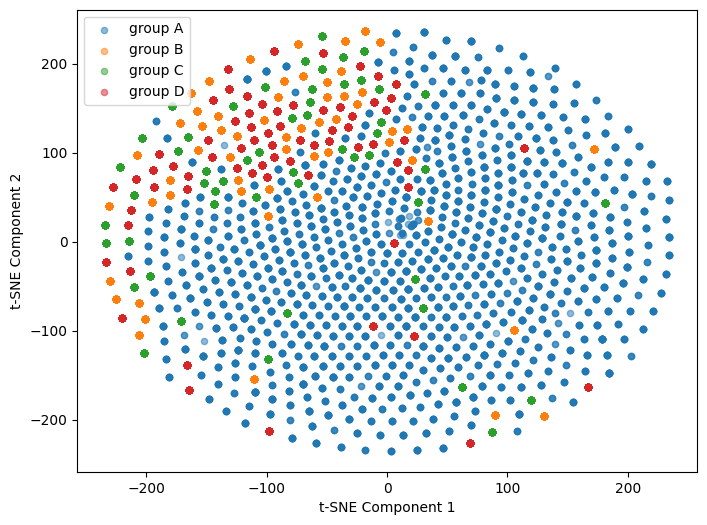}
\end{minipage}%
}%

\centering
\vspace{-3mm}
\caption{Visualization for balanced and non-balanced representation for different time steps from Causal Transformer.}
\label{rep_visual}
\vspace{-6mm}
\end{figure}

\textbf{Learned representations for reconstruction of covariates}.
We analyze if models, specifically the Causal Transformer and CRN, can accurately reconstruct original covariates with and without the balancing module. The goal is to determine if the balancing module leads to information loss. After training these models using standard supervised learning, we compare the representations of test data before and after applying the balance module. We then use an LSTM-based decoder to reconstruct the covariates from these representations. The reconstruction's accuracy is measured using the MSE between the reconstructed and original covariates.

Figure \ref{fig_reconsreuction_loss} shows the reconstruction loss for the Causal Transformer with balanced and non-balanced representations. At a high treatment bias ($\gamma=10$), the balanced representation's error doesn't converge, suggesting a significant loss of original covariate information. Conversely, the non-balanced representation's error decreases, indicating better retention of covariate information. When $\gamma=1$, both representations perform adequately, but the non-balanced one has a lower final loss, because, at low treatment biases, the balanced representation's negative impact is less noticeable. The results for the CRN model, not shown due to space constraints, follow a similar pattern.

\begin{table*}[t]
\centering
\caption{RMSE (mean $\pm$ std) comparison between sequential and non-sequential models in terms of different strengths of treatment bias.}
\label{compaison_rmse}
\vskip 0.10in
\renewcommand{\arraystretch}{1.1}
\resizebox{1.0\textwidth}{!}{
\begin{tabular}{|cc|c|c|c|c|c|c|c|}
\hline
\multicolumn{2}{|c|}{Tumor}                                         & $\gamma$=0    & 1            & 2            & 3            & 6            & 8            & 10           \\ \hline
\multicolumn{1}{|c|}{\multirow{4}{*}{Sequential}}     & CT          & 0.307$\pm$0.072  & 0.259$\pm$0.049 & 0.444$\pm$0.142 & 0.935$\pm$0.178 & 3.657$\pm$1.670 & 4.336$\pm$1.335 & 3.827$\pm$1.952 \\ \cline{2-9} 
\multicolumn{1}{|c|}{}                                & CT w/o BRM  & 0.299$\pm$0.071  & 0.325$\pm$0.105 & 0.456$\pm$0.109 & 0.852$\pm$0.234 & 2.406$\pm$0.646 & 3.108$\pm$0.307 & 3.891$\pm$3.241 \\ \cline{2-9} 
\multicolumn{1}{|c|}{}                                & CRN         & 0.239$\pm$0.017  & 0.295$\pm$0.088 & 0.323$\pm$0.049 & 0.488$\pm$0.124 & 0.812$\pm$0.164 & 1.146$\pm$0.151 & 1.483$\pm$0.173 \\ \cline{2-9} 
\multicolumn{1}{|c|}{}                                & CRN w/o BRM & 0.243$\pm$0.018  & 0.271$\pm$0.042 & 0.282$\pm$0.042 & 0.496$\pm$0.220 & 0.902$\pm$0.127 & 1.197$\pm$0.173 & 1.324$\pm$0.133 \\ \hline
\multicolumn{1}{|c|}{\multirow{5}{*}{Non-Sequential}} & CFR         & 0.340$\pm$0.024  & 0.390$\pm$0.036 & 0.511$\pm$0.103 & 0.749$\pm$0.139 & 1.727$\pm$0.235 & 2.107$\pm$0.278 & 2.419$\pm$0.297 \\ \cline{2-9} 
\multicolumn{1}{|c|}{}                                & TARnet      & 0.330$\pm$0.025  & 0.393$\pm$0.040 & 0.512$\pm$0.097 & 0.767$\pm$0.124 & 1.827$\pm$0.236 & 2.382$\pm$0.286 & 2.842$\pm$0.330 \\ \cline{2-9} 
\multicolumn{1}{|c|}{}                                & OLS1        & 0.557$\pm$0.017  & 0.617$\pm$0.033 & 0.738$\pm$0.056 & 0.951$\pm$0.075 & 1.667$\pm$0.131 & 1.872$\pm$0.136 & 1.866$\pm$0.128 \\ \cline{2-9} 
\multicolumn{1}{|c|}{}                                & OLS2        & 0.594$\pm$0.036  & 0.590$\pm$0.009 & 0.595$\pm$0.031 & 0.629$\pm$0.038 & 0.800$\pm$0.021 & 0.892$\pm$0.020 & 0.959$\pm$0.046 \\ \cline{2-9} 
\multicolumn{1}{|c|}{}                                & Random      & 0.824$\pm$0.025  & 0.868$\pm$0.034 & 0.940$\pm$0.055 & 1.076$\pm$0.062 & 1.538$\pm$0.120 & 1.724$\pm$0.122 & 1.835$\pm$0.131 \\ \hline
\multicolumn{2}{|c|}{MIMIC III}                                     & $\gamma$=0.25 & 1.0          & 2.5          & 5            & 10           & 15           & 50           \\ \hline
\multicolumn{1}{|c|}{\multirow{4}{*}{Sequential}}     & CT          & 0.084$\pm$0.015  & 0.108$\pm$0.016 & 0.128$\pm$0.012 & 0.152$\pm$0.023 & 0.188$\pm$0.049 & 0.226$\pm$0.066 & 0.490$\pm$0.128 \\ \cline{2-9} 
\multicolumn{1}{|c|}{}                                & CT w/o BRM  & 0.085$\pm$0.014  & 0.106$\pm$0.017 & 0.125$\pm$0.010 & 0.145$\pm$0.014 & 0.177$\pm$0.045 & 0.202$\pm$0.031 & 0.480$\pm$0.130 \\ \cline{2-9} 
\multicolumn{1}{|c|}{}                                & CRN         & 0.104$\pm$0.013  & 0.159$\pm$0.009 & 0.225$\pm$0.044 & 0.265$\pm$0.048 & 0.310$\pm$0.054 & 0.352$\pm$0.057 & 0.445$\pm$0.094 \\ \cline{2-9} 
\multicolumn{1}{|c|}{}                                & CRN w/o BRM & 0.099$\pm$0.012  & 0.122$\pm$0.013 & 0.133$\pm$0.016 & 0.147$\pm$0.028 & 0.188$\pm$0.081 & 0.238$\pm$0.108 & 0.503$\pm$0.137 \\ \hline
\multicolumn{1}{|c|}{\multirow{5}{*}{Non-Sequential}} & CFR         & 0.265$\pm$0.058  & 0.307$\pm$0.055 & 0.328$\pm$0.051 & 0.336$\pm$0.049 & 0.341$\pm$0.049 & 0.341$\pm$0.044 & 0.366$\pm$0.052 \\ \cline{2-9} 
\multicolumn{1}{|c|}{}                                & TARnet      & 0.271$\pm$0.057  & 0.318$\pm$0.059 & 0.349$\pm$0.061 & 0.358$\pm$0.060 & 0.364$\pm$0.054 & 0.375$\pm$0.048 & 0.428$\pm$0.046 \\ \cline{2-9} 
\multicolumn{1}{|c|}{}                                & OLS1        & 0.465$\pm$0.091  & 0.525$\pm$0.071 & 0.535$\pm$0.072 & 0.545$\pm$0.076 & 0.559$\pm$0.080 & 0.566$\pm$0.083 & 0.586$\pm$0.097 \\ \cline{2-9} 
\multicolumn{1}{|c|}{}                                & OLS2        & 0.300$\pm$0.042  & 0.353$\pm$0.038 & 0.385$\pm$0.049 & 0.403$\pm$0.065 & 0.405$\pm$0.073 & 0.403$\pm$0.065 & 0.437$\pm$0.078 \\ \cline{2-9} 
\multicolumn{1}{|c|}{}                                & Random      & 0.353$\pm$0.004  & 0.427$\pm$0.038 & 0.440$\pm$0.070 & 0.443$\pm$0.081 & 0.452$\pm$0.087 & 0.458$\pm$0.088 & 0.470$\pm$0.090 \\ \hline
\end{tabular}
}
\end{table*}

\textbf{Learned representation visualization with or without balanced module}. 
We visualized the representations learned with and without balancing modules at various time steps to investigate their balancing differences. Given that previous results showed minor covariate distribution differences among treatment groups at low treatment bias intensities, we focus on visualizations at $\gamma=10$ in the synthetic tumor dataset for a clearer demonstration of the balancing module's distribution adjustment capabilities. The representation visualization for the Causal Transformer is illustrated in Figure \ref{rep_visual}.  We can see that the representations with or without the balancing module do not differ much in terms of balancing degree across different groups (similar observations are also made for CRN). However, it is evident that the distribution of representations after balancing significantly differs from that before balancing. In other words, the balancing module not only fails to eliminate treatment bias, but also leads to a decline in predictive performance due to the altered distribution of covariate representations.

The experiments above suggest that models with balanced module struggle to reconstruct covariates under high treatment bias since they lose essential information during training. Conversely, with lower treatment bias and minor differences in group covariate distributions, the balanced module can result in over-adjustment of variables and decreased predictive performance for counterfactual estimation.

\subsection{When Are the Temporal Dependencies Informative for Counterfactual Estimation?}

We examined the effectiveness of temporal dependency modeling in counterfactual outcome estimation by comparing the performance of sequential models (Causal Transformer, CRN, and their ERM variants) against classical non-sequential models. The non-sequential baselines included: (1) Counterfactual Regression (CFR) \cite{shalit2017estimating}; (2) TarNet \cite{shalit2017estimating}, a CFR variant without the balancing module; (3) OLS1, a linear regression model combining treatment assignment and covariates; (4) OLS2, with separate linear regression models for treatment and control groups; and (5) the Random method, predicting outcomes by randomly selecting from historical outcomes.

The results are shown in Table \ref{compaison_rmse}. When the treatment bias is relatively low, sequential models outperform non-sequential models and the Random method. However, as the treatment bias increases to a certain extent, the sequential models, as mentioned in the previous analysis, tend to lose a considerable amount of original covariate information due to excessive treatment bias. Consequently, the performance of counterfactual estimation falls behind that of non-sequential models, and in some cases, even the Random method. This suggests that temporal dependency information can be overshadowed when the treatment bias is too large, rendering it ineffective for model generalization.

\section{Summary and Future Work}

In this paper, we conduct a thorough empirical study on existing popular methods  for counterfactual estimation in temporal settings. It has been observed that the effectiveness and applicability of representation-based balancing strategies could differ from previously reported results.  We identify the limitations of existing representation balance techniques in various scenarios. Our analysis results suggest three promising directions to pursue when developing new solutions to  temporal counterfactual estimation:

\begin{itemize}
    \item \textbf{Trade-off between balancing and prediction accuracy}. Although balancing strategies can mitigate the impact of treatment bias, they inevitably disrupt the data distribution, which in turn affects the accuracy of outcome predictions. Therefore, one promising future direction is to investigate appropriate strategies to balance these two factors.

    \item \textbf{More stable balancing technologies}. Our analysis reveals that balancing strategies can lead to instability in temporal model training, resulting in high variance in prediction. Therefore, new balancing strategies that can offer strong stability should be explored.

    \item \textbf{Treatment bias checking in advance}. Excessive treatment bias can results in various challenges, such as masking important temporal information in the data and leading to overall model collapse. Furthermore, treatment bias may change over time (e.g., gradually diminishing). Therefore, exploring strategies to assess how significant treatment bias is and decide if the balancing strategies are needed could effectively help to mitigate the model collapse issue.
    
\end{itemize}

\section*{Acknowledgements}
The authors would like to thank the anonymous referees for their valuable comments. The work is done while the first author is a visiting student at USC. Qiang Huang and Yi Chang are supported by the National Key R\&D Program of China under Grant No.2023YFF0905400 and the National Natural Science Foundation of China
(No.U2341229). Chuizheng Meng, Defu Cao, and Yan Liu are supported in part by NSF Research Grant \# 2226087 and \#1837131. 

\section*{Impact Statement}
This paper presents work whose goal is to advance the field of Machine Learning and Causal Inference. There are many potential societal consequences of our work, none which we feel must be specifically highlighted here.

\nocite{langley00}

\bibliography{example_paper}
\bibliographystyle{icml2024}

\newpage
\appendix
\onecolumn
\section*{\centering Appendix}

\section{Assumptions}
The task of counterfactual outcome estimation is based on the potential outcome framework, where three fundamental assumptions ensure its causal identifiability. Here, we introduce the three basic assumptions of causal identifiability in a temporal setting.

\begin{assumption}
    \textbf{Consistency}. If $\bar{\boldsymbol{A}}_{t}$=$\bar{\boldsymbol{a}}_t$ represents a sequence of treatments assigned to a specific individual, then under this treatment sequence, the potential outcome $\bar{\boldsymbol{Y}}_{t+1}(\bar{\boldsymbol{A}}_t=\bar{\boldsymbol{a}}_t)=\boldsymbol{Y}_t$, where $\boldsymbol{Y}_t$ is the observed outcome conditioned on $\bar{\boldsymbol{A}}_t=\bar{\boldsymbol{a}}_t$. This implies that the potential outcome for an individual, given their observed exposure history, corresponds to the actual (factual) outcome that will be observed for that individual.
\end{assumption}
\begin{assumption}
\textbf{Overlap}. In the entire historical timeline, there is a nonzero likelihood of either receiving or not receiving any treatment, i.e., given the history of an individual, the probability of treatment assignment holds $0<P(\boldsymbol{A}_t=\boldsymbol{a}_t|\bar{\boldsymbol{H}}_t=\bar{\boldsymbol{h}}_t)<1$ if $P(\bar{\boldsymbol{H}}_t=\bar{\boldsymbol{h}}_t)>0$. 
\end{assumption}

\begin{assumption}
\textbf{Unconfoundedness}. Given the observed history, the current treatment assignment is independent of the potential outcome, i.e., $\boldsymbol{A}_t \perp \boldsymbol{Y}_{t+1}(\boldsymbol{a}_t) | \bar{\boldsymbol{H}}_t, \forall \boldsymbol{a}_t $. This implies that all the possible confounders that affect both treatment and outcome are observed.
\end{assumption}

\section{Data Generation of Pure Synthetic Dataset}
We generate random time-series data using the universal method from literature \cite{bica2020time}, analyzing covariate distribution across treatment groups. This involves generating initial covariates and treatments using Gaussian and Bernoulli distributions, and then creating time-series data under the proposed causal structure. The generated dataset features 10-dimensional covariates, a maximum sequence length of 30, and the two-dimensional treatment variables, following a $p$-order autoregressive process:
\begin{equation}
\small
\begin{aligned}
    X_t = &\frac{1}{p}\sum_{i=1}^p(\beta_i X_{t-i} +\sum_{j=1}^k\lambda_{i,j}A_{t-i,j}+\epsilon_t),\\
    \pi_{tj} &= \gamma \sum_{i=1}^p X_{t-i},
    \quad A_{tj}|\pi_{tj} \sim B(\sigma(\pi_{tj})),\\
    &Y_t = \lambda_Y \cdot f(W \cdot Z_t) + \epsilon_Y,
\end{aligned}
\end{equation}
where $k$ denotes the dimension of treatment variables, $\beta_i \sim \mathcal{N}(1-(\frac{i}{p}),\frac{1}{p})$, $\epsilon \sim \mathcal{N}(0,0.01)$; $\sigma$ denotes the sigmoid active function, $f$ denotes the nonlinear mapping function
(e.g., tanh function), $\gamma$ represents the strength of treatment bias and $\gamma_Y$ controls the amount of confounding applied to the outcome.

\section{Related works}

\textbf{Counterfactual outcome estimation under I.I.D setting}. Due to the prohibitive costs and potential ethical concerns associated with randomized experiments \cite{guo2020survey}, there has been a growing interest in recent years in causal inference from observational data, with a majority of existing studies primarily focusing on independent and identically distributed (i.i.d.) data. Traditional statistical methods aim to construct one or more pseudo-populations that differ from the original data to achieve a balance in the distribution of sample covariates among different groups, like sample re-weighting \cite{rosenbaum1983central,cerulli2014treatrew} based on IPTW \cite{chesnaye2022introduction}, Stratification \cite{imbens2015causal,miratrix2013adjusting} and matching \cite{abadie2004implementing,abadie2006large}.
Some instrument variables-based methods have also been introduced. DeepIV \cite{hartford2017deep}, for instance, introduced a two-stage approach comprising a first-stage network for treatment prediction and a second-stage network incorporating a loss function that encompasses integration over the conditional treatment distribution for counterfactual predictions. Confounder Balanced IV Regression (CB-IV) \cite{wu2022instrumental} jointly remove the bias from the unmeasured confounders and balance the observed
confounders by twp stage IV regression. 
Recently, more methods have been proposed based on representation learning to enable more effective and more
accurate counterfactual estimation. CFR \cite{shalit2017estimating} frames counterfactual inference as a form of domain adaptation and employs neural networks to learn ITEs by creating balanced representations through the minimization of distribution differences between control and treated groups. CEVAE \cite{louizos2017causal} proposes a novel approach, mapping the original observed features to a latent space to capture hidden confounders using a variational autoencoder \cite{kingma2014auto}. Atan et al. \cite{atan2018deep} introduce Deep-Treat, which mitigates bias by learning representations and crafting effective treatment policies using deep neural networks on transformed data for counterfactual estimation. Yao et al. \cite{yao2018representation} introduce a method for counterfactual estimation, known as SITE, grounded in deep representation learning, capable of capturing hidden confounders and preserving the local similarity of data. Hassanpour et al. \cite{hassanpour2019learning} proposed to learn three underlying sources of the observation data as the disentangled representations to account for the treatment bias and then conduct unbiased counterfactual outcome estimation. Then MIM-DRCFR \cite{cheng2022learning} was proposed to identify disentangled representations by mutual information minimization for the counterfactual outcome estimation. 

\textbf{Counterfactual outcome estimation over time}. 
Early methods for temporal counterfactual outcome estimation primarily included G-computation, marginal structural models (MSMs), and structural nested models \cite{robins1986new,robins1994correcting,robins2000marginal,robins2008estimation}. However, these methods heavily rely on linear assumptions to predict counterfactual outcomes, and they become inadequate when dealing with data that contains intricate temporal dependencies.
Subsequent research endeavors have aimed to overcome the limitations of model expressiveness. This has been accomplished through the adoption of Bayesian non-parametric methods \cite{xu2016bayesian,soleimani2017treatment,schulam2017reliable} or more sophisticated deep neural networks, such as recurrent neural networks (RNNs). Notably, in the realm of Bayesian non-parametric methods, recurrent marginal structural networks (RMSNs) \cite{lim2018forecasting} have emerged, replacing the linear model of MSM with an RNN-based architecture to forecast treatment outcomes. Similarly, G-Net \cite{li2021g} integrates RNNs into the g-computation framework in place of classical regression models. Inspired by the successes of representation learning for domain adaptation and generalization \cite{ganin2016domain,tzeng2015simultaneous}, recent research ventures have delved into the development of learned representations that serve both predictive purposes for outcome estimation and alleviating treatment bias within training data. For instance, the Counterfactual Recurrent Network (CRN) \cite{Bica2020Estimating} employs an RNN-based model trained with dual objectives: the factual outcome regression loss and the gradient reversal \cite{ganin2016domain} with respect to treatment prediction. The former fosters the development of informative outcome-predictive representations, while the latter encourages the creation of uniform representations across different treatments. This joint training target yields representations that are both informative and balanced. Motivated by similar objectives, a recent study by \cite{melnychuk2022causal} replaces the RNN-based architecture with a Transformer-based \cite{vaswani2017attention} one and employs a domain confusion loss \cite{tzeng2015simultaneous} to facilitate the learning of treatment-agnostic representations.

\begin{table}[t]
\caption{RMSE (mean $\pm$ std) with $\gamma=2,6,8$ on Tumor dataset in standard supervised learning.}
\centering
\label{tumor_2_6_8_standard}
\renewcommand{\arraystretch}{1.5}
\scalebox{0.8}{
\begin{tabular}{|c|c|l|l|l|l|l|l|}
\hline
\multirow{8}{*}{$\gamma=2$} & Methods     & \multicolumn{1}{c|}{$\tau$=1}            & \multicolumn{1}{c|}{2}                   & \multicolumn{1}{c|}{3}                   & \multicolumn{1}{c|}{4}                   & \multicolumn{1}{c|}{5}                   & \multicolumn{1}{c|}{6} \\ \cline{2-8} 
                            & CT          & 0.8607\(\pm\)0.0636                      & 0.7316\(\pm\)0.0381                      & 0.7889\(\pm\)0.0362                      & 0.8443\(\pm\)0.0497                      & 0.8979\(\pm\)0.0605                      & 0.9441\(\pm\)0.0765    \\ \cline{2-8} 
                            & CT w/o BRM  & 0.8589\(\pm\)0.0605                      & 0.7193\(\pm\)0.0399                      & 0.7695\(\pm\)0.0543                      & 0.8156\(\pm\)0.0608                      & 0.8603\(\pm\)0.0709                      & 0.9004\(\pm\)0.079     \\ \cline{2-8} 
                            & CRN         & 0.8874\(\pm\)0.0643                      & 0.7512\(\pm\)0.042                       & 0.8365\(\pm\)0.0647                      & 0.9305\(\pm\)0.087                       & 1.0186\(\pm\)0.1041                      & 1.0914\(\pm\)0.1144    \\ \cline{2-8} 
                            & CRN w/o BRM & 0.9081\(\pm\)0.0833                      & 0.7253\(\pm\)0.0648                      & 0.7647\(\pm\)0.0711                      & 0.8093\(\pm\)0.0791                      & 0.8549\(\pm\)0.0865                      & 0.8978\(\pm\)0.09      \\ \cline{2-8} 
                            & RMSN        & 1.1192\(\pm\)0.1139                      & 0.9905\(\pm\)0.1068                      & 1.0193\(\pm\)0.1189                      & 1.0485\(\pm\)0.1388                      & 1.0817\(\pm\)0.1596                      & 1.1184\(\pm\)0.1782    \\ \cline{2-8} 
                            & GNET        & 1.0014\(\pm\)0.0542                      & 1.0301\(\pm\)0.07                        & 1.1816\(\pm\)0.0999                      & 1.2527\(\pm\)0.135                       & 1.2944\(\pm\)0.1635                      & 1.3194\(\pm\)0.1811    \\ \cline{2-8} 
                            & MSM         & \multicolumn{1}{c|}{1.4126\(\pm\)0.1091} & \multicolumn{1}{c|}{2.0351\(\pm\)0.2576} & \multicolumn{1}{c|}{2.387\(\pm\)0.3079}  & \multicolumn{1}{c|}{2.5739\(\pm\)0.3327} & \multicolumn{1}{c|}{2.6381\(\pm\)0.3357} & 2.6038\(\pm\)0.3251    \\ \hline
\multirow{7}{*}{$\gamma=6$} & CT          & 2.9329\(\pm\)1.0845                      & 3.6032\(\pm\)1.7021                      & 4.1373\(\pm\)1.9675                      & 4.4685\(\pm\)2.1355                      & 4.6556\(\pm\)2.2183                      & 4.6891\(\pm\)2.2337    \\ \cline{2-8} 
                            & CT w/o BRM  & 2.8848\(\pm\)0.2065                      & 2.8963\(\pm\)1.0772                      & 3.1507\(\pm\)1.1061                      & 3.2695\(\pm\)1.1335                      & 3.2695\(\pm\)1.1335                      & 3.3876\(\pm\)1.146     \\ \cline{2-8} 
                            & CRN         & 2.5019\(\pm\)0.2458                      & 3.0962\(\pm\)1.2731                      & 3.5662\(\pm\)1.6316                      & 3.763\(\pm\)1.7179                       & 3.8416\(\pm\)1.7144                      & 3.8249\(\pm\)1.66      \\ \cline{2-8} 
                            & CRN w/o BRM & 2.452\(\pm\)0.2341                       & 3.1151\(\pm\)1.1059                      & 3.5853\(\pm\)1.3143                      & 3.7225\(\pm\)1.3755                      & 3.6914\(\pm\)1.3697                      & 3.5311\(\pm\)1.3101    \\ \cline{2-8} 
                            & RMSN        & 2.4291\(\pm\)0.2032                      & 1.9708\(\pm\)0.6216                      & 2.3183\(\pm\)0.8332                      & 2.5352\(\pm\)0.8919                      & 2.6429\(\pm\)0.866                       & 2.6645\(\pm\)0.7951    \\ \cline{2-8} 
                            & GNET        & 2.1158\(\pm\)0.15                        & 1.6777\(\pm\)0.3713                      & 2.099\(\pm\)0.4588                       & 2.3431\(\pm\)0.5212                      & 2.4959\(\pm\)0.5712                      & 2.5896\(\pm\)0.6363    \\ \cline{2-8} 
                            & MSM         & \multicolumn{1}{c|}{2.9828\(\pm\)0.2093} & \multicolumn{1}{c|}{2.1097\(\pm\)0.67}   & \multicolumn{1}{c|}{1.8604\(\pm\)0.605}  & \multicolumn{1}{c|}{1.8862\(\pm\)0.5998} & \multicolumn{1}{c|}{1.8448\(\pm\)0.5613} & 1.656\(\pm\)0.4577     \\ \hline
\multirow{7}{*}{$\gamma=8$} & CT          & \multicolumn{1}{c|}{4.045\(\pm\)0.7598}  & \multicolumn{1}{c|}{4.9419\(\pm\)1.9872} & \multicolumn{1}{c|}{5.5432\(\pm\)2.1325} & \multicolumn{1}{c|}{5.9236\(\pm\)2.2364} & \multicolumn{1}{c|}{6.1506\(\pm\)2.346}  & 6.189\(\pm\)2.445      \\ \cline{2-8} 
                            & CT w/o BRM  & \multicolumn{1}{c|}{4.1487\(\pm\)1.0624} & \multicolumn{1}{c|}{4.4726\(\pm\)1.9057} & \multicolumn{1}{c|}{4.8615\(\pm\)1.8993} & \multicolumn{1}{c|}{5.05\(\pm\)1.8558}   & \multicolumn{1}{c|}{5.1302\(\pm\)1.7835} & 5.0791\(\pm\)1.6694    \\ \cline{2-8} 
                            & CRN         & \multicolumn{1}{c|}{3.8468\(\pm\)0.2727} & \multicolumn{1}{c|}{5.7004\(\pm\)2.4747} & \multicolumn{1}{c|}{6.294\(\pm\)3.0082}  & \multicolumn{1}{c|}{6.4696\(\pm\)3.0877} & \multicolumn{1}{c|}{6.4373\(\pm\)2.9626} & 6.2295\(\pm\)2.7222    \\ \cline{2-8} 
                            & CRN w/o BRM & \multicolumn{1}{c|}{3.5992\(\pm\)0.2349} & \multicolumn{1}{c|}{4.6192\(\pm\)1.3005} & \multicolumn{1}{c|}{5.3959\(\pm\)1.5165} & \multicolumn{1}{c|}{5.709\(\pm\)1.5527}  & \multicolumn{1}{c|}{5.7371\(\pm\)1.504}  & 5.5302\(\pm\)1.3913    \\ \cline{2-8} 
                            & RMSN        & \multicolumn{1}{c|}{3.4263\(\pm\)0.4333} & \multicolumn{1}{c|}{2.8244\(\pm\)1.2188} & \multicolumn{1}{c|}{3.5567\(\pm\)1.4583} & \multicolumn{1}{c|}{3.7758\(\pm\)1.4774} & \multicolumn{1}{c|}{3.7013\(\pm\)1.5067} & 3.5593\(\pm\)1.3595    \\ \cline{2-8} 
                            & GNET        & \multicolumn{1}{c|}{3.1364\(\pm\)0.3073} & \multicolumn{1}{c|}{2.856\(\pm\)0.8067}  & \multicolumn{1}{c|}{3.5623\(\pm\)1.0225} & \multicolumn{1}{c|}{3.9387\(\pm\)1.143}  & \multicolumn{1}{c|}{4.1182\(\pm\)1.1899} & 4.1359\(\pm\)1.1703    \\ \cline{2-8} 
                            & MSM         & \multicolumn{1}{c|}{4.1831\(\pm\)0.371}  & \multicolumn{1}{c|}{1.5761\(\pm\)0.5207} & \multicolumn{1}{c|}{2.1267\(\pm\)0.7158} & \multicolumn{1}{c|}{2.5203\(\pm\)0.853}  & \multicolumn{1}{c|}{2.7747\(\pm\)0.9489} & 2.9511\(\pm\)1.0315    \\ \hline
\end{tabular}
}
\end{table}

\begin{table}[t]
\centering
\caption{RMNSE (mean $\pm$ std) for M5 real-world dataset in factual outcome estiamtion.}
\label{M5_dataset_prediction}
\renewcommand{\arraystretch}{1.5}
\scalebox{0.8}{
\begin{tabular}{|c|c|c|c|c|c|c|}
\hline
Methods     & $\tau$=1            & 2                   & 3                    & 4                    & 5                    & 6                    \\ \hline
CT          & 4.5934\(\pm\)0.0823 & 8.9863\(\pm\)0.1888 & 9.5868\(\pm\)0.1934  & 9.9135\(\pm\)0.2266  & 10.1454\(\pm\)0.2535 & 10.3476\(\pm\)0.2838 \\ \hline
CT w/o BRM  & 4.5908\(\pm\)0.0855 & 8.9802\(\pm\)0.1814 & 9.5806\(\pm\)0.1789  & 9.9029\(\pm\)0.2047  & 10.1342\(\pm\)0.2273 & 10.3354\(\pm\)0.2551 \\ \hline
CRN         & 4.9843\(\pm\)0.316  & 9.1488\(\pm\)0.1674 & 9.7878\(\pm\)0.1619  & 10.1161\(\pm\)0.1683 & 10.3767\(\pm\)0.1921 & 10.5927\(\pm\)0.2228 \\ \hline
CRN w/o BRM & 4.6808\(\pm\)0.0554 & 9.0483\(\pm\)0.1636 & 9.6803\(\pm\)0.1535  & 10.004\(\pm\)0.1625  & 10.2566\(\pm\)0.1779 & 10.4691\(\pm\)0.1971 \\ \hline
RMSN        & 5.1912\(\pm\)0.1169 & 9.7308\(\pm\)0.2557 & 10.4807\(\pm\)0.3374 & 11.069\(\pm\)0.4258  & 11.632\(\pm\)0.532   & 12.1559\(\pm\)0.6548 \\ \hline
\end{tabular}
}
\end{table}

\begin{table}[h]
\centering
\caption{RMSE (mean $\pm$ std) performance on Tumor for short-term history cold-start case.}
\label{short_term_2_6_8}
\renewcommand{\arraystretch}{1.5}
\scalebox{0.8}{
\begin{tabular}{|c|c|l|l|l|l|l|l|}
\hline
\multirow{8}{*}{$\gamma=2$} & Methods     & \multicolumn{1}{c|}{$\tau$=1}            & \multicolumn{1}{c|}{2}                   & \multicolumn{1}{c|}{3}                   & \multicolumn{1}{c|}{4}                   & \multicolumn{1}{c|}{5}                   & \multicolumn{1}{c|}{6} \\ \cline{2-8} 
                            & CT          & 0.9233\(\pm\)0.0605                      & 0.7834\(\pm\)0.0572                      & 0.8694\(\pm\)0.0643                      & 0.9514\(\pm\)0.0669                      & 1.0158\(\pm\)0.0741                      & 1.0733\(\pm\)0.0801    \\ \cline{2-8} 
                            & CT w/o BRM  & 0.9215\(\pm\)0.0527                      & 0.7955\(\pm\)0.0575                      & 0.8797\(\pm\)0.0698                      & 0.9564\(\pm\)0.077                       & 1.0177\(\pm\)0.0863                      & 1.0709\(\pm\)0.0884    \\ \cline{2-8} 
                            & CRN         & 0.9411\(\pm\)0.045                       & 0.848\(\pm\)0.0838                       & 0.9751\(\pm\)0.0981                      & 1.1063\(\pm\)0.1163                      & 1.2235\(\pm\)0.1292                      & 1.3169\(\pm\)0.1323    \\ \cline{2-8} 
                            & CRN w/o BRM & 0.9632\(\pm\)0.0622                      & 0.8061\(\pm\)0.0956                      & 0.8835\(\pm\)0.1173                      & 0.9631\(\pm\)0.1374                      & 1.0368\(\pm\)0.1517                      & 1.1018\(\pm\)0.1573    \\ \cline{2-8} 
                            & RMSN        & 1.1651\(\pm\)0.0931                      & 1.0225\(\pm\)0.1023                      & 1.0943\(\pm\)0.1043                      & 1.1698\(\pm\)0.1118                      & 1.2383\(\pm\)0.1237                      & 1.3018\(\pm\)0.1319    \\ \cline{2-8} 
                            & GNET        & 1.0559\(\pm\)0.058                       & 5.0544\(\pm\)0.1537                      & 6.1204\(\pm\)0.2822                      & 6.4523\(\pm\)0.4056                      & 6.4235\(\pm\)0.5026                      & 6.174\(\pm\)0.5921     \\ \cline{2-8} 
                            & MSM         & \multicolumn{1}{c|}{1.4994\(\pm\)0.0684} & \multicolumn{1}{c|}{2.4804\(\pm\)0.1105} & \multicolumn{1}{c|}{2.9169\(\pm\)0.1328} & \multicolumn{1}{c|}{3.1447\(\pm\)0.1411} & \multicolumn{1}{c|}{3.2205\(\pm\)0.142}  & 3.1734\(\pm\)0.1437    \\ \hline
\multirow{7}{*}{$\gamma=6$} & CT          & 3.1077\(\pm\)1.0199                      & 4.3539\(\pm\)3.144                       & 5.0768\(\pm\)3.6509                      & 5.4948\(\pm\)3.9159                      & 5.7599\(\pm\)3.951                       & 5.7262\(\pm\)3.8615    \\ \cline{2-8} 
                            & CT w/o BRM  & 2.8129\(\pm\)0.3076                      & 3.0805\(\pm\)1.2024                      & 3.4053\(\pm\)1.2094                      & 3.6102\(\pm\)0.9977                      & 3.8604\(\pm\)0.8948                      & 3.991\(\pm\)0.8221     \\ \cline{2-8} 
                            & CRN         & 2.7107\(\pm\)0.295                       & 4.5699\(\pm\)1.1237                      & 5.1567\(\pm\)1.2849                      & 5.356\(\pm\)1.289                        & 5.3911\(\pm\)1.2332                      & 5.2859\(\pm\)1.1368    \\ \cline{2-8} 
                            & CRN w/o BRM & 2.6608\(\pm\)0.2909                      & 4.7513\(\pm\)0.871                       & 5.4321\(\pm\)0.8854                      & 5.5918\(\pm\)0.8353                      & 5.4986\(\pm\)0.7715                      & 5.2123\(\pm\)0.6938    \\ \cline{2-8} 
                            & RMSN        & 2.6146\(\pm\)0.174                       & 2.7629\(\pm\)0.6637                      & 3.4115\(\pm\)0.7996                      & 3.7641\(\pm\)0.8695                      & 3.9148\(\pm\)0.8473                      & 3.9114\(\pm\)0.7832    \\ \cline{2-8} 
                            & GNET        & 2.2245\(\pm\)0.2132                      & 4.6513\(\pm\)0.4998                      & 5.4084\(\pm\)0.6527                      & 5.6466\(\pm\)0.7452                      & 5.6574\(\pm\)0.7861                      & 5.5463\(\pm\)0.7823    \\ \cline{2-8} 
                            & MSM         & \multicolumn{1}{c|}{3.2872\(\pm\)0.2421} & \multicolumn{1}{c|}{3.0459\(\pm\)0.4111} & \multicolumn{1}{c|}{2.7305\(\pm\)0.3868} & \multicolumn{1}{c|}{2.7901\(\pm\)0.4116} & \multicolumn{1}{c|}{2.7366\(\pm\)0.4222} & 2.4671\(\pm\)0.4189    \\ \hline
\multirow{7}{*}{$\gamma=8$} & CT          & \multicolumn{1}{c|}{4.1338\(\pm\)0.5542} & \multicolumn{1}{c|}{5.7549\(\pm\)2.494}  & \multicolumn{1}{c|}{6.7353\(\pm\)2.8143} & \multicolumn{1}{c|}{7.2575\(\pm\)2.9254} & \multicolumn{1}{c|}{7.4752\(\pm\)2.8458} & 7.6408\(\pm\)2.7925    \\ \cline{2-8} 
                            & CT w/o BRM  & \multicolumn{1}{c|}{4.0003\(\pm\)0.6324} & \multicolumn{1}{c|}{4.0751\(\pm\)1.4148} & \multicolumn{1}{c|}{4.568\(\pm\)1.5578}  & \multicolumn{1}{c|}{4.8\(\pm\)1.5702}    & \multicolumn{1}{c|}{4.8626\(\pm\)1.5929} & 5.1146\(\pm\)1.4664    \\ \cline{2-8} 
                            & CRN         & \multicolumn{1}{c|}{4.0312\(\pm\)0.2933} & \multicolumn{1}{c|}{8.3633\(\pm\)3.9551} & \multicolumn{1}{c|}{9.1368\(\pm\)4.2186} & \multicolumn{1}{c|}{9.3202\(\pm\)4.063}  & \multicolumn{1}{c|}{9.23\(\pm\)3.8222}   & 8.8752\(\pm\)3.4921    \\ \cline{2-8} 
                            & CRN w/o BRM & \multicolumn{1}{c|}{3.7808\(\pm\)0.2239} & \multicolumn{1}{c|}{6.6869\(\pm\)1.2202} & \multicolumn{1}{c|}{7.7847\(\pm\)1.3929} & \multicolumn{1}{c|}{8.2069\(\pm\)1.4314} & \multicolumn{1}{c|}{8.217\(\pm\)1.4154}  & 7.8825\(\pm\)1.3351    \\ \cline{2-8} 
                            & RMSN        & \multicolumn{1}{c|}{3.6366\(\pm\)0.4517} & \multicolumn{1}{c|}{3.8651\(\pm\)1.5634} & \multicolumn{1}{c|}{5.0449\(\pm\)1.6389} & \multicolumn{1}{c|}{5.3907\(\pm\)1.5291} & \multicolumn{1}{c|}{5.248\(\pm\)1.5433}  & 5.0034\(\pm\)1.3334    \\ \cline{2-8} 
                            & GNET        & \multicolumn{1}{c|}{3.1837\(\pm\)0.2193} & \multicolumn{1}{c|}{5.6514\(\pm\)0.9173} & \multicolumn{1}{c|}{6.5778\(\pm\)1.0536} & \multicolumn{1}{c|}{6.9325\(\pm\)1.0519} & \multicolumn{1}{c|}{6.987\(\pm\)0.9693}  & 6.8128\(\pm\)0.8408    \\ \cline{2-8} 
                            & MSM         & \multicolumn{1}{c|}{4.3752\(\pm\)0.4191} & \multicolumn{1}{c|}{2.2088\(\pm\)0.3828} & \multicolumn{1}{c|}{2.9991\(\pm\)0.5359} & \multicolumn{1}{c|}{3.5619\(\pm\)0.6461} & \multicolumn{1}{c|}{3.9347\(\pm\)0.7295} & 4.2067\(\pm\)0.8001    \\ \hline
\end{tabular}
}
\end{table}

\section{Additional Experiment Results}
\subsection{Detailed Description of Baselines}
\textbf{Transformer-based}:
\begin{itemize}
    \item Causal Transformer (CT) \cite{melnychuk2022causal}: A transformer-based counterfactual outcome prediction model, which is specifically designed to capture complex, long-range dependencies among time-varying confounders. The model proposed a domain confusion module (CDC) to obtain the adversarial balanced representations for addressing confounding bias.
\end{itemize}

\textbf{LSTM-based}:
\begin{itemize}
    \item Counterfactual Recurrent Network(CRN) \cite{Bica2020Estimating}:  A sequence-to-sequence model that leverages the increasingly available patient observational data to estimate treatment effects over time and maximize the loss of treatment classifier for eliminating the confounding bias by adversarial gradient reversal.

    \item Recurrent Marginal Structural Networks(RMSN) \cite{lim2018forecasting}: A LSTM-based model for time series data, which uses propensity weighting to adjust for time-dependent confounders.

    \item G-Net \cite{li2021g}:  A sequential deep learning framework for counterfactual prediction under dynamic time-varying treatment strategies in complex longitudinal settings based on G-computation.
\end{itemize}
\textbf{Linear-based}:
\begin{itemize}
    \item Marginal Structural Model (MSM) \cite{robins2000marginal}. A linear marginal structural model based on the inverse-probability-of-treatment weighted estimator. 
\end{itemize}

\begin{table}[t]
\centering
\caption{RMSE (mean $\pm$ std) performance on Tumor for distribution shift cold-start case.}
\label{disreibution_shift_2_6_8}
\renewcommand{\arraystretch}{1.5}
\scalebox{0.8}{
\begin{tabular}{|c|c|l|l|l|l|l|l|}
\hline
\multirow{8}{*}{$\gamma=2$} & Methods     & \multicolumn{1}{c|}{$\tau$=1}            & \multicolumn{1}{c|}{2}                   & \multicolumn{1}{c|}{3}                   & \multicolumn{1}{c|}{4}                   & \multicolumn{1}{c|}{5}                   & \multicolumn{1}{c|}{6} \\ \cline{2-8} 
                            & CT          & 1.7095\(\pm\)1.0786                      & 2.4904\(\pm\)2.5066                      & 3.0249\(\pm\)3.2811                      & 3.4384\(\pm\)3.7743                      & 3.5949\(\pm\)3.8012                      & 3.702\(\pm\)3.8142     \\ \cline{2-8} 
                            & CT w/o BRM  & 1.0343\(\pm\)0.1086                      & 1.0491\(\pm\)0.102                       & 1.1595\(\pm\)0.1364                      & 1.2747\(\pm\)0.1549                      & 1.3739\(\pm\)0.1682                      & 1.4078\(\pm\)0.1844    \\ \cline{2-8} 
                            & CRN         & 1.626\(\pm\)0.0804                       & 2.0827\(\pm\)0.2604                      & 2.1672\(\pm\)0.4988                      & 2.2263\(\pm\)0.5693                      & 2.2678\(\pm\)0.563                       & 2.284\(\pm\)0.5354     \\ \cline{2-8} 
                            & CRN w/o BRM & 1.4259\(\pm\)0.0824                      & 1.4815\(\pm\)0.1531                      & 1.6756\(\pm\)0.1583                      & 1.7945\(\pm\)0.1491                      & 1.8476\(\pm\)0.1399                      & 1.834\(\pm\)0.1368     \\ \cline{2-8} 
                            & RMSN        & 1.4809\(\pm\)0.1192                      & 1.3126\(\pm\)0.1                         & 1.3464\(\pm\)0.1102                      & 1.3807\(\pm\)0.1043                      & 1.4222\(\pm\)0.1102                      & 1.4675\(\pm\)0.1519    \\ \cline{2-8} 
                            & GNET        & 1.1463\(\pm\)0.0576                      & 4.022\(\pm\)0.0451                       & 4.5349\(\pm\)0.1598                      & 4.6286\(\pm\)0.2901                      & 4.6283\(\pm\)0.4205                      & 4.6054\(\pm\)0.5494    \\ \cline{2-8} 
                            & MSM         & \multicolumn{1}{c|}{1.3067\(\pm\)0.1205} & \multicolumn{1}{c|}{0.5554\(\pm\)0.0677} & \multicolumn{1}{c|}{0.8374\(\pm\)0.1064} & \multicolumn{1}{c|}{1.0729\(\pm\)0.1382} & \multicolumn{1}{c|}{1.2844\(\pm\)0.1669} & 1.4894\(\pm\)0.1901    \\ \hline
\multirow{7}{*}{$\gamma=6$} & CT          & 3.3542\(\pm\)1.0787                      & 4.0092\(\pm\)2.9922                      & 4.7749\(\pm\)3.7048                      & 5.2517\(\pm\)4.0412                      & 5.4713\(\pm\)4.013                       & 5.5775\(\pm\)3.9294    \\ \cline{2-8} 
                            & CT w/o BRM  & 2.2628\(\pm\)0.1719                      & 2.0385\(\pm\)0.6964                      & 2.4144\(\pm\)0.8376                      & 2.5979\(\pm\)0.8042                      & 2.7848\(\pm\)0.8568                      & 2.8334\(\pm\)0.8094    \\ \cline{2-8} 
                            & CRN         & 2.9282\(\pm\)0.0577                      & 3.831\(\pm\)1.3974                       & 4.1685\(\pm\)1.8389                      & 4.3323\(\pm\)1.9419                      & 4.4152\(\pm\)1.8572                      & 4.4311\(\pm\)1.6918    \\ \cline{2-8} 
                            & CRN w/o BRM & 2.7046\(\pm\)0.0917                      & 3.5286\(\pm\)1.2803                      & 4.0347\(\pm\)1.3757                      & 4.2852\(\pm\)1.3728                      & 4.3668\(\pm\)1.3267                      & 4.2864\(\pm\)1.2416    \\ \cline{2-8} 
                            & RMSN        & 2.8807\(\pm\)0.1817                      & 2.8372\(\pm\)1.121                       & 2.9339\(\pm\)1.1156                      & 2.9181\(\pm\)1.0219                      & 2.893\(\pm\)0.8884                       & 2.8541\(\pm\)0.7413    \\ \cline{2-8} 
                            & GNET        & 2.1707\(\pm\)0.1345                      & 4.068\(\pm\)0.4381                       & 4.5058\(\pm\)0.6647                      & 4.6623\(\pm\)0.8378                      & 4.723\(\pm\)0.9577                       & 4.7171\(\pm\)1.0359    \\ \cline{2-8} 
                            & MSM         & \multicolumn{1}{c|}{3.1887\(\pm\)0.2687} & \multicolumn{1}{c|}{1.0831\(\pm\)0.3845} & \multicolumn{1}{c|}{1.6306\(\pm\)0.5812} & \multicolumn{1}{c|}{2.0902\(\pm\)0.7477} & \multicolumn{1}{c|}{2.5028\(\pm\)0.9024} & 2.9196\(\pm\)1.0679    \\ \hline
\end{tabular}
}
\end{table}

\subsection{Factual Outcome Estimation on Real-world M5 dataset}
The dataset for M5 Forecasting, as referenced in \cite{makridakis2022m5}, encompasses daily transaction figures from Walmart outlets in three American states, supplemented by detailed information on products and stores, in addition to factors like pricing and notable occurrences. This dataset is restructured for the purpose of estimating treatment outcomes, identifying the pricing of products as the treatment factor and product sales figures as the result variable. All other attributes are considered as covariate variables.

Given that this dataset is derived from real-world data and lacks information on counterfactual outcomes, we demonstrate the performance of various models and their Empirical Risk Minimization (ERM) variants in estimating factual outcomes. We omit GENT and MSM here because they can not converge on this dataset. Similarly, we set the horizon $\tau$ from 1 to 6, and the results, as shown in Table \ref{M5_dataset_prediction}, indicate that ERM variants without a balancing module generally outperform those with a balancing module. This outcome is intuitive since we are dealing with factual outcomes, and the presence of a balancing module alters the original distribution of covariates, causing a loss of information beneficial to outcome prediction.

\subsection{Multi-step Counterfactual Outcome in Standard Supervised Learning}
reports on the performance of various models and their ERM variants in counterfactual estimation on the Tumor dataset, with the treatment bias strength parameter $\gamma$, set as 2, 6, and 8. The results, illustrated in Table \ref{tumor_2_6_8_standard}, corroborate the phenomena discussed in the main text. It was found that ERM variants without a balancing module generally exhibit superior predictive performance compared to those with a balancing module. Moreover, as the intensity of treatment bias increases, the negative impact on model performance brought about by the balancing strategy becomes more pronounced (as evidenced by the increase in the Root Mean Square Error (RMSE) average) and more unstable (as can be discerned from the standard deviation of the RMSE). This highlights the challenges in managing treatment bias in models, especially when balancing mechanisms are employed, suggesting a careful consideration of the trade-off between bias correction and predictive accuracy.

\subsection{Additional Short-term History Cold-start Case for Counterfactual Estimation}
In this report, we detail additional findings on the performance of counterfactual outcome estimation in the context of short-term history cold-start cases on the Tumor dataset, under varying intensities of treatment bias. Specifically, we compared the performance of different counterfactual estimation models and their Empirical Risk Minimization (ERM) variants in multi-step prediction scenarios, with $\gamma$ values set at 2, 6, and 8, as depicted in Table \ref{short_term_2_6_8}. Our observations reveal that while models with a balancing strategy may exhibit lower average Root Mean Square Error (RMSE) values in certain cases compared to those without a balancing strategy, the significant difference in the standard deviation between the two groups undermines the potential advantages of balanced models in these instances. Furthermore, it is noteworthy that in the majority of cases, models lacking a balancing strategy tend to perform better. This suggests that the purported benefits of implementing a balancing mechanism for counterfactual outcome prediction in cold-start scenarios with short-term history may not be consistently demonstrable across different levels of treatment bias, especially when considering the variability in model performance.

\begin{figure}[t]
\centering
\subfigure[Causal Transformer, $\gamma=1$]{
\begin{minipage}{0.33\linewidth}
\centering
\includegraphics[width=1.6in]{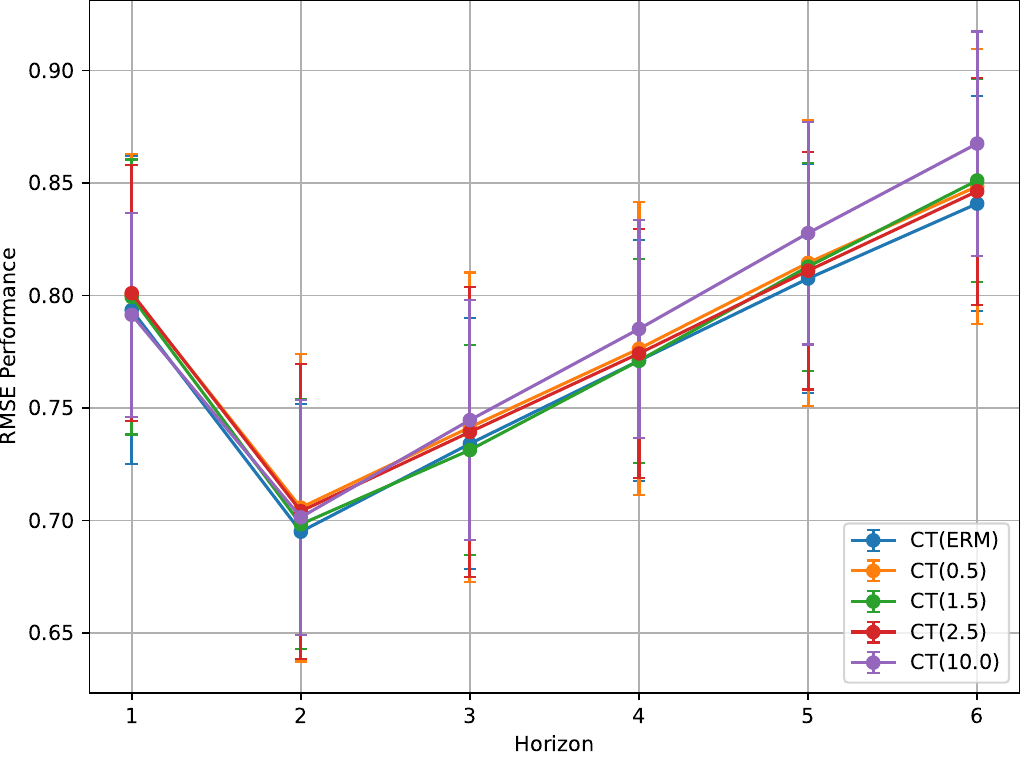}
\end{minipage}%
}%
\centering
\subfigure[Causal Transformer, $\gamma=3$]{
\begin{minipage}{0.33\linewidth}
\centering
\includegraphics[width=1.6in]{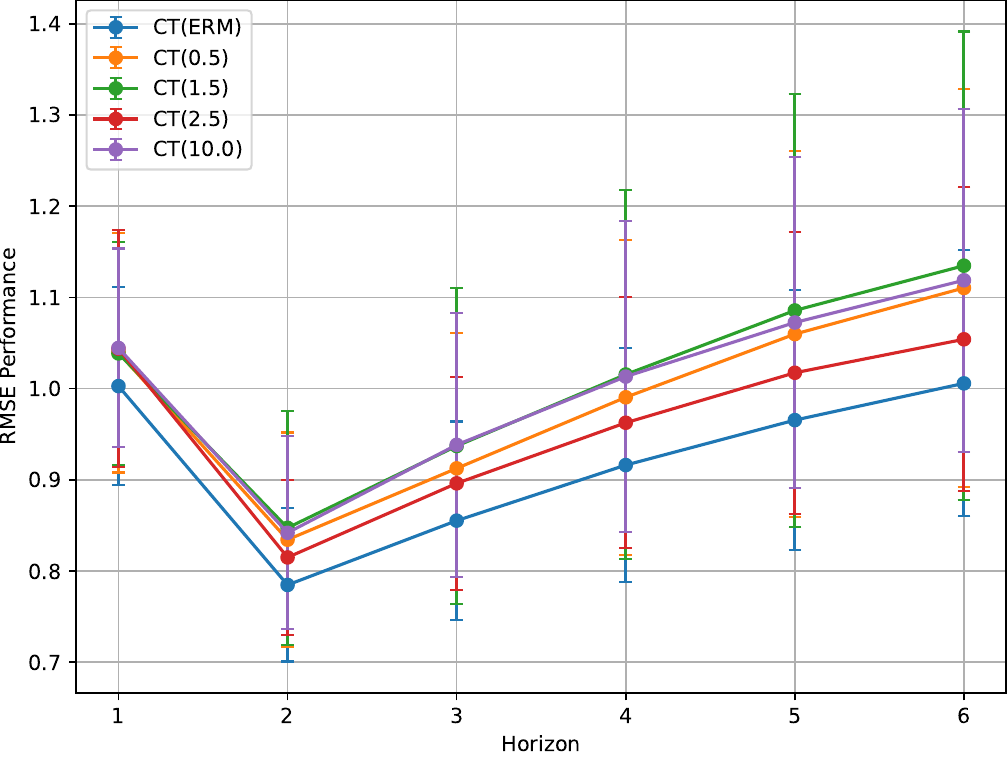}
\end{minipage}%
}%
\subfigure[Causal Transformer, $\gamma=10$]{
\begin{minipage}{0.33\linewidth}
\centering
\includegraphics[width=1.6in]{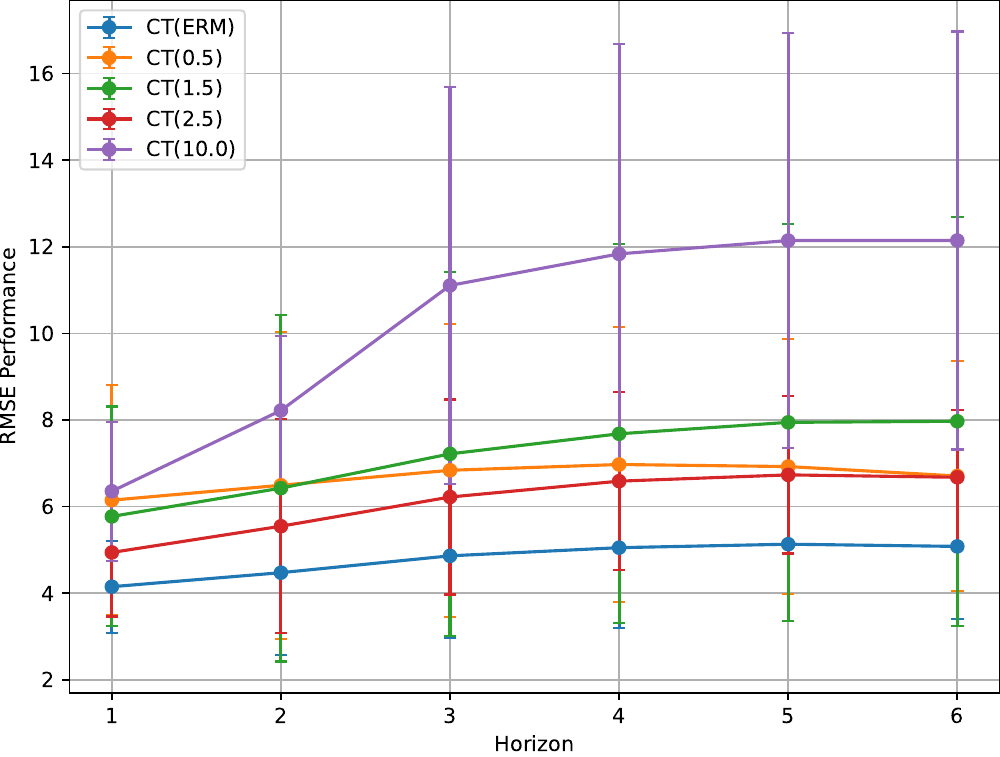}
\end{minipage}%
}%
\\
\centering
\subfigure[CRN, $\gamma=1$]{
\begin{minipage}{0.33\linewidth}
\centering
\includegraphics[width=1.6in]{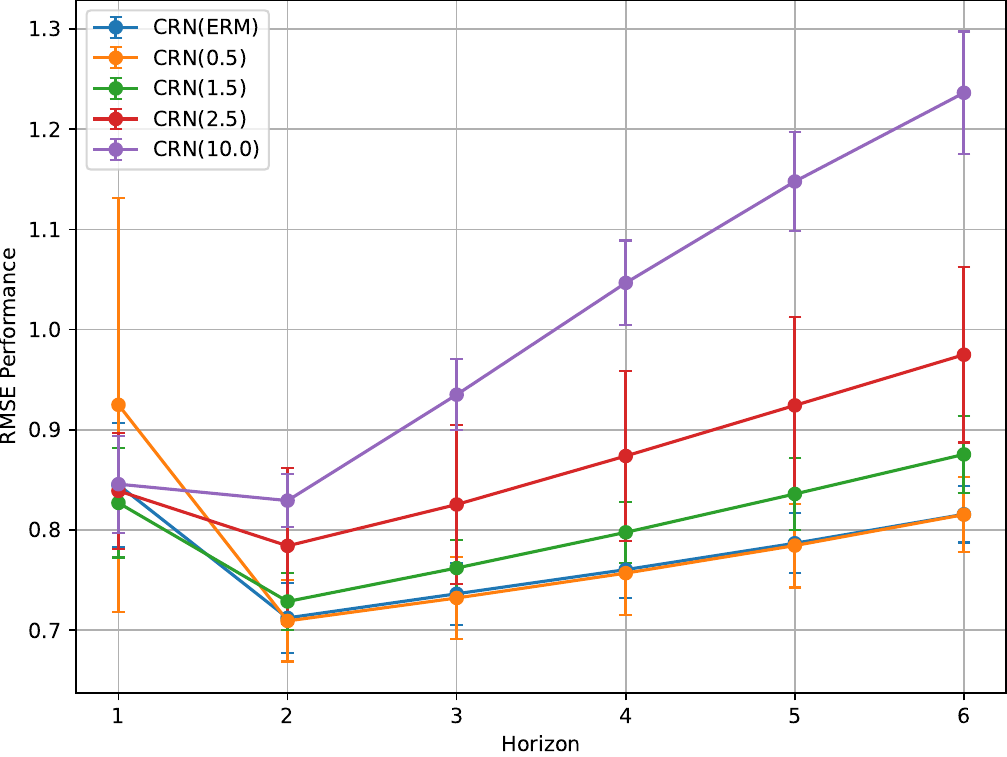}
\end{minipage}%
}%
\centering
\subfigure[CRN, $\gamma=3$]{
\begin{minipage}{0.33\linewidth}
\centering
\includegraphics[width=1.6in]{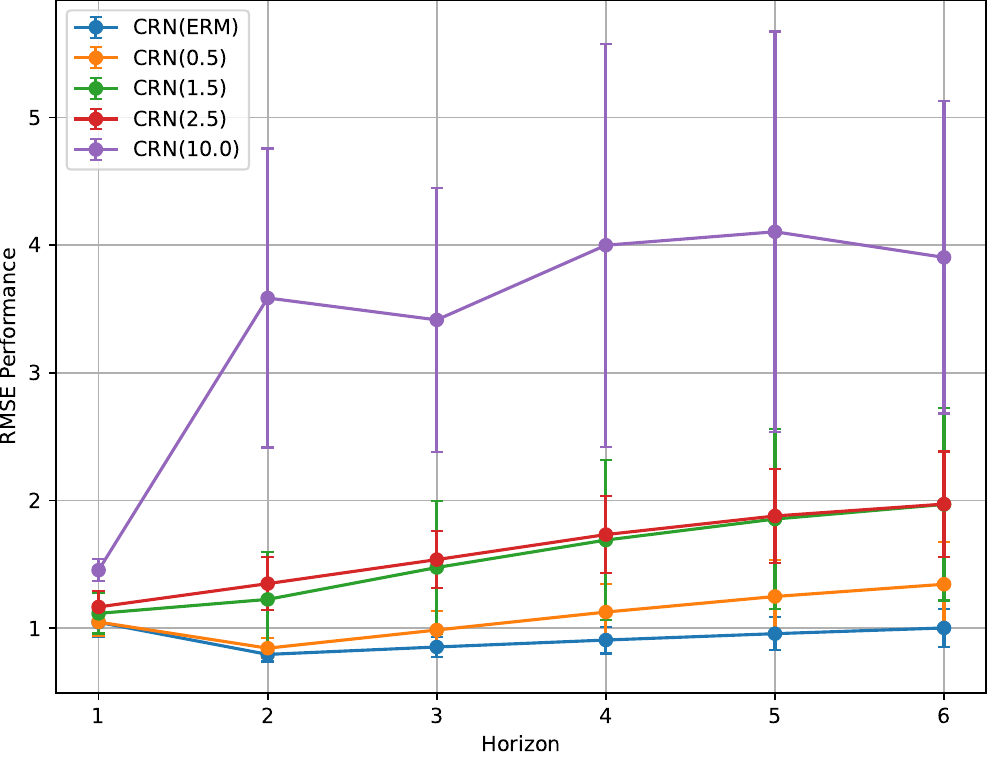}
\end{minipage}%
}%
\subfigure[CRN, $\gamma=10$]{
\begin{minipage}{0.33\linewidth}
\centering
\includegraphics[width=1.6in]{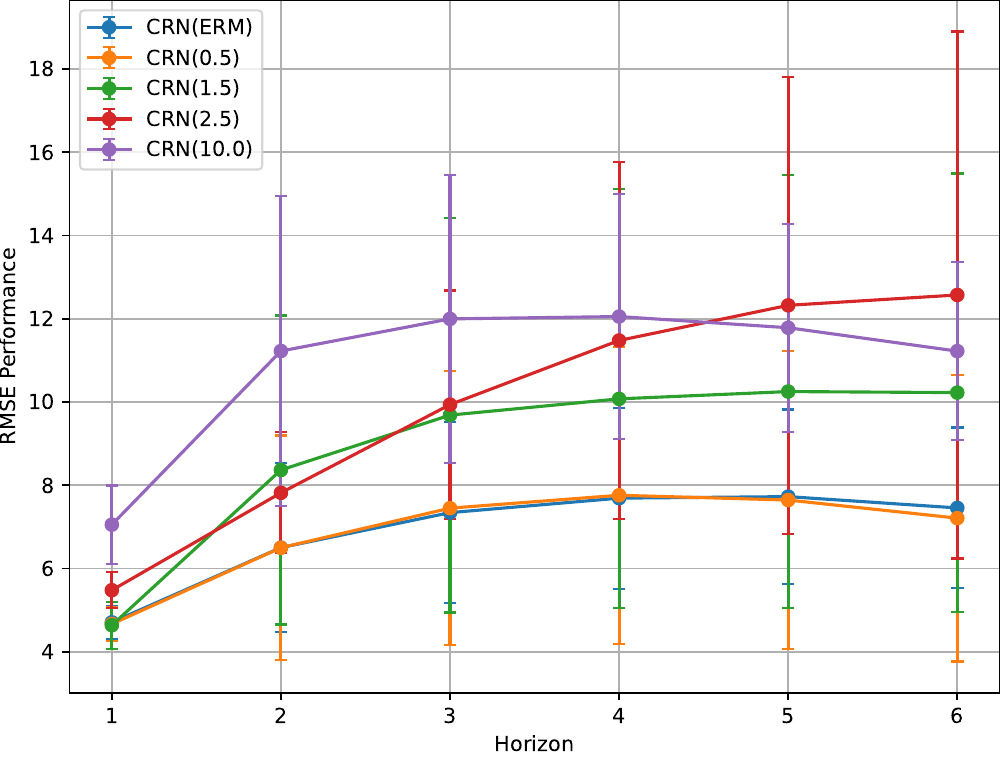}
\end{minipage}%
}%
\centering
\caption{Model performance comparison in different weights of balancing module, where CT with Counterfactual Domain Confusion, CRN with Adversarial Gradient Reversal.}
\label{fig_balance_weight}
\end{figure}

\subsection{Additional Distribution Shift Cold-start Case for Counterfactual Estimation}

In this report, we present additional experimental results on counterfactual outcome estimation under the scenario of distribution shift cold-start Case, involving various models and their ERM variants. These models were trained with a treatment bias strength of $\gamma=10$, and then tested on datasets constructed with the treatment bias strength of 
$\gamma=2,6$ to evaluate the effectiveness of balancing strategies when there are distributional differences between the source and target domains. The findings, as displayed in Table \ref{disreibution_shift_2_6_8}, consistently show that ERM variants without a balancing module generally outperform those with a balancing module. We did not observe any benefit from the balancing strategy in the context of distribution shift cases for counterfactual estimation tasks; instead, it appears to have a detrimental effect, introducing instability into the model's performance. This suggests that in scenarios where the training and testing data distributions differ significantly, employing a balancing strategy may not enhance, and could potentially impair, the stability and accuracy of counterfactual outcome predictions.

\subsection{Exploring the Impact of Contribution Weight for Balancing Module}
In this section, we conduct the experiments to explore the impact of the contribution weights of the two balanced representation methods: Adversarial Gradient Reversal, Counterfactual Domain Confusion on model performance in temporal counterfactual outcome estimation tasks for the synthetic Tumor Growth dataset. Due to the poor performance of PS-based Contrastive Balancing, we omit the exploration for it.  These methods were initially mentioned in Section 3. In our analysis, we denote the contribution weights of these balancing modules as $\alpha$, with values set within the range of [0.5, 1.5, 2.5, 10.0]. We then compare their results with the ERM version (where $\alpha=0$) of each model, while keeping other hyperparameters consistent with those used in standard supervised learning settings. Experimental results, as illustrated in Figure \ref{fig_balance_weight}, demonstrate the performance of different balanced representation module weights under varying values of $\gamma=[1,3,10]$. 

Our observations indicate that when $\gamma$ values are relatively low (e.g., 1 or 3), signifying less significant treatment bias, the weight values of the balanced representation modules have a minimal impact on the performance of counterfactual outcome estimation, the model performance in terms of different balancing weights is almost the same. However, as the $\gamma$ value increases, a general trend emerges: with the rise in the balancing module’s weight, there is a corresponding decline in performance. Notably, compared to the ERM models, the performance of models with higher balancing module weights tends to be less effective, which is consistent with the previous observation in the standard supervised learning setting. This finding underscores the negative role that balancing module weights play in temporal counterfactual outcome estimation, particularly in scenarios characterized by more pronounced treatment biases.

\subsection{Detailed Analysis for the Covariate Distribution}
\begin{figure}[t]
\centering
\subfigure[$\gamma=0$]{
\begin{minipage}{0.166\linewidth}
\centering
\includegraphics[width=1.0in]{first_gamma_0.pdf}
\end{minipage}%
}%
\centering
\subfigure[$\gamma=1$]{
\begin{minipage}{0.166\linewidth}
\centering
\includegraphics[width=1.0in]{first_gamma_1.pdf}
\end{minipage}%
}%
\subfigure[$\gamma=2$]{
\begin{minipage}{0.166\linewidth}
\centering
\includegraphics[width=1.0in]{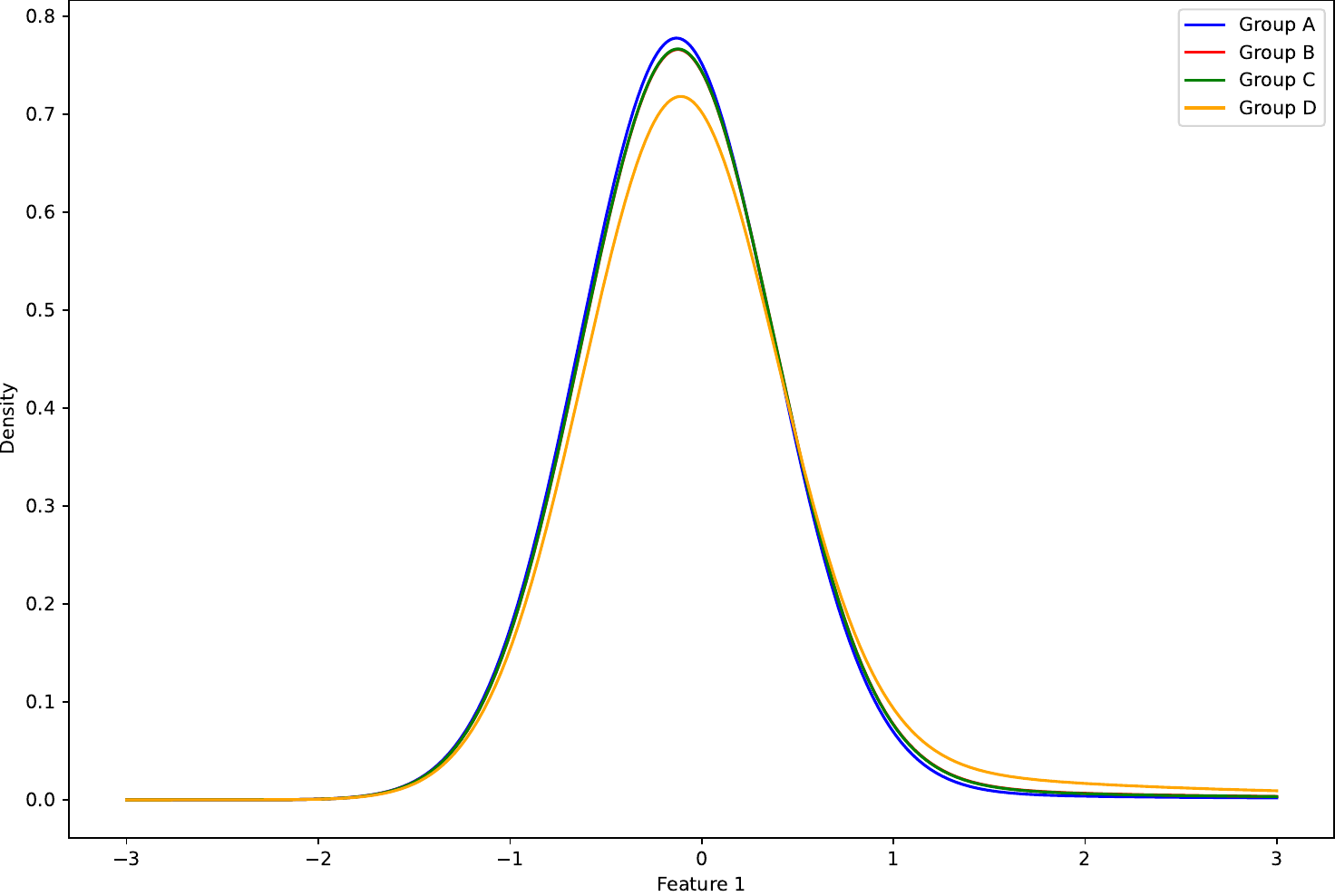}
\end{minipage}%
}%
\centering
\subfigure[$\gamma=3$]{
\begin{minipage}{0.166\linewidth}
\centering
\includegraphics[width=1.0in]{first_gamma_3.pdf}
\end{minipage}%
}%
\centering
\subfigure[$\gamma=6$]{
\begin{minipage}{0.166\linewidth}
\centering
\includegraphics[width=1.0in]{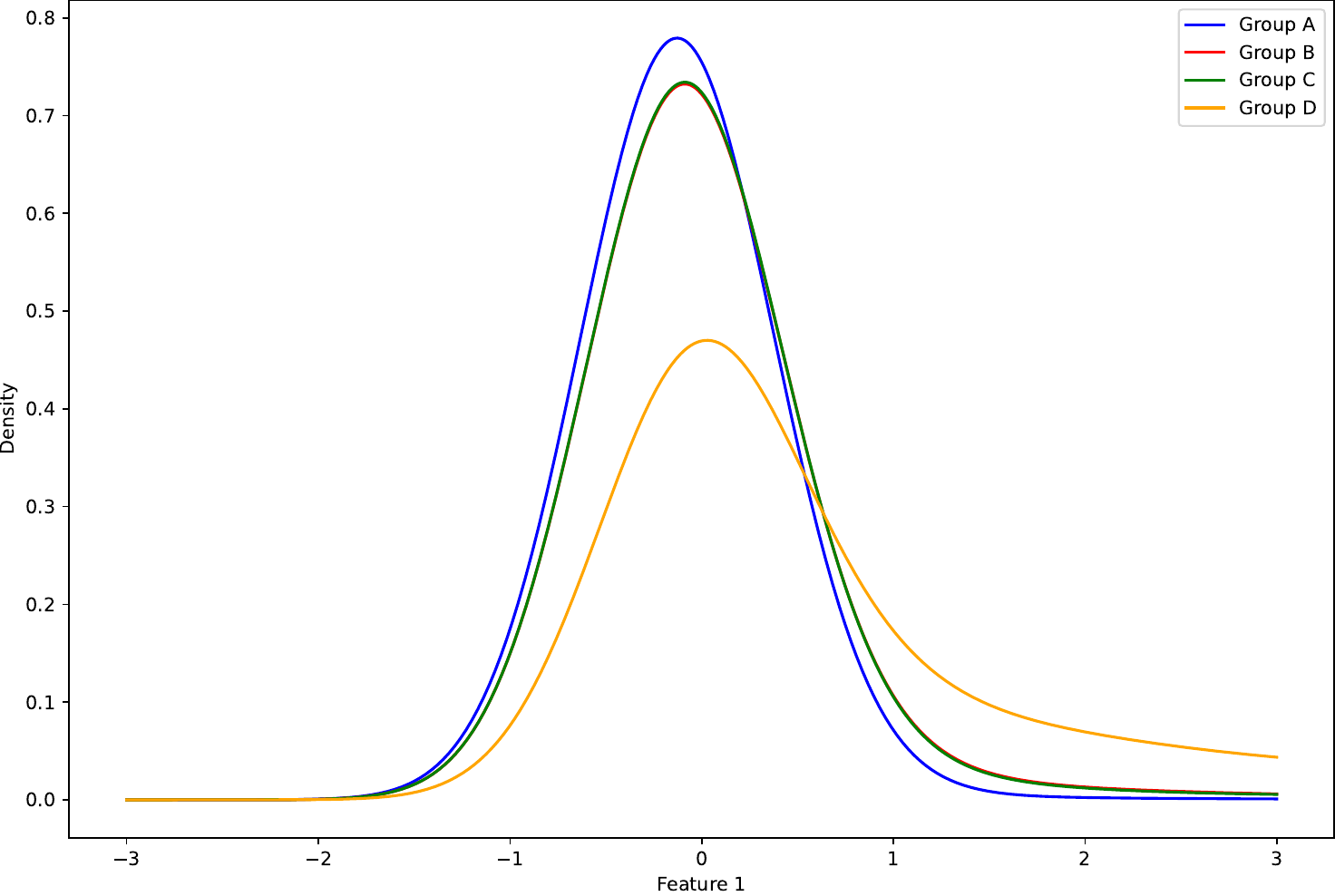}
\end{minipage}%
}%
\subfigure[$\gamma=10$]{
\begin{minipage}{0.166\linewidth}
\centering
\includegraphics[width=1.0in]{first_gamma_10.pdf}
\end{minipage}%
}%
\centering
\\
\subfigure[$\gamma=0$]{
\begin{minipage}{0.166\linewidth}
\centering
\includegraphics[width=1.0in]{second_gamma_0.pdf}
\end{minipage}%
}%
\centering
\subfigure[$\gamma=1$]{
\begin{minipage}{0.166\linewidth}
\centering
\includegraphics[width=1.0in]{second_gamma_1.pdf}
\end{minipage}%
}%
\subfigure[$\gamma=2$]{
\begin{minipage}{0.166\linewidth}
\centering
\includegraphics[width=1.0in]{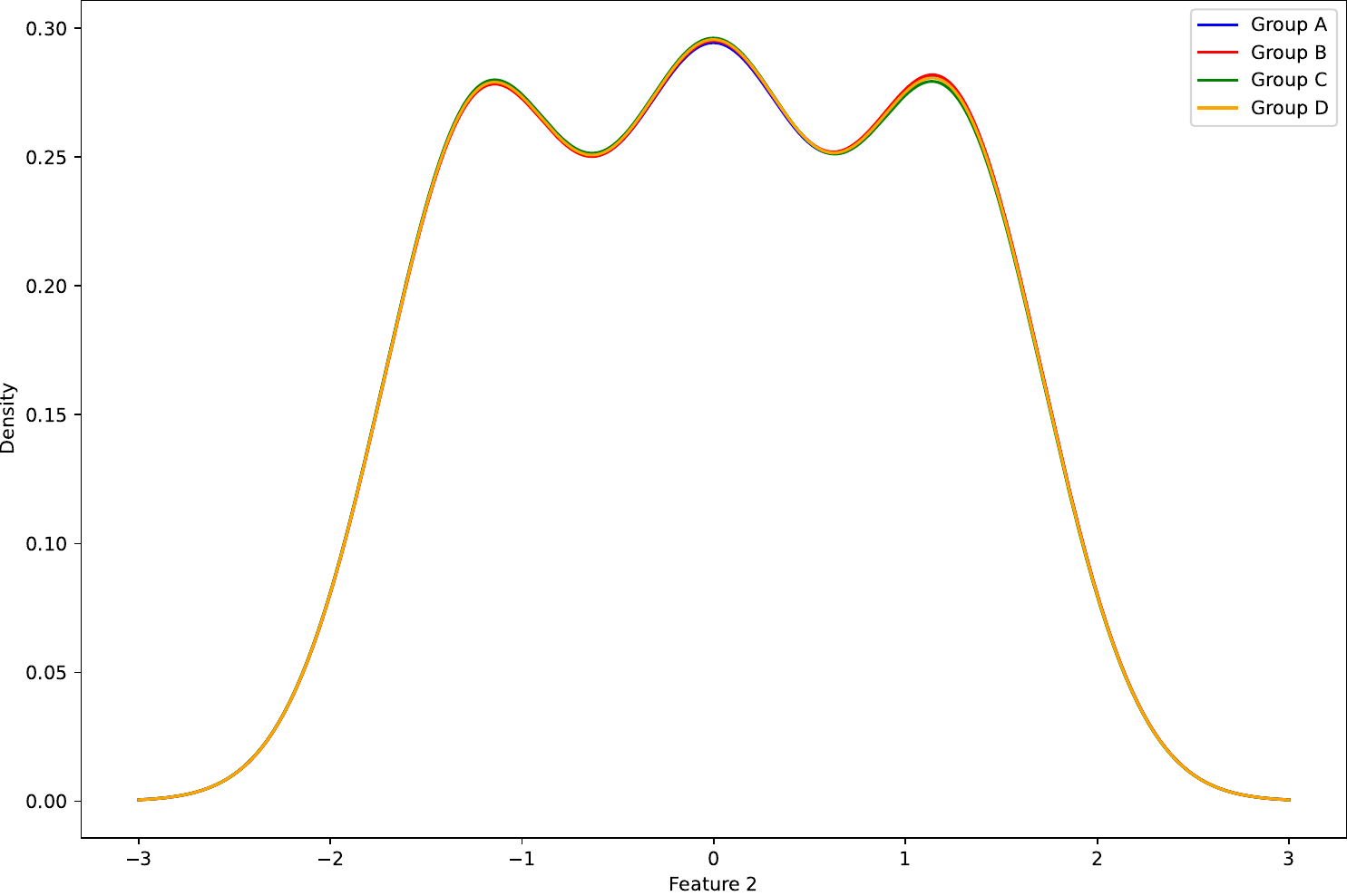}
\end{minipage}%
}%
\centering
\subfigure[$\gamma=3$]{
\begin{minipage}{0.166\linewidth}
\centering
\includegraphics[width=1.0in]{second_gamma_3.pdf}
\end{minipage}%
}%
\centering
\subfigure[$\gamma=6$]{
\begin{minipage}{0.166\linewidth}
\centering
\includegraphics[width=1.0in]{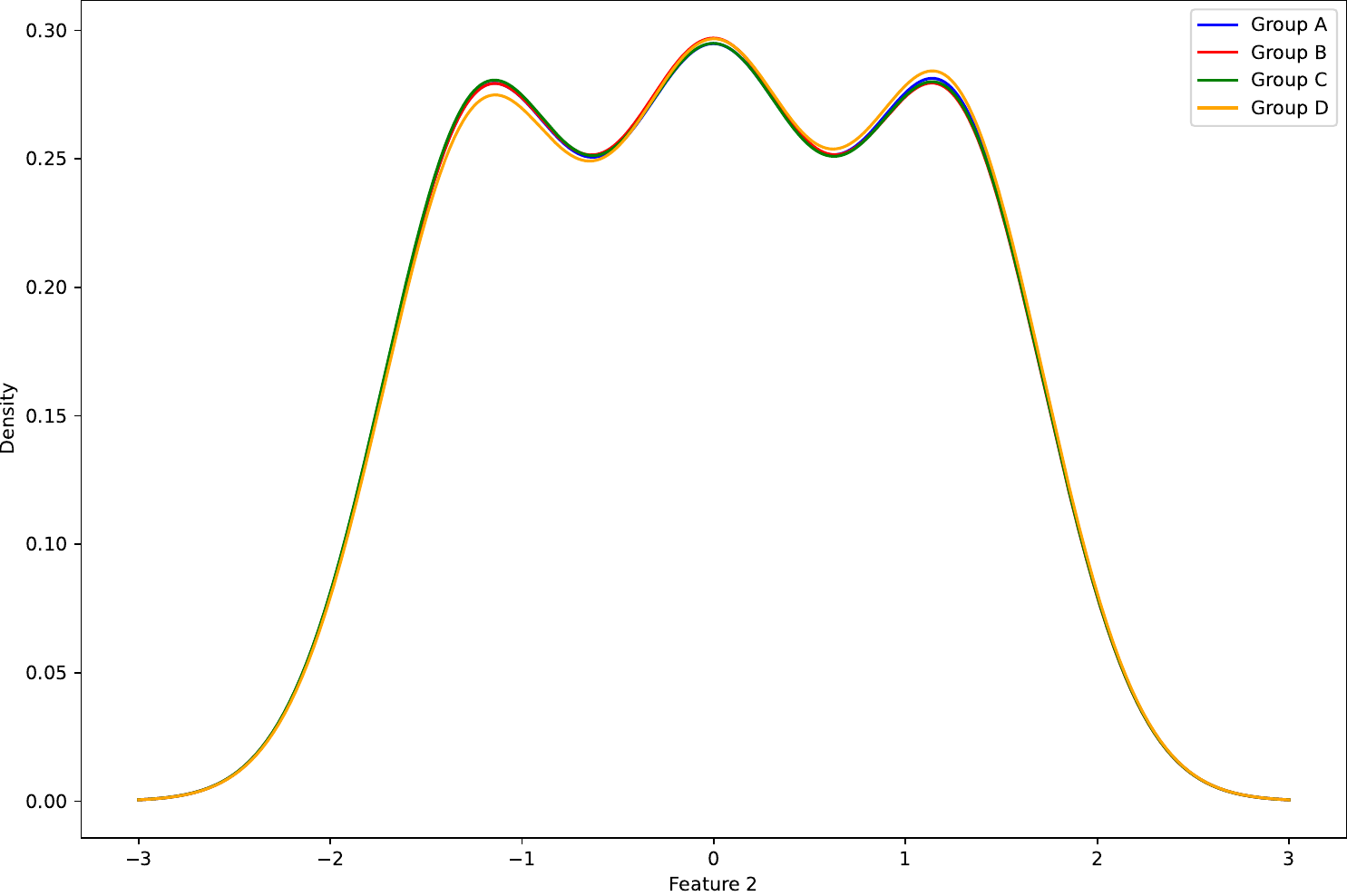}
\end{minipage}%
}%
\subfigure[$\gamma=10$]{
\begin{minipage}{0.166\linewidth}
\centering
\includegraphics[width=1.0in]{second_gamma_10.pdf}
\end{minipage}%
}%
\caption{Distribution shape for the two covariates under different strengths of treatment bias. (a)-(f) for the first covariate, (g)-(l) for the second covariate.}
\label{fig_dis_cov_appendix}
\end{figure}

\textbf{Tumor Growth}. Here we present the covariate distribution averaged on the all-time steps for treatment bias strength $gamma=\{0,1,2,3,6,10\}$. The Gaussian fit distributions of the first covariate and the second covariate are shown in Figures \ref{fig_dis_cov_appendix}. We can observe that for the first covariate, when the treatment bias intensity is relatively low (e.g., $\gamma$ equals 0, 1, 2, 3), the differences in the distribution of this variable among different groups are not very pronounced. It's only when the treatment bias is relatively high that noticeable differences in distribution between groups emerge. As for the second covariate, regardless of the intensity of the treatment bias, the distribution of this covariate among different groups remains nearly identical. 

Therefore, based on the results presented above, this may be the reason why the balanced representation module of the causal time-series model doesn't work: in this dataset, the distribution of one covariate is the same across different groups, and the distribution of the other covariate only exhibits noticeable differences among groups when the treatment bias is relatively high. As mentioned at the outset, the Balanced Representation Module aims to reduce the distance between covariate distributions among different groups. However, if the disparities in covariate distributions among different groups in the data are not sufficiently large, this module is likely to become a burden to the model, leading to a decrease in predictive performance.

\begin{figure}[h]
\centering
\subfigure[$\gamma=0.2, t=1$]{
\begin{minipage}{0.2\linewidth}
\centering
\includegraphics[width=1.0in]{pure_gamma_0.2_step_1.pdf}
\end{minipage}%
}%
\centering
\subfigure[$\gamma=0.2, t=2$]{
\begin{minipage}{0.2\linewidth}
\centering
\includegraphics[width=1.0in]{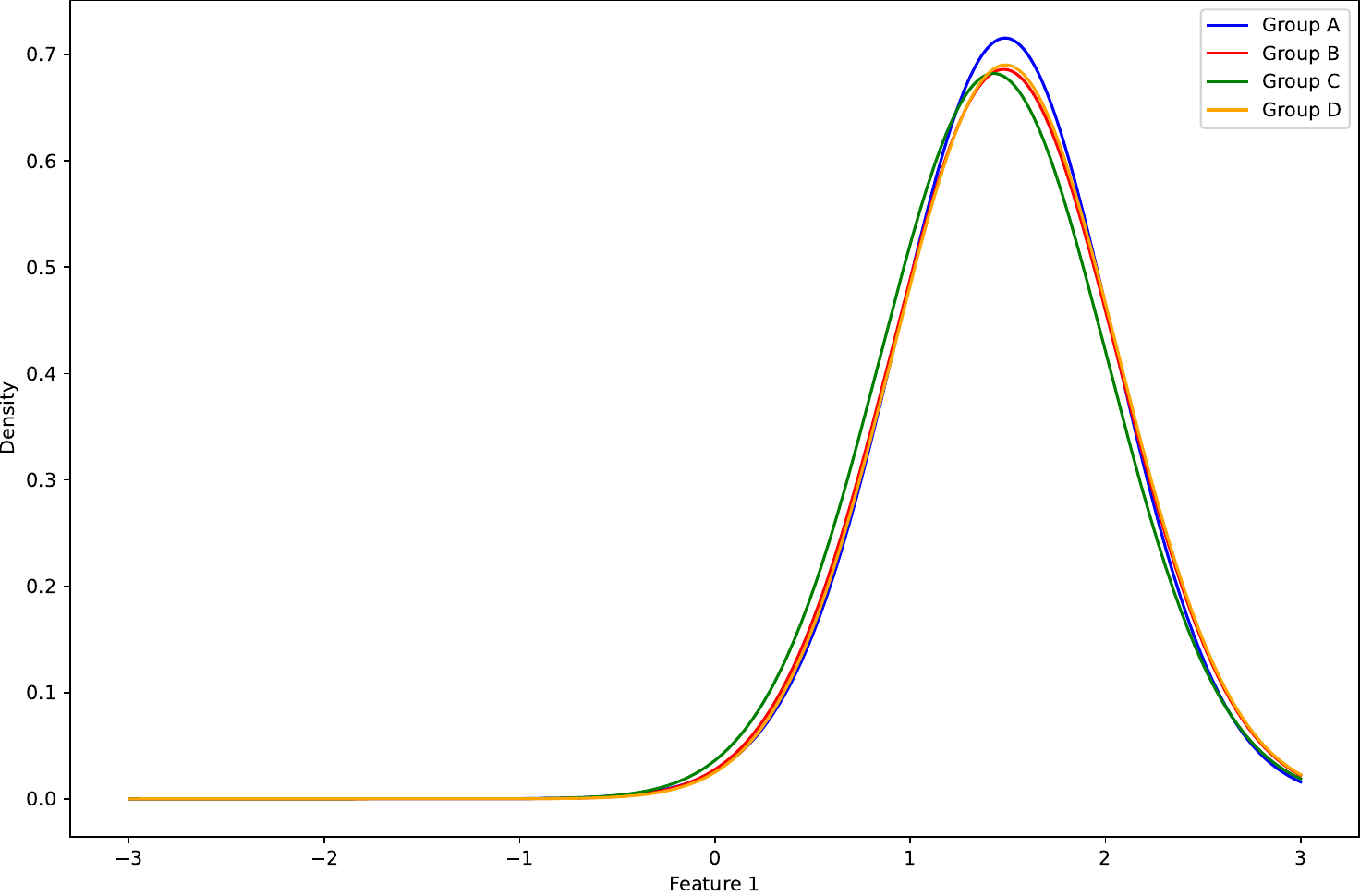}
\end{minipage}%
}%
\subfigure[$\gamma=0.2, t=4$]{
\begin{minipage}{0.2\linewidth}
\centering
\includegraphics[width=1.0in]{pure_gamma_0.2_step_4.pdf}
\end{minipage}%
}%
\centering
\subfigure[$\gamma=0.2, t=8$]{
\begin{minipage}{0.2\linewidth}
\centering
\includegraphics[width=1.0in]{pure_gamma_0.2_step_8.pdf}
\end{minipage}%
}%
\centering
\subfigure[$\gamma=0.2, t=16$]{
\begin{minipage}{0.2\linewidth}
\centering
\includegraphics[width=1.0in]{pure_gamma_0.2_step_16.pdf}
\end{minipage}%
}%
\\
\subfigure[$\gamma=0.4, t=1$]{
\begin{minipage}{0.2\linewidth}
\centering
\includegraphics[width=1.0in]{pure_gamma_0.4_step_1.pdf}
\end{minipage}%
}%
\centering
\subfigure[$\gamma=0.4, t=2$]{
\begin{minipage}{0.2\linewidth}
\centering
\includegraphics[width=1.0in]{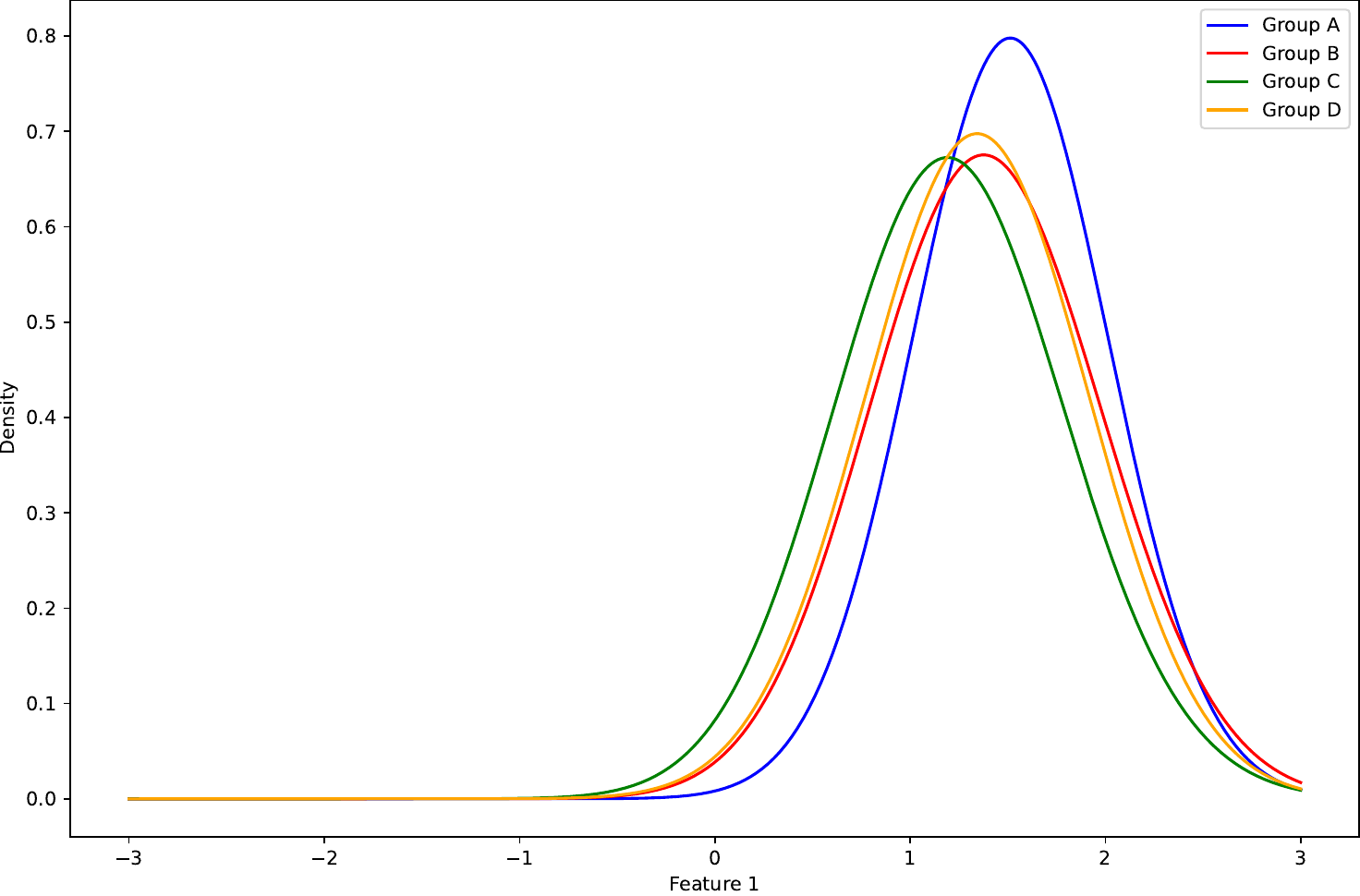}
\end{minipage}%
}%
\subfigure[$\gamma=0.4, t=4$]{
\begin{minipage}{0.2\linewidth}
\centering
\includegraphics[width=1.0in]{pure_gamma_0.4_step_4.pdf}
\end{minipage}%
}%
\centering
\subfigure[$\gamma=0.4, t=8$]{
\begin{minipage}{0.2\linewidth}
\centering
\includegraphics[width=1.0in]{pure_gamma_0.4_step_8.pdf}
\end{minipage}%
}%
\centering
\subfigure[$\gamma=0.4, t=16$]{
\begin{minipage}{0.2\linewidth}
\centering
\includegraphics[width=1.0in]{pure_gamma_0.4_step_16.pdf}
\end{minipage}%
}%
\caption{Distribution shape for the covariate under different strengths of treatment bias and time-steps on pure synthetic dataset.}
\label{fig_dis_cov_pure_appendix}
\end{figure}

\textbf{Pure Synthetic Data}. Furthermore, to avoid losing generality, we present the visualization of covariate distribution for more time-step selection. We set the sampled time step to \{1,2,4,8,16\}. We plot the visualization of one certain covariate over the sampled time steps as shown in Figure \ref{fig_dis_cov_pure_appendix}. The results reveal that at the onset of the timeline (e.g., at time-steps 1 and 2), an increase in the gamma value correlates with a more pronounced disparity in covariate distributions among different treatment groups. Yet, as the timeline progresses to later stages (e.g., at time step 8 and 16), the variance in covariate distributions between treatment groups diminishes. This pattern suggests that, in line with typical time series assumptions, treatment bias tends to decrease over time, thereby reducing the efficacy of strategies aimed at balancing representation distributions.

\subsection{Learned Representations for Reconstruction of Covariates}

\begin{figure}[h]
\centering
\centering
\subfigure[$\gamma=10$, Balanced CRN]{
\begin{minipage}{0.25\linewidth}
\centering
\includegraphics[width=1.6in]{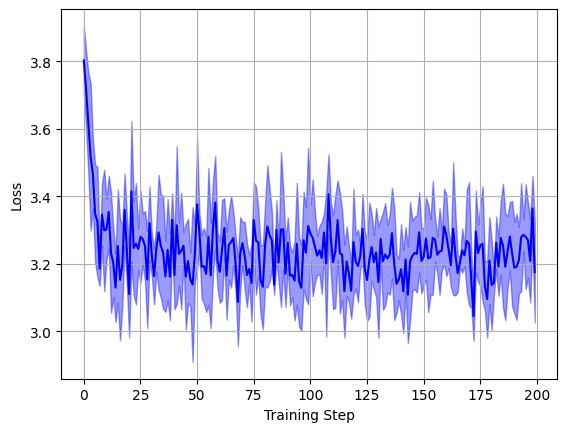}
\end{minipage}%
}%
\centering
\subfigure[$\gamma=10$, Non-balanced CRN]{
\begin{minipage}{0.25\linewidth}
\centering
\includegraphics[width=1.6in]{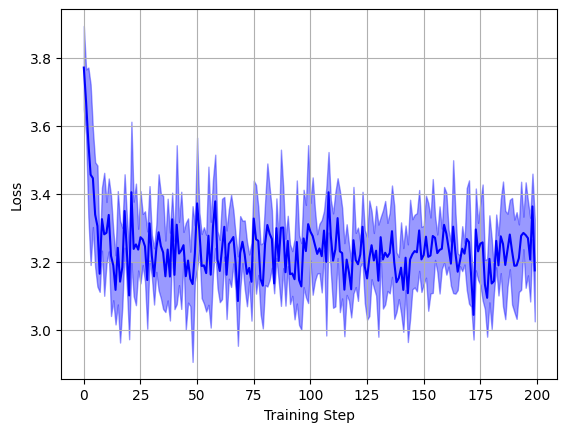}
\end{minipage}%
}%
\subfigure[$\gamma=1$, Balanced CRN]{
\begin{minipage}{0.25\linewidth}
\centering
\includegraphics[width=1.6in]{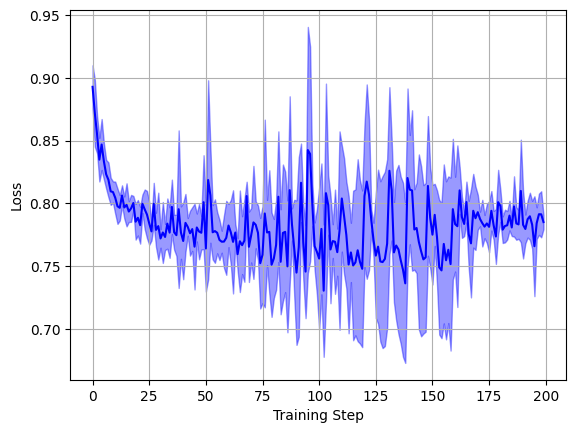}
\end{minipage}%
}%
\subfigure[$\gamma=1$, Non-balanced CRN]{
\begin{minipage}{0.25\linewidth}
\centering
\includegraphics[width=1.6in]{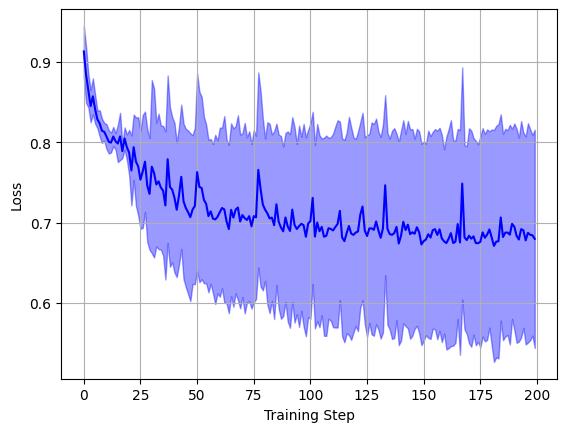}
\end{minipage}%
}
\centering
\caption{Reconstruction loss curves for CRN model.}
\label{fig_reconsreuction_loss_crn}
\end{figure}

In the main text, we present the reconstruction results of the original covariates by the Causal Transformer under scenarios of both mild and significant treatment bias, indicated by $\gamma=1$ and 10, respectively. This is done to validate the impact of the balanced representation strategy on the loss of covariate information when employing balancing strategies versus not. Here, we also display the capability of the CRN model to reconstruct the original covariates under the same settings, using two types of representations as illustrated in Figure \ref{fig_reconsreuction_loss_crn}. We observe that under conditions of mild treatment bias ($\gamma=1$), the training for the task of reconstructing original covariates with non-balanced representation is more stable than with balanced representation. The loss for the former reasonably and steadily declines, whereas the latter's loss fluctuates continuously. However, under conditions of significant treatment bias ($\gamma=10$), both balanced and non-balanced representations show poor convergence in training. This suggests that excessive treatment bias can distort the training process of capturing useful information from the original covariates, leading to unstable prediction outcomes. 

\subsection{Representation Visualization for CRN model}

\begin{figure}[t]
\centering
\subfigure[$t=1$, CRN, Balanced]{
\begin{minipage}{0.25\linewidth}
\centering
\includegraphics[width=1.6in]{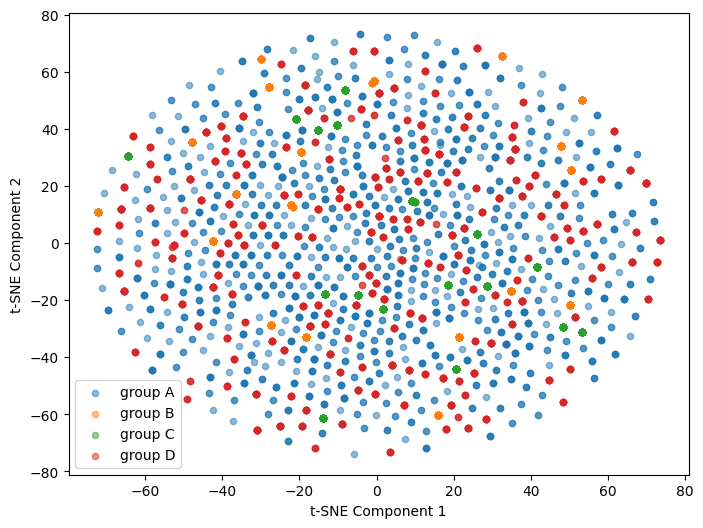}
\end{minipage}%
}%
\centering
\subfigure[$t=10, CRN, Balanced$]{
\begin{minipage}{0.25\linewidth}
\centering
\includegraphics[width=1.6in]{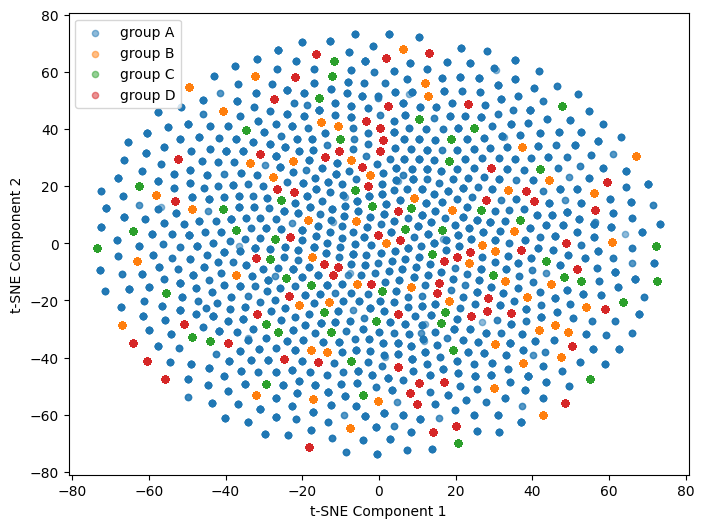}
\end{minipage}%
}%
\centering
\subfigure[$t=1$, CRN, Non-balanced]{
\begin{minipage}{0.25\linewidth}
\centering
\includegraphics[width=1.6in]{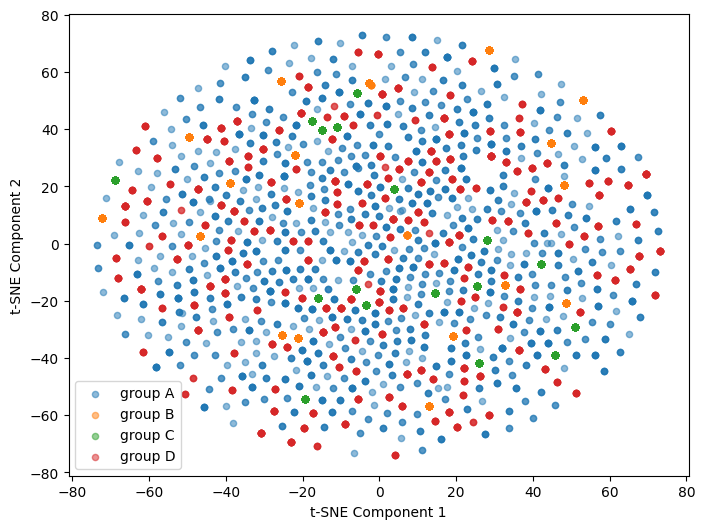}
\end{minipage}%
}%
\centering
\subfigure[$t=10$, CRN, Non-balanced]{
\begin{minipage}{0.25\linewidth}
\centering
\includegraphics[width=1.6in]{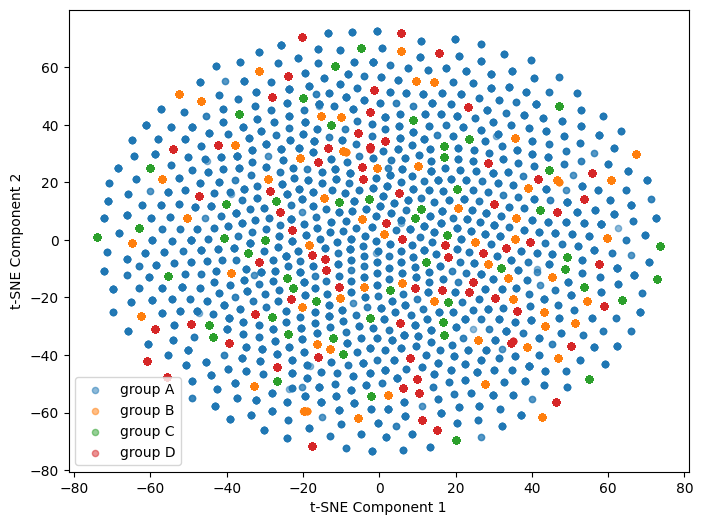}
\end{minipage}%
}%
\centering
\caption{Visualization for balanced and non-balanced representation for different time steps for CRN models.}
\label{rep_visual_crn}
\end{figure}

Here, we report on the visualization of representations derived from the CRN model, both with and without the implementation of a balancing strategy. Similarly, we set the intensity of treatment bias, $\gamma$, to 10 and obtained representations for time steps $t=1$ and $t=10$. The visualization results, as shown in Figure \ref{rep_visual_crn}, indicate that the distribution differences between the balanced and non-balanced representations, after dimensionality reduction, are not significant. Under conditions of pronounced treatment bias, the balancing module does not appear to be effective. This observation corroborates our previous discussion and analysis that excessive treatment bias can disrupt the model's learning process. The visualizations provide a tangible insight into how, in the presence of significant treatment bias, both balanced and non-balanced strategies result in similar distributions of representations. This suggests a limitation in the balancing module's ability to mitigate the effects of treatment bias when it is substantial, affirming the notion that an overly pronounced treatment bias poses a challenge to the model's ability to learn and differentiate effectively between treated and untreated groups based on the original covariates.

\end{document}

%% file: math_defi.tex
\usepackage{amsmath,amsfonts,bm}

\def\eqref#1{equation~\ref{#1}}

\def\1{\bm{1}}

\def\va{{\bm{a}}}

\def\vv{{\bm{v}}}

\def\vx{{\bm{x}}}
\def\vy{{\bm{y}}}

\def\mA{{\bm{A}}}

\def\mH{{\bm{H}}}

\def\mV{{\bm{V}}}

\def\mX{{\bm{X}}}
\def\mY{{\bm{Y}}}

\DeclareMathAlphabet{\mathsfit}{\encodingdefault}{\sfdefault}{m}{sl}
\SetMathAlphabet{\mathsfit}{bold}{\encodingdefault}{\sfdefault}{bx}{n}

\newcommand{\E}{\mathbb{E}}

